\documentclass[twoside,11pt]{article}

\usepackage[
    a4paper,
    left=3cm,
    right=3cm,
    top=3cm,
    bottom=3cm
]{geometry}

\usepackage[T1]{fontenc}
\usepackage[utf8]{inputenc}
\usepackage{lmodern}
\usepackage{libertine}

\usepackage{xcolor}
\definecolor{linkblue}{RGB}{0,90,180}
\definecolor{lightlinkblue}{RGB}{70,150,230}
\definecolor{condPink}{HTML}{F472B6}
\definecolor{condBlue}{HTML}{38BDF8}
\definecolor{condGreen}{HTML}{34D399}

\usepackage{amsmath}
\usepackage{amssymb}
\usepackage{mathtools}
\usepackage{amsthm}
\usepackage{mathrsfs}
\usepackage{bbm}
\usepackage{bm}
\usepackage{upgreek}

\usepackage{graphicx}
\usepackage{booktabs}
\usepackage{tabularx}
\usepackage{makecell}
\usepackage{multirow,multicol}
\usepackage{enumitem}
\usepackage[font=small,labelfont=bf]{caption}
\usepackage{subcaption}
\usepackage{quoting}
\usepackage{xspace}
\usepackage{comment}
\usepackage{pifont}
\usepackage{lineno}
\usepackage{tikz}
\usetikzlibrary{arrows.meta}

\graphicspath{{JMLR/}{JMLR/images/}{./}}
\DeclareGraphicsExtensions{.pdf,.png,.jpg,.jpeg}

\usepackage[numbers,sort&compress]{natbib}

\usepackage[
    pdftex,
    colorlinks=true,
    bookmarksopen=true,
    linkcolor=lightlinkblue,
    citecolor=linkblue,
    urlcolor=linkblue
]{hyperref}
\usepackage{cleveref}

\theoremstyle{plain}
\newtheorem{theorem}{Theorem}[section]
\newtheorem{proposition}[theorem]{Proposition}
\newtheorem{lemma}[theorem]{Lemma}
\newtheorem{corollary}[theorem]{Corollary}
\newtheorem{definition}[theorem]{Definition}

\newtheorem{remark}[theorem]{Remark}
\newtheorem{example}[theorem]{Example}

\DeclareMathOperator{\diam}{diam}

\DeclareMathOperator{\supp}{supp}

\newcommand{\diff}{\,\mathrm{d}}

\newcommand{\Lip}{\mathrm{Lip}}

\newcommand{\Id}{\mathsf{Id}}
\newcommand{\KL}{\mathsf{KL}}
\newcommand{\BV}{\mathsf{BV}}
\newcommand{\TV}{\mathsf{TV}}
\newcommand{\JS}{\mathsf{JS}}
\newcommand{\KR}{\mathsf{KR}}

\renewcommand{\leq}{\leqslant}
\renewcommand{\geq}{\geqslant}

\newcommand{\N}{\mathbb{N}}

\newcommand{\R}{\mathbb{R}}

\newcommand{\Z}{\mathbb{Z}}

\newcommand{\cP}{\mathcal{P}}

\newcommand{\Pac}{\mathcal{P}_{\mathrm{ac}}}

\newcommand{\UIP}{\ensuremath{\mathrm{UIP}}\xspace}
\newcommand{\UAP}{\ensuremath{\mathrm{UAP}}\xspace}
\newcommand{\uUIP}{\ensuremath{\mathrm{u}\text{-}\mathrm{UIP}}\xspace}
\newcommand{\DSP}{\ensuremath{\mathrm{DSP}}\xspace}

\newcommand\restr[2]{{%
  \left.\kern-\nulldelimiterspace
  #1
  \vphantom{\big|}
  \right|_{#2}
}}

\newcommand{\zl}[1]{}
\newcommand{\cj}[1]{}
\newcommand{\nbaa}[1]{}
\newcommand{\nbll}[1]{}

\newcommand{\cjp}[1]{\textcolor{cyan}{\text{Jingpu: }#1}}

\title{Point interpolation and distributional universality in flow-based generative models}

\author{
Antonio \'Alvarez-L\'opez\textsuperscript{1} \qquad
Jingpu Cheng\textsuperscript{2} \qquad
Liyan Zhang\textsuperscript{2} \qquad
Qianxiao Li\textsuperscript{2,3}
}

\date{\vspace{-1cm}}

\begin{document}

\maketitle

\footnotetext[1]{
Nanyang Technological University, 50 Nanyang Avenue, Singapore 639798.
}

\footnotetext[2]{
Department of Mathematics, National University of Singapore, 10 Lower Kent Ridge Road, Singapore 119076.
}

\footnotetext[3]{
Institute for Functional Intelligent Materials, National University of Singapore, 10 Lower Kent Ridge Road, Singapore 119076.
}

\begin{abstract}
Motivated by the expressive power of neural ordinary differential equations in flow-based generative modeling, we study the approximation of probability measures by pushforwards generated by parameterized families of Lipschitz vector fields. For Wasserstein distances, we establish a quantitative relationship between finite-sample point interpolation and distributional approximation, and derive universality criteria based on $L^p$-approximation of transport maps. In particular, affine-invariant families containing a non-affine vector field are universal from compactly supported absolutely continuous base measures. For total variation, we show that coordinatewise dynamics are insufficient in dimensions $d\geq 2$ and establish sufficient conditions based on one-dimensional Sobolev approximation and shear-type mixing. We further examine the role of point interpolation, establishing when different regimes relate to one another and when approximate interpolation can be made exact.

\vspace{0.3cm}
\noindent \textbf{Keywords:} Flow-based generative models, neural ordinary differential equations, interpolation, universal approximation, Wasserstein and total variation distances.
\end{abstract}

\setcounter{tocdepth}{2}
\tableofcontents
\vspace{0.5cm}

\section{Introduction}\label{s:gm:intro}

Modern generative models transform randomness into complex data distributions. Transport-based models do so by evolving samples along a deterministic flow. Starting from a random variable $X_0\in\R^d$ with law $\mu_{\mathrm B}$, one solves
\begin{equation}\label{eq:ode}
    \dot X_t = f_{\theta(t)}(X_t), \qquad X_{|t=0} = X_0,
\end{equation}
for $t\in[0,T]$, and generates the sample $X_T=\Phi_T^\theta(X_0)$, where $\Phi_T^\theta$ is the flow map of \eqref{eq:ode} driven by the signal $\theta(\cdot)$. At the level of probability measures, the generated law is the pushforward $\mu_T=(\Phi_T^\theta)_\#\mu_{\mathrm B}$. Equivalently, $\mu_t$ is a weak solution of the continuity equation
\begin{equation}\label{eq:gm:continuity}
    \partial_t\mu_t+\operatorname{div}\left(f_{\theta(t)}\mu_t\right)=0,
    \qquad \mu_{|t=0}=\mu_{\mathrm B}.
\end{equation}
Training amounts to optimizing the signal $\theta(\cdot)$ using samples from a target measure $\mu_\star$ so that the generated law $\mu_T$ approximates $\mu_\star$.

This formulation naturally defines a control problem. Because the same signal acts on every initial state, prescribing the terminal positions of finitely many samples is a problem of simultaneous, or finite-ensemble, controllability, whereas prescribing the terminal law $\mu_T$ leads to distributional controllability of \eqref{eq:gm:continuity}. We study the interplay between these two levels when the velocity fields are drawn from a fixed family $\mathcal F$, and seek structural conditions under which the induced pushforwards can approximate compatible target measures.

Many generative models share this transport structure while differing in their parameterization and training objective. Examples include continuous normalizing flows trained by maximum likelihood~\citep{Chen2018Neural,Grathwohl2019Ffjord} or by flow matching~\citep{Lipman2023FlowMatching}. Score-based diffusion models use stochastic dynamics, but, under suitable regularity assumptions and with the exact score, their time-marginal laws also admit a deterministic probability-flow ODE representation~\citep{Song2021ScoreSDE}. The approximation power of controlled flows is thus relevant across several generative modeling paradigms.

{
Understanding their reachable space is essential not only for training generative models, but also for evaluating them, as different evaluation criteria probe distinct structural properties of the generated law. A standard benchmark, the \emph{Fr\'echet Inception Distance (FID)}, compares Gaussian approximations of feature distributions via the squared $2$-Wasserstein distance~\citep{Heusel2017FID}, and is therefore blind to higher-order differences beyond means and covariances. \emph{Precision--recall} criteria~\citep{Sajjadi2018PRGAN,Kynkaanniemi2019Improved} instead quantify the fraction of generated mass lying in the support of the target distribution (precision) and the fraction of target mass covered by the model (recall). In the formulation of~\citet{Sajjadi2018PRGAN}, the largest common value of precision and recall between two laws $\mu$ and $\nu$ equals their shared mass $1-|\mu-\nu|_{\TV}$. 


These connections motivate the two discrepancies studied in this work. Wasserstein-$p$ distances ($W_p$) measure the spatial cost of transporting mass, whereas total variation ($\TV$) controls the probability error uniformly across all measurable events:
\begin{equation}\label{eq:tv.intro}
 |\mu(E)-\nu(E)|\le |\mu-\nu|_{\TV}
    \qquad\text{for every Borel subset }E\subset\R^d.   
\end{equation}
In particular, assigning no mass to a target region of mass $m$ incurs a $\TV$ error of at least $m$, even when the generated mass lies arbitrarily close to it. Generating spatially accurate samples ($W_p$) and accurately approximating event probabilities ($\TV$) thus impose fundamentally different requirements.

Event probabilities are particularly relevant when generative models represent random fields or PDE solutions, as in function-space flow matching~\citep{Kerrigan2024Functional}. Consider, for instance, steady conduction $-\nabla\cdot(\kappa\nabla u)=f$ with a random source $f$ or random conductivity $\kappa$. In reliability analysis, one seeks the probability that a critical observable, such as the peak temperature~\citep{WanZhou2018Elliptic} or the boundary flux~\citep{Wagner2021RareEvents}, exceeds a fixed safety threshold $\tau$. For a discretized solution in $\R^d$ and an observable $q:\R^d\to\R$, this corresponds to the exceedance event $\{q>\tau\}\subset\R^d$. By \eqref{eq:tv.intro}, $\TV$ controls the error in the failure probability uniformly over $\tau$, without assumptions on the concentration of $q$ near the threshold. While $\TV$ bounds absolute rather than relative errors, so accurately resolving rare failure events still requires a correspondingly fine tolerance, it remains the natural metric to prevent structural underestimation of critical thresholds.


Returning to our control perspective, a natural question is how point interpolation and map approximation~\citep{Cheng2025Interpolation} relate to distributional approximation under $W_p$ and $\TV$. Passing from points or maps to measures modifies the problem in two opposing ways. First, point assignments are relaxed: whenever $\mathsf P:\R^d\to\R^d$ preserves $\mu_{\mathrm B}$, then
\begin{equation*}
    (\Phi\circ\mathsf P)_\#\mu_{\mathrm B}=\Phi_\#\mu_{\mathrm B},
\end{equation*}
so distinct maps (or permutations of an empirical sample) produce the exact same terminal law. Conversely, transporting continuous mass demands stronger regularity: since finite sets have Lebesgue measure zero, interpolating arbitrary finite samples does not control how the flow transports the bulk of the mass between them, and thus need not imply $W_p$ universality, while matching densities in $\TV$ further requires controlling local volume distortion.


Our central question is therefore:
\par\smallskip
\noindent\emph{Under what conditions do the point interpolation or functional approximation properties of a control family imply universality of the induced pushforward measures in $W_p$ and $\TV$?}
\par\smallskip

\par\smallskip

\noindent\emph{Related work. Transport by flows.}
The transport of probability densities by flows has a classical foundation in the theory of volume forms. Moser's theorem links smooth positive volume forms of equal mass on a closed connected manifold via a time-dependent flow~\citep{Moser1965Volume}, while the Dacorogna--Moser construction treats bounded domains with regularity estimates and boundary-fixing maps~\citep{DacorognaMoser1990Jacobian}, later refined to control the support near the boundary~\citep{Teixeira2017Support}. In optimal transport, the quadratic optimal map is a convex gradient~\citep{Brenier1991Polar} with well-known regularity theory~\citep{Caffarelli1992Regularity}, and is dynamically recast via the continuity equation by~\citet{BenamouBrenier2000Computational}. In all these classical results, the vector fields are unconstrained.

\noindent\emph{Restricted vector fields.}
A nonholonomic version of Moser's theorem establishes transport under time-dependent vector fields constrained to a bracket-generating distribution~\citep{KhesinLee2009Nonholonomic}. For finite compositions of autonomous flows, \citet{AgrachevCaponigro2009Diffeomorphisms} show that a bracket-generating family closed under multiplication by smooth functions generates the identity component of the diffeomorphism group on closed manifolds; see also~\citet{Caponigro2011Diffeomorphisms} and~\citet{GrongSchmeding2025Boundary} for manifolds with boundary. Under suitable controllability and regularity assumptions, \citet{CaponigroVicari2025Transport} further establish the exact realization of quadratic optimal transport maps on smooth bounded uniformly convex domains. Transport by a single autonomous vector field and the approximation power gained through compositions are studied by~\citet{DeNittiFernandezReal2025Autonomous} and~\citet{RouhvarziKratsios2025Incremental}.

\noindent\emph{Ensemble controllability.}
Steering multiple initial states simultaneously by a common control is studied in ensemble control, with Lie-algebraic criteria available for finite and continuous ensembles~\citep{AgrachevBaryshnikovSarychev2016,AgrachevSarychev2020Ensembles}. Our universal interpolation property is a finite-ensemble requirement. \citet{CuchieroLarssonTeichmann2020} establish exact interpolation using five suitable vector fields, as well as interpolation for almost every choice of field parameters and input configurations under suitable analyticity assumptions. Related neural ODE results study the required complexity~\citep{RuizBalet2023Neural, AlvarezLopez2024Interplay,AlvarezLopez2026Constructive} or links between interpolation and function approximation~\citep{Cheng2025Interpolation}. At the distributional level, controllability of the Liouville equation and realization of diffeomorphisms through controlled flows (including feedback controls) are studied by~\citet{Brockett2012Liouville} and~\citet{Raginsky2024Liouville}, while \citet{DuprezMoranceyRossi2019Continuity} obtain approximate and exact transport using localized Lipschitz vector fields.
}

\noindent\emph{Generative modeling.}
Linear-control systems can approximate diffeomorphisms isotopic to the identity~\citep{Scagliotti2023Diffeomorphisms} and optimal transport maps~\citep{ScagliottiFarinelli2025Normalizing}. For neural ODEs, distributional approximation has been established in Wasserstein distance~\citep{RuizBalet2023Neural,Elamvazhuthi2022Trajectory}, total variation~\citep{RuizBalet2024Control}, and relative entropy~\citep{AlvarezLopez2025Constructive}, as well as simultaneously for maps and pushforwards~\citep{Geshkovski2026Constructive}; statistical guarantees are developed in~\citet{Marzouk2024Distribution}. Closer to our perspective, \citet{Ishikawa2023Universal} establish general criteria linking functional approximation to distributional universality, including Sobolev bounds for $\TV$, with applications to neural ODEs.
{

\par\smallskip
\noindent\emph{Our contributions.}
Existing universality results rely on Lie-algebraic or architecture-specific constructions, as well as functional approximation hypotheses~\citep{Ishikawa2023Universal}. Meanwhile, general principles linking finite-sample interpolation to approximation have been developed only at the map level~\citep{Cheng2025Interpolation}. This motivates identifying what structural properties a general architecture must satisfy for point- or map-level expressivity to lift to probability laws. In this work, we characterize when these properties imply distributional universality, when they fail, and which additional mechanisms yield $W_p$- and $\TV$-universality.

}


\par\smallskip
\noindent\emph{Notation.}
We use $[n]\coloneqq\{1,\dots,n\}$ and let $S_n$ denote the set of permutations of $[n]$. For an open set $\Omega\subset\R^d$, we use $C^k(\Omega;\R^q)$, $C^{0,1}(\Omega;\R^q)$, $L^p(\Omega;\R^q)$, and $W^{m,p}(\Omega;\R^q)$ for the spaces of $k$-times continuously differentiable, globally Lipschitz (with Lipschitz constant $\Lip(\cdot)$), $p$-integrable, and Sobolev maps, respectively. For a function space $\mathcal X$, the closure $\overline{\mathcal X}^{\,\tau}$ is taken with respect to the topology $\tau$. The subscripts $_c$ and $_{\mathrm{loc}}$ denote compact support and the corresponding local space, respectively. For $d=1$, the subscript $_{\mathrm{nd}}$ denotes the subspace of nondecreasing functions (or those admitting a nondecreasing representative). The $i$-th coordinate projection is $\pi^i(x)=x^{(i)}$.

We denote by $\mathcal B(\R^d)$ the Borel $\sigma$-algebra on $\R^d$, by $\mathcal P(\R^d)$ the space of Borel probability measures, and by $\mathcal L^d$ the $d$-dimensional Lebesgue measure, with $|\Omega|\coloneqq\mathcal L^d(\Omega)$ for $\Omega\in\mathcal B(\R^d)$. We set $\Pac(\R^d)\coloneqq\{\mu\in\mathcal P(\R^d):\mu\ll\mathcal L^d\}$ and denote by $\mathcal P_p(\R^d)$ the set of probability measures with finite $p$-th moment. Weak convergence is denoted by $\mu_n\rightharpoonup\mu$. Finally, for $0<|\Omega|<\infty$, we write $\mathcal U(\Omega)\coloneqq \mathbf 1_\Omega/|\Omega|\,\mathcal L^d$ for the uniform probability measure on $\Omega$.

\section{Setup and main results}\label{s:gm:setup}

Fix $d\ge 1$ and a parameter space $\Theta$. We formalize the generative model by considering a \emph{control family} of globally Lipschitz vector fields,
\begin{equation}\label{eq:gm:control_family}
\mathcal{F} \coloneqq \{f_\theta:\R^d\to\R^d : \theta\in\Theta\} \subset {C}^{0,1}(\R^d;\R^d).
\end{equation}
A relevant example is the family of one-layer neural ODEs, defined for a componentwise activation function $\upsigma\in{C}^{0,1}(\R)$ by
\begin{equation}\label{eq:gm:node}
f_\theta(x)=W\,\upsigma(Ax+b),
\qquad
\theta=(W,A,b)\in\R^{d\times d}\times\R^{d\times d}\times\R^d.
\end{equation}

We constrain the control input $t \mapsto \theta(t)$ to be piecewise constant with finitely many switches, so that
\begin{equation}\label{eq:gm:ode}
\dot x(t)=f_{\theta(t)}\left(x(t)\right),\qquad x(0)=x_0,
\end{equation}
evolves through a finite sequence of autonomous ODEs. Let $\Phi_t^f$ denote the unique flow map generated by a vector field $f \in {C}^{0,1}(\R^d;\R^d)$ for $t\ge0$. Then the flow map of \eqref{eq:gm:ode} at any time $T>0$ is a finite composition of these autonomous flows.

We define the sets of maps attainable at time $T>0$ and in finite time as
\begin{equation*}
\mathcal A_{\mathcal{F},T}
\coloneqq
\left\{
\Phi_{t_m}^{f_m} \circ \cdots \circ \Phi_{t_1}^{f_1} : m \ge 1, \, f_i \in \mathcal{F}, \, t_i > 0, \, \sum_{i=1}^m t_i = T
\right\},\qquad
\mathcal A_{\mathcal{F}}
\coloneqq
\bigcup_{T>0}\mathcal A_{\mathcal{F},T}.
\end{equation*}

As outlined in Section~\ref{s:gm:intro}, these dynamics naturally lift to probability measures via the pushforward. For a base measure $\mu_{\mathrm B}\in\mathcal P(\R^d)$, the corresponding sets of attainable measures at time $T>0$ and in finite time are
\begin{equation*}
\mathcal A_{\mathcal{F},T}(\mu_{\mathrm B})
\coloneqq
\left\{\Phi_\#\mu_{\mathrm B}\in\mathcal P(\R^d): \Phi\in \mathcal A_{\mathcal{F},T}\right\},
\qquad
\mathcal A_{\mathcal{F}}(\mu_{\mathrm B})
\coloneqq
\bigcup_{T>0}\mathcal A_{\mathcal{F},T}(\mu_{\mathrm B}).
\end{equation*}
In particular, when $\mathcal F={C}^{0,1}(\R^d;\R^d)$, the set $\mathcal A_{{C}^{0,1}}(\mu_{\mathrm B})$ comprises all measures reachable from $\mu_{\mathrm B}$ through finite concatenations of flows generated by globally Lipschitz vector fields.

\begin{definition}[Scaling and Affine Invariance]\label{def:gm:scaling_affine}
A control family $\mathcal{F}$ is said to be:
\begin{itemize}[leftmargin=*]
    \item \emph{scaling invariant} if $f\in\mathcal{F} \Longrightarrow \lambda f\in\mathcal{F}$ for all $\lambda>0$;
    \item \emph{affine invariant} if $f\in\mathcal{F} \Longrightarrow \left(x\mapsto W f(Ax+b)\right)\in\mathcal{F}$ for all $W,A \in \R^{d\times d}$, $b \in \R^d$.
\end{itemize}
\end{definition}
Note that affine invariance implies scaling invariance (take $W=\lambda\Id$, $A=\Id$, $b=0$).

\begin{remark}\label{rem:gm:Lip_normalization}
When $\mathcal{F}$ is scaling invariant and the terminal time is free, restricting the family to vector fields with a bounded Lipschitz constant does not alter the attainable sets. More precisely, setting $\mathcal{F}_1 \coloneqq \{f \in \mathcal{F} : \Lip(f) \le 1\}$, we have
\begin{equation*}
\mathcal A_{\mathcal{F}_1} = \mathcal A_{\mathcal{F}}
\qquad\text{and}\qquad
\mathcal A_{\mathcal{F}_1}(\mu_{\mathrm B}) = \mathcal A_{\mathcal{F}}(\mu_{\mathrm B})
\quad\text{for every }\mu_{\mathrm B}\in\mathcal P(\R^d).
\end{equation*}
\end{remark}

We now recall the following definition from~\citet{Cheng2025Interpolation}.

\begin{definition}[Universal Interpolation Property]\label{def:interp:UIP}
For $d\ge 2$, the system driven by a control family $\mathcal{F}$ possesses the \emph{approximate universal interpolation property (approximate \UIP)} if, for every $N\ge 1$, $\varepsilon>0$, and pairs $(x_1,y_1),\dots,(x_N,y_N) \in \R^d\times\R^d$ with $x_i \neq x_j$ and $y_i \neq y_j$ whenever $i \neq j$, there exists $\Phi\in\mathcal A_{\mathcal{F}}$ such that
\begin{equation}\label{eq:matching}
\|\Phi(x_i)-y_i\| < \varepsilon \qquad \text{for all } i\in[N].    
\end{equation}
If $\Phi$ can always be chosen such that $\Phi(x_i)=y_i$ for all $i\in[N]$, the system possesses the \emph{(exact) universal interpolation property (\UIP)}. When $d=1$, the same definitions apply after relabeling the pairs so that $x_1<\cdots<x_N$ and restricting to targets satisfying $y_1<\cdots<y_N$.
\end{definition}

Alongside point interpolation, we consider the ability to approximate functions.

\begin{definition}[Universal Approximation Property]
Let $\mathcal H_{\mathrm{loc}}(\R^d;\R^d)$ be a local function space (e.g., $L^p_{\mathrm{loc}}$ or $W^{1,1}_{\mathrm{loc}}$) equipped with a family of seminorms $\|\cdot\|_{\mathcal H(K)}$ defined on restrictions to subsets $K \Subset \R^d$. The system driven by a control family $\mathcal{F}$ possesses the \emph{$\mathcal H$-universal approximation property ($\mathcal H$-\UAP)} if, for every $h\in \mathcal H_{\mathrm{loc}}(\R^d;\R^d)$, every $K\Subset\R^d$, and every $\varepsilon>0$, there exists $\Phi\in\mathcal A_{\mathcal{F}}$ such that
\begin{equation*}
\|(\Phi-h)|_K\|_{\mathcal H(K)}<\varepsilon.
\end{equation*}
\end{definition}

A natural starting point is Theorem~2.6 of~\citet{Cheng2025Interpolation}, which states that an affine-invariant control family possesses the exact \UIP\ if and only if it contains at least one vector field that is non-affine in the state variable (referred to as nonlinear by~\citet{Cheng2025Interpolation}). The central goal of this work is to determine how, and to what extent, this property lifts to probability measures.

This transition reveals an immediate structural obstruction: because every $\Phi \in \mathcal{A}_{\mathcal{F}}$ is a bi-Lipschitz homeomorphism, atomic parts remain atomic and Lebesgue null sets are preserved. Consequently, exact transport between arbitrary probability measures is generally impossible within this class of flows, and even approximate transport depends heavily on the underlying topology. 

Any notion of universality at the measure level must therefore be tied to the chosen discrepancy. To make this precise, we fix a reference class of probability measures $\mathcal X\subseteq~\mathcal P(\R^d)$.

\begin{definition}[Discrepancy]\label{def:discrepancy}
A map $\mathsf D:\mathcal X\times\mathcal X\to[0,+\infty]$ is called a \emph{discrepancy} if:
\begin{enumerate}
    \item $\mathsf{D}(\mu,\nu)=0$ if and only if $\mu=\nu$;
    \item $\mathsf{D}(\mu_n,\mu) \to 0$ implies $\mu_n \rightharpoonup \mu$ (i.e., $\int \varphi \diff\mu_n \to \int \varphi \diff\mu$ for all $\varphi \in C_b(\R^d)$).
\end{enumerate}
\end{definition}


For $W_p$ we take $\mathcal X=\mathcal P_p(\R^d)$, whereas for $\TV$ we take $\mathcal X=\mathcal P(\R^d)$. To characterize feasible targets, we introduce the following notion.

\begin{definition}[Measure Compatibility]\label{def:gm:D-compatible}
Given a discrepancy $\mathsf{D}$ in the sense of Definition~\ref{def:discrepancy}, we say that $(\mu_{\mathrm B},\mu_\star)\in\mathcal X\times\mathcal X$ is \emph{$\mathsf{D}$-compatible} if there exist $\Psi_n\in\mathcal A_{{C}^{0,1}}$ such that $(\Psi_n)_\#\mu_{\mathrm B}\in\mathcal X$ and $\mathsf{D}\left((\Psi_n)_\#\mu_{\mathrm B},\mu_\star\right)\to 0$ as $n\to\infty$.
\end{definition}


For the two discrepancies we consider, $\mathsf{D}$-compatibility admits concrete characterizations: every target in $\mathcal P_p(\R^d)$ is $W_p$-compatible with any atomless base measure in $\mathcal P_p(\R^d)$ (Proposition~\ref{prop:gm:Wp_compat}), whereas for any base measure in $\Pac(\R^d)$, $\TV$-compatibility simply amounts to absolute continuity of the target (Proposition~\ref{prop:gm:TV_compat_Pac}). 

With this framework in place, we define measure-level universality.

\begin{definition}[Distributional UAP]\label{def:gm:D-UAP}
The system driven by a control family $\mathcal{F}$ possesses the \emph{$\mathsf{D}$-universal approximation property ($\mathsf{D}$-\UAP)} if, for every $\mathsf{D}$-compatible pair $(\mu_{\mathrm B},\mu_\star)$ and every $\varepsilon>0$, there exists $\nu\in\mathcal A_{\mathcal F}(\mu_{\mathrm B})\cap\mathcal X$ such that $\mathsf D(\nu,\mu_\star)<\varepsilon$.  

Given a subclass $\mathcal X_0\subseteq\mathcal X$, we say that the system possesses the \emph{$\mathsf D$-\UAP\ on $\mathcal X_0$} if the above property holds for every $\mathsf D$-compatible pair $(\mu_{\mathrm B},\mu_\star)$ with $\mu_{\mathrm B}\in\mathcal X_0$ and $\mu_\star\in\mathcal X$.
\end{definition}

\begin{figure}[tbp]
    \centering
    \resizebox{\linewidth}{!}{%
    \begingroup
\providecolor{condPink}{HTML}{BE4082}
\providecolor{condBlue}{HTML}{1979A9}
\providecolor{condGreen}{HTML}{2E8B57}
\providecolor{figref}{HTML}{526078}
\providecolor{nonimplRed}{HTML}{C43C35}
\usetikzlibrary{calc,arrows.meta}

\newcommand{\impDef}[1]{{\normalfont\unboldmath\tiny\textcolor{figref}{%
  \hyperref[#1]{Def.~\ref*{#1}}}}}
\newcommand{\impThm}[1]{\hyperref[#1]{Thm.~\ref*{#1}}}
\newcommand{\impProp}[1]{\hyperref[#1]{Prop.~\ref*{#1}}}
\newcommand{\impCor}[1]{\hyperref[#1]{Cor.~\ref*{#1}}}
\newcommand{\impEq}[1]{\hyperref[#1]{(\ref*{#1})}}
\newcommand{\impSub}[1]{{\normalfont\unboldmath\tiny #1}}

\begin{tikzpicture}[
  x=1cm,y=1cm,
  box/.style={
    draw=black!30,
    line width=.75pt,
    rounded corners=2pt,
    fill=white,
    align=center,
    font=\scriptsize,
    text=black!62,
    text width=1.9cm,
    minimum height=.7cm,
    inner xsep=3pt,
    inner ysep=3pt,
    outer sep=1pt
  },
  key/.style={
    box,
    draw=figref!85!black,
    line width=1.2pt,
    fill=white,
    text=black,
    font=\scriptsize\bfseries\boldmath,
    text width=2.35cm,
    minimum height=.92cm,
    inner xsep=4pt,
    inner ysep=4pt
  },
  hyp/.style={
    box,
    draw=black!45,
    rounded corners=0pt,
    fill=black!5,
    text=black!80,
    font=\scriptsize\itshape,
    text width=1.32cm
  },
  tv/.style={draw=condGreen},
  interp/.style={draw=condPink},
  wasser/.style={draw=condBlue},
  impl/.style={
    -{Stealth[length=1.9mm]},
    line width=.55pt,
    draw=black!70
  },
  faint/.style={
    draw=black!32,
    line width=.45pt
  },
  equiv/.style={
    {Stealth[length=1.9mm]}-{Stealth[length=1.9mm]},
    line width=.55pt,
    draw=black!70
  },
  mainequiv/.style={
    equiv,
    line width=1pt,
    draw=black
  },
  cond/.style={
    impl,
    dash pattern=on 2.5pt off 2pt
  },
  nonimpl/.style={
    -{Stealth[length=2mm]},
    line width=.75pt,
    solid,
    draw=nonimplRed
  },
  lbl/.style={
    font=\tiny,
    text=figref,
    fill=white,
    inner sep=2pt,
    align=center
  },
  cross/.style={
    fill=white,
    inner sep=.4pt,
    text=nonimplRed,
    font=\Large\bfseries\boldmath
  }
]

\def\rTV{3.30}
\def\rI{0}
\def\rW{-3.10}


\node[box,tv] (tvemp) at (0,\rTV)
  {TV-UAP\\\impSub{on $\mathcal P_{\mathrm{emp}}(\R^d)$}};
\node[box,tv] (tvuap) at (3.3,\rTV)
  {TV-UAP\\\impDef{def:gm:D-UAP}};
\node[key,tv,fill=condGreen!8] (tvac) at (6.6,\rTV)
  {TV-UAP\\\impSub{on $\Pac(\R^d)$}};

\node[hyp] (aff) at (10.5,\rTV)
  {Affine invariance\\\impDef{def:gm:scaling_affine}};

\node[box,interp] (epiuip) at (0,\rI)
  {Exact $\pi$-UIP\\\impDef{def:gm:permUIP}};
\node[key,interp,fill=condPink!8] (euip) at (3.3,\rI)
  {Exact UIP\\\impDef{def:interp:UIP}};
\node[box,interp] (auip) at (6.6,\rI)
  {Approximate UIP\\\impDef{def:interp:UIP}};
\node[key,interp,fill=condPink!8] (apiuip) at (10.5,\rI)
  {Approximate $\pi$-UIP\\\impDef{def:gm:permUIP}};
\node[key,interp,fill=condPink!8] (lpiuip) at (13.8,\rI)
  {$\Lambda$-approximate $\pi$-UIP\\
   \impDef{def:gm:Lip_piUIP}};

\node[box,wasser] (wpac) at (3.3,\rW)
  {$W_p$-UAP\\\impSub{on $\Pac(\R^d)$}};
\node[box,wasser] (wpuap) at (6.6,\rW)
  {$W_p$-UAP\\\impDef{def:gm:D-UAP}};
\node[key,wasser,fill=condBlue!8] (wpemp) at (10.5,\rW)
  {$W_p$-UAP\\
   \impSub{on $\mathcal P_{\mathrm{emp}}(\R^d)$}};
\node[key,wasser,fill=condBlue!8] (lwp) at (13.8,\rW)
  {$\Lambda$-$W_p$-UAP\\\impDef{def:gm:Lip_WpUAP}};


\draw[impl] (tvuap.east) -- (tvac.west);
\draw[impl,faint] (tvuap.west) -- (tvemp.east);
\draw[impl] (euip.west) -- (epiuip.east);
\draw[impl] (euip.east) -- (auip.west);

\draw[impl,rounded corners=3pt]
  ([xshift=6mm]epiuip.north)
  -- (.60,1.05) -- (9.85,1.05)
  -- ([xshift=-6.5mm]apiuip.north);

\draw[equiv] (auip.east) --
  node[lbl,above=2pt,fill=none,inner sep=1pt]
    {\impProp{prop:approx_labeled_unlabeled}}
  (apiuip.west);

\draw[impl] (lpiuip.west) -- (apiuip.east);
\draw[impl,faint] (wpuap.west) -- (wpac.east);
\draw[impl] (wpuap.east) -- (wpemp.west);
\draw[impl,rounded corners=3pt]
  (lwp.south) -- (13.8,-4.10)
  -- (6.6,-4.10) -- (wpuap.south);

\draw[cond]
  (aff.west) --
  node[lbl,above=2pt] {\impCor{cor:gm:gen_TV}}
  (tvac.east);

\draw[cond,rounded corners=3pt]
  (auip.south) -- (6.6,-.92)
  -- node[lbl,below=2pt]
       {\impProp{prop:approx_exact_UIP}}
     (3.85,-.92)
  -- ([xshift=5.5mm]euip.south);


\draw[equiv,faint]
  (tvemp.south) --
  node[lbl,right=1pt]
    {\impEq{eq:gm:TV_empirical_matching}}
  (epiuip.north);

\draw[mainequiv]
  (apiuip.south) --
  node[lbl,right=1pt]
    {\impThm{thm:gm:Wp_permUIP}}
  (wpemp.north);

\draw[mainequiv]
  (lpiuip.south) --
  node[lbl,right=1pt]
    {\impThm{thm:gm:Lip_WpUAP_equiv}}
  (lwp.north);


\draw[nonimpl,rounded corners=3pt,
      preaction={draw=white,line width=3pt,solid,-}]
  ([xshift=-6.5mm]euip.north)
  -- (2.65,2.15)
  -- node[pos=.3485,cross] {$\times$}
     node[pos=.3485,lbl,below=7pt]
       {\impProp{prop:uip_tv_independent}\\[-1pt] $d\ge2$}
     (5.95,2.15)
  -- ([xshift=-6.5mm]tvac.south);

\draw[nonimpl,rounded corners=3pt]
  ([xshift=6.5mm]tvac.south)
  -- (7.25,1.65)
  -- node[midway,cross] {$\times$}
     node[midway,lbl,above=7pt]
       {\impProp{prop:uip_tv_independent}\\[-1pt] $d\ge2$}
     (11.15,1.65)
  -- ([xshift=6.5mm]apiuip.north);

\draw[nonimpl,rounded corners=3pt]
  ([xshift=6.5mm]wpac.north)
  -- (3.95,-1.75)
  -- node[pos=.52,cross] {$\times$}
     node[pos=.52,lbl,below=7pt]
       {\impProp{prop:gm:piUIP_not_WpUAP}\\[-1pt] $d\ge2$}
     (9.85,-1.75)
  -- ([xshift=-6.5mm]apiuip.south);

\draw[nonimpl]
  ([xshift=-7mm]euip.south) --
  node[cross] {$\times$}
  node[lbl,left=8pt]
    {\impProp{prop:gm:piUIP_not_WpUAP}\\[-1pt] $d\ge2$}
  ([xshift=-7mm]wpac.north);

\end{tikzpicture}
\endgroup%
}
    \caption{Logical implications among the main properties we study.
    \textcolor{condGreen}{Green}, \textcolor{condPink}{pink}, and \textcolor{condBlue}{blue} colors indicate $\TV$ approximation, point interpolation, and $W_p$ approximation, respectively, while highlighted boxes mark the properties featured in our main results.
    Solid arrows denote implications, double-headed arrows indicate equivalence, and \textcolor{red}{red} arrows with a cross indicate strict non-implications.
    Finally, dashed arrows represent conditional implications that hold under the assumptions stated in the referenced results.}
    \label{fig:implications}
\end{figure}
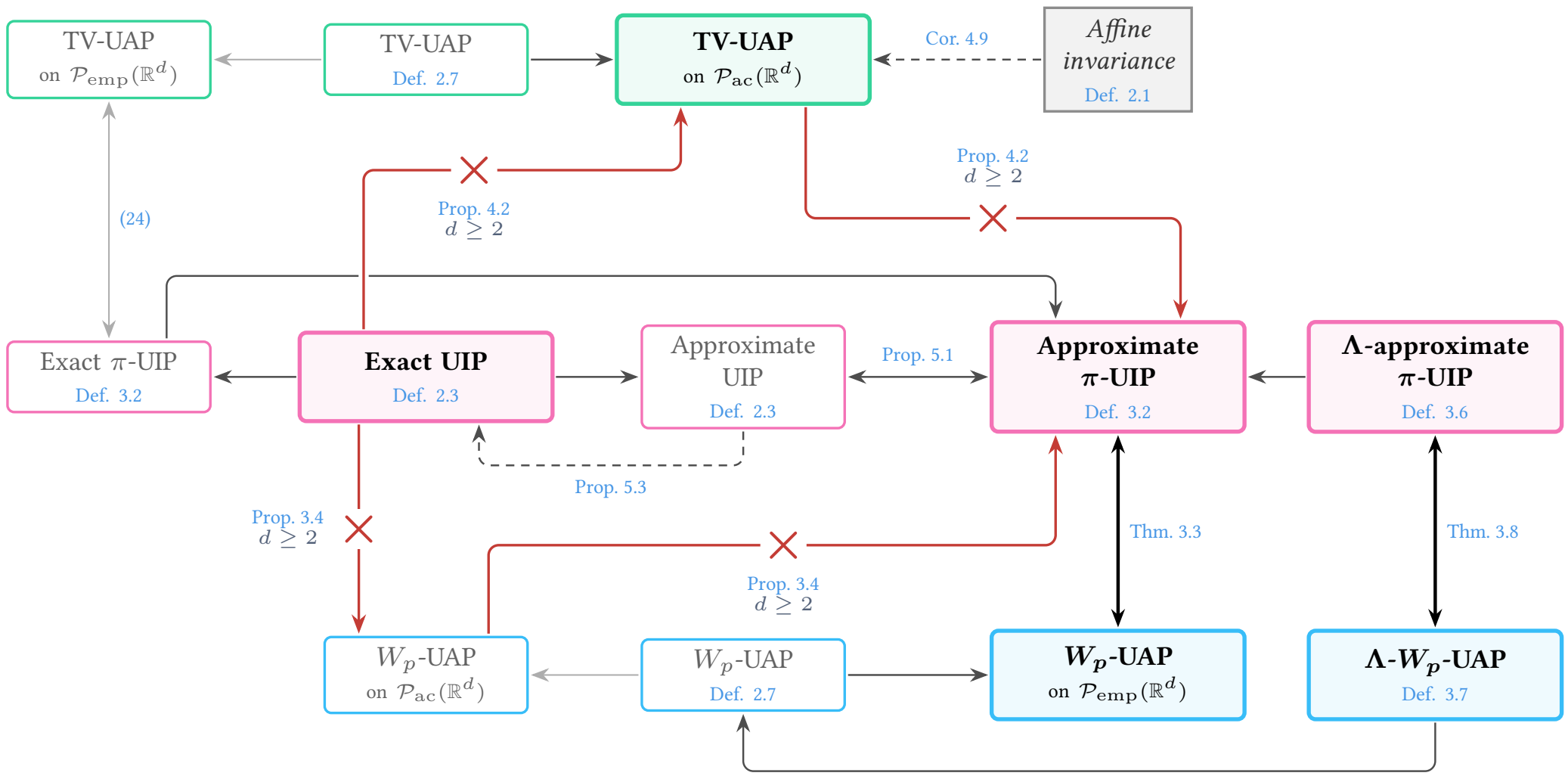

We now summarize the main results of this work according to the chosen discrepancy. Their logical relationships are illustrated in Figure~\ref{fig:implications}.

\medskip
\noindent\textbf{Wasserstein.}
In Section~\ref{sec:gm:wasserstein}, we establish the link between Wasserstein approximation and point interpolation. For uniform empirical measures, we prove that the $W_p$-\UAP\ is equivalent to the approximate \UIP\ up to permutation (Theorem~\ref{thm:gm:Wp_permUIP}). For general probability measures, however, interpolation alone is insufficient. We address this obstruction in two complementary ways: first, we prove a quantitative equivalence between Lipschitz-controlled interpolation and Wasserstein approximation (Theorem~\ref{thm:gm:Lip_WpUAP_equiv}); second, we derive sufficient criteria based on the $L^p$-approximation of transport maps, yielding Wasserstein universality for affine-invariant control families (Corollary~\ref{cor:gm:WpUAP_affine}).

\medskip
\noindent\textbf{Total variation.}
In Section~\ref{sec:gm:TV}, we develop sufficient mechanisms for total variation universality. In $d=1$, our criterion relies on approximating monotone rearrangements (Proposition~\ref{prop:d1interp}). In $d\ge2$, independent coordinatewise transport fails: the system must be capable of generating genuine coordinate coupling, for instance through shear-type flows. Under suitable structural assumptions, we establish an abstract criterion for the $\TV$-\UAP\ on $\Pac(\R^d)$ (Theorem~\ref{thm:TV-d}) and apply it to affine-invariant control families (Corollary~\ref{cor:gm:gen_TV}).

\medskip
We conclude in Section~\ref{s.discussion} with remarks on the scope and extensions of our results. All proofs and technical lemmas are collected in Appendices~\ref{app:preliminaries}--\ref{app:interpolation}.


\section{Wasserstein distance}\label{sec:gm:wasserstein}

For $p\in[1,\infty)$, the Wasserstein-$p$ distance between $\mu,\nu\in\mathcal P_p(\R^d)$ is defined by
\begin{equation}\label{eq:gm:Wp}
W_p(\mu,\nu)
\coloneqq
\inf_{\gamma\in\Gamma(\mu,\nu)}
\left(\int_{\R^d\times\R^d}\|x-y\|^p\diff\gamma(x,y)\right)^{1/p},
\end{equation}
where $\Gamma(\mu,\nu)$ is the set of couplings between $\mu$ and $\nu$, 
\begin{equation}\label{eq:set:couplings}
\Gamma(\mu,\nu)
\coloneqq
\left\{\gamma\in\mathcal P(\R^d\times\R^d):\ \gamma(\cdot\times\R^d)=\mu,\ \gamma(\R^d\times\cdot)=\nu\right\}.
\end{equation}
For later use, we denote the class of uniform empirical measures by
\begin{equation}\label{def.Pemp}
\mathcal P_{\mathrm{emp}}(\R^d)
\coloneqq
\left\{
\frac1N\sum_{i=1}^N \delta_{x_i}
\;:\;
N\ge1,\ x_i\neq x_j \text{ for } i\neq j
\right\}.
\end{equation}

We repeatedly use two standard properties of $W_p$:

\noindent\textbf{(i)} \emph{Lipschitz pushforwards.} For every $f\in{C}^{0,1}(\R^d;\R^d)$ and $\mu,\nu\in\mathcal P_p(\R^d)$,
\begin{equation}\label{eq:gm:Wp_Lip}
W_p(f_\#\mu,f_\#\nu)\le \Lip(f)\,W_p(\mu,\nu).
\end{equation}

\noindent\textbf{(ii)} \emph{Empirical matching.} Given $\mu_N=\frac1N\sum_{i=1}^N\delta_{x_i}$ and $\nu_N=\frac1N\sum_{i=1}^N\delta_{y_i}$ in $\mathcal P_{\mathrm{emp}}(\R^d)$,
\begin{equation}\label{eq:gm:Wp_empirical_assign}
W_p^p(\mu_N,\nu_N)
=
\min_{\pi\in S_N}\frac1N\sum_{i=1}^N\|x_i-y_{\pi(i)}\|^p.
\end{equation}

We begin by characterizing $W_p$-compatibility.

\begin{proposition}\label{prop:gm:Wp_compat}
Let $d\ge2$, $p\in[1,\infty)$, and $\mu_{\mathrm B},\mu_\star\in\mathcal P_p(\R^d)$. Then $(\mu_{\mathrm B},\mu_\star)$ is $W_p$-compatible if and only if there exists a map $g$ assigning each atom of $\mu_{\mathrm B}$ to an atom of $\mu_\star$ such that
\begin{equation*}
\sum_{\substack{x:\ \mu_{\mathrm B}(\{x\})>0\\ g(x)=y}}
\mu_{\mathrm B}(\{x\})
\le
\mu_\star(\{y\})
\qquad
\text{for every } y\in\R^d \text{ with } \mu_\star(\{y\})>0.
\end{equation*}
\end{proposition}


This condition reflects the fact that bi-Lipschitz homeomorphisms cannot split atoms: in the $W_p$-limit, each base atom must be transported into a target atom, and while distinct base atoms may merge, their combined mass cannot exceed that of the receiving target atom. In particular, when $\mu_{\mathrm B}$ is atomless (as is standard in generative modeling) the condition holds vacuously, so every target $\mu_\star\in\mathcal P_p(\R^d)$ is $W_p$-compatible with $\mu_{\mathrm B}$.

\subsection{Empirical approximation}\label{ss.emp.approx}

Fix $d\ge1$. We first consider base measures in $\mathcal P_{\mathrm{emp}}(\R^d)$. For two such measures with the same number of atoms, \eqref{eq:gm:Wp_empirical_assign} shows that the Wasserstein distance is obtained by minimizing over permutations. This naturally motivates the following unlabeled variant of the \UIP.

\begin{definition}[Unlabeled Universal Interpolation Property]\label{def:gm:permUIP}
For $d\ge1$, the system driven by a control family $\mathcal{F}$ possesses the \emph{approximate unlabeled-\UIP\ (approximate \uUIP)} if, for every $N\ge 1$, $\varepsilon>0$, and pairs $(x_1,y_1),\dots,(x_N,y_N)\in\R^d\times\R^d$ with $x_i\neq x_j$ and $y_i\neq y_j$ whenever $i\neq j$, there exist $\Phi\in\mathcal A_{\mathcal{F}}$ and $\pi\in S_N$ such that
\begin{equation}\label{eq:gm:permUIP}
\max_{i\in[N]}\|\Phi(x_i)-y_{\pi(i)}\| < \varepsilon.
\end{equation}
If $\Phi$ and $\pi$ can always be chosen such that $\Phi(x_i)=y_{\pi(i)}$ for all $i\in[N]$, the system possesses the \emph{(exact) unlabeled-\UIP\ (\uUIP)}.
\end{definition}

A closer comparison of the \UIP\ and \uUIP\ is deferred to Section~\ref{s.discussion.interpolation}; here, we show that the approximate \uUIP\ characterizes Wasserstein universality on $\mathcal P_{\mathrm{emp}}(\R^d)$.

\begin{theorem}\label{thm:gm:Wp_permUIP}
Let $\mathcal{F}$ be a Lipschitz control family on $\R^d$. The following are equivalent:
\begin{enumerate}[label=\textup{(\roman*)}, ref=\textup{(\roman*)}]
\item the associated system possesses the approximate \uUIP;
\item the associated system possesses the $W_p$-\UAP\ on $\mathcal P_{\mathrm{emp}}(\R^d)$ for all $p\in[1,\infty)$.
\end{enumerate}
\end{theorem}

{
To pass from empirical to general probability laws, recall that every $\mu\in\mathcal P_p(\R^d)$ admits a sequence $\mu_N\in\mathcal P_{\mathrm{emp}}(\R^d)$ with $W_p(\mu_N,\mu)\to0$. Propagating this convergence via transport maps $\Phi_N$, however, requires stability bounds such as uniform Lipschitz control on $\Phi_N$.

Without such control, point interpolation and $W_p$-universality from continuous base measures are mutually independent if $d\ge2$. Two geometric mechanisms---paralleling the function-level counterexamples in \cite[Thm.~2.9]{Cheng2025Interpolation}---explain this: volume preservation obstructs mass concentration without preventing point interpolation, whereas a common fixed point obstructs interpolation without preventing density approximation.
}

We illustrate these two mechanisms using the divergence-free family
\begin{equation}\label{eq:VrhoBdivfree}
\mathcal F_{\mathrm{div}}
\coloneqq
\left\{
x\mapsto w(a\cdot x+b)_+:
w,a\in\R^d,\ b\in\R,\ w\cdot a=0
\right\}
\end{equation}
and, for a globally Lipschitz activation $\upsigma$, the anchored family
\begin{equation}\label{eq:Fsigma0}
\mathcal F_{\upsigma,0}
\coloneqq
\left\{
x\mapsto w(\upsigma(a\cdot x+b)-\upsigma(b)):
w,a\in\R^d,\ b\in\R
\right\}.
\end{equation}


\begin{proposition}\label{prop:gm:piUIP_not_WpUAP}
Let $d\ge2$ and $p\in[1,\infty)$. Then:
\begin{enumerate}[
    label=\textup{(\roman*)},
    ref=\textup{(\roman*)},
    leftmargin=*
]
\item
The system associated with $\mathcal F_{\mathrm{div}}$ in \eqref{eq:VrhoBdivfree} possesses the exact \UIP, but fails the $W_p$-\UAP\ on $\Pac(\R^d)\cap\mathcal P_p(\R^d)$.

\item
For $\upsigma(s)=s_+$, the system associated with $\mathcal F_{\upsigma,0}$ in \eqref{eq:Fsigma0} possesses the $W_p$-\UAP\ on $\Pac(\R^d)\cap\mathcal P_p(\R^d)$, but fails the approximate \uUIP.

\end{enumerate}
\end{proposition}

\begin{figure}[tbp]
    \centering
    \includegraphics[width=0.9\linewidth]{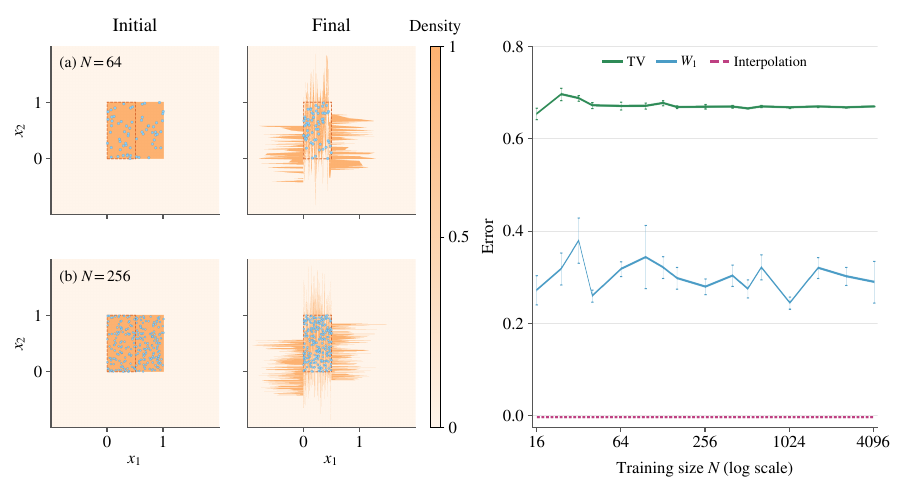}
    \caption{{Divergence-free interpolation without distributional approximation.
    (a,b) Initial and final densities ({\color{orange}orange}) for $N=64$ and $N=256$ sampled points ({\color{condBlue}blue}).
    The initial law is uniform on $[0,1]^2$; the target is uniform on $[0,1/2]\times[0,1]$, outlined by dashed lines.
    (c) Errors versus sample size $N$: sample points are matched, whereas the population errors in $W_1$ and $\TV$ remain bounded away from zero. Vertical bars indicate confidence intervals for the sampling-based $W_1$ and $\TV$ estimates, computed from five independent repetitions.}}
    \label{fig:divfree_numerics}
\end{figure}

Figure~\ref{fig:divfree_numerics} illustrates Proposition~\ref{prop:gm:piUIP_not_WpUAP}\textup{(i)} numerically: although the empirical samples are interpolated accurately, the population-level $W_1$ and $\TV$ errors remain bounded away from zero as the sample size $N$ increases. 
In fact, as formalized in the next remark, matching empirical samples in the presence of a population-level obstruction necessarily forces the Lipschitz constants of the interpolating flows to blow up as $N\to\infty$.

\begin{remark}[Lipschitz blow-up]\label{rem:gm:divfree_lip_blowup}
Suppose the system fails the $W_p$-\UAP. Then there exist compatible $\mu_{\mathrm B},\mu_\star\in\mathcal P_p(\R^d)$ and $\delta>0$ such that $W_p(\Phi_\#\mu_{\mathrm B},\mu_\star)\ge\delta$ for all $\Phi\in\mathcal A_{\mathcal{F}}$. Take sequences $\mu_{\mathrm B}^N,\mu_\star^N\in \mathcal P_{\mathrm{emp}}(\R^d)$ with equal numbers of atoms such that $W_p(\mu_{\mathrm B}^N,\mu_{\mathrm B})\to 0$ and $W_p(\mu_\star^N,\mu_\star)\to 0$. If the system possesses the approximate \uUIP, there exist $\Phi_N\in\mathcal A_{\mathcal{F}}$ such that $W_p((\Phi_N)_\#\mu_{\mathrm B}^N,\mu_\star^N) \to 0$. By the triangle inequality and \eqref{eq:gm:Wp_Lip},
\begin{equation*}
\delta
\le
W_p\left((\Phi_N)_\#\mu_{\mathrm B},\mu_\star\right)
\le
\Lip(\Phi_N)\,W_p(\mu_{\mathrm B},\mu_{\mathrm B}^N)
+
W_p\left((\Phi_N)_\#\mu_{\mathrm B}^N,\mu_\star^N\right)
+
W_p(\mu_\star^N,\mu_\star).
\end{equation*}
Since the last two terms vanish as $N\to\infty$, the first term must remain bounded away from zero; as $W_p(\mu_{\mathrm B},\mu_{\mathrm B}^N)\to0$, this forces $\Lip(\Phi_N)\to\infty$.
\end{remark}


{

Remark~\ref{rem:gm:divfree_lip_blowup} shows that passing from empirical interpolation to distributional approximation requires uniform Lipschitz control on the approximating flows. To formalize this quantitative link, we benchmark the controlled system against unrestricted Lipschitz flows of bounded complexity. For $r\ge0$, set
\begin{equation}\label{eq:gm:reference_budget}
\mathcal A_r\coloneqq
\left\{
\Phi_{t_m}^{f_m}\circ\cdots\circ\Phi_{t_1}^{f_1}
\in\mathcal A_{C^{0,1}}:
\sum_{j=1}^m t_j\Lip(f_j)\le r
\right\}.
\end{equation}
By Gr\"onwall's inequality, every $\Psi\in\mathcal A_r$ satisfies $\Lip(\Psi)\le e^r$, so $r$ quantifies the Lipschitz budget of the reference flow.

We capture the quantitative cost of approximation via a profile $\Lambda:[0,\infty)\to[1,\infty)$, where $\Lambda(r)$ bounds the Lipschitz constant of the admissible flows needed to approximate any reference map $\Psi\in\mathcal A_r$, uniformly in the number of points $N$ and accuracy $\varepsilon$.

\begin{definition}[$\Lambda$-Approximate \uUIP]
\label{def:gm:Lip_piUIP}
Let $\Lambda:[0,\infty)\to[1,\infty)$ be nondecreasing.
The system driven by a control family $\mathcal{F}$ possesses the \emph{$\Lambda$-approximate \uUIP}
if, for every $r\ge0$, $\Psi\in\mathcal A_r$, $N\ge1$,
distinct points $x_1,\dots,x_N\in\R^d$, and $\varepsilon>0$,
there exist $\Phi\in\mathcal A_{\mathcal F}$ and $\pi\in S_N$
such that
\begin{equation*}
\max_{i\in[N]}\|\Phi(x_i)-\Psi(x_{\pi(i)})\|<\varepsilon,
\qquad \Lip(\Phi)\le\Lambda(r).
\end{equation*}
\end{definition}

At the distributional level, the same profile $\Lambda(r)$ bounds the Lipschitz constant required to approximate the pushforward of any base measure under a reference flow $\Psi\in\mathcal A_r$.

\begin{definition}[$\Lambda$-$W_p$-\UAP]\label{def:gm:Lip_WpUAP}
Fix $p\in[1,\infty)$ and let
$\Lambda:[0,\infty)\to[1,\infty)$ be nondecreasing.
The system driven by a control family $\mathcal F$ possesses the \emph{$\Lambda$-$W_p$-\UAP}
if, for every $r\ge0$, $\Psi\in\mathcal A_r$,
$\mu_{\mathrm B}\in\mathcal P_p(\R^d)$, and $\varepsilon>0$,
there exists $\Phi\in\mathcal A_{\mathcal F}$ such that
\begin{equation*}
W_p(\Phi_\#\mu_{\mathrm B},\Psi_\#\mu_{\mathrm B})<\varepsilon,
\qquad \Lip(\Phi)\le\Lambda(r).
\end{equation*}
\end{definition}

The next theorem establishes that empirical and distributional approximation under a Lipschitz budget are equivalent and share the exact same profiles.

\begin{theorem}\label{thm:gm:Lip_WpUAP_equiv}
Fix $p\in[1,\infty)$ and let
$\Lambda:[0,\infty)\to[1,\infty)$ be nondecreasing.
A system driven by a control family possesses the
$\Lambda$-approximate \uUIP\ if and only if it possesses
the $\Lambda$-$W_p$-\UAP.
\end{theorem}

Any profile $\Lambda$ satisfying the conditions of Theorem~\ref{thm:gm:Lip_WpUAP_equiv} must obey the lower bound
\begin{equation}\label{eq:lowbdlip}
  \Lambda(r)\ge e^r.  
\end{equation}
Indeed, the linear map $\Psi(x)=(e^r x^{(1)},x^{(2)},\dots,x^{(d)})$ belongs to $\mathcal A_r$, and approximating its action on $\{0,e_1\}$ within error $\varepsilon$ requires $\Lip(\Phi)\ge e^r-2\varepsilon$ regardless of the permutation.
The following examples show that the optimal bound \eqref{eq:lowbdlip} can be attained, whereas qualitative universality alone need not admit any quantitative profile.

\begin{example}[Quantitative profiles]\label{ex:gm:basic_quant_profiles}
By Theorem~\ref{thm:gm:Lip_WpUAP_equiv}, a profile $\Lambda$ is valid whenever each $\Psi\in\mathcal A_r$ can be approximated locally uniformly by $\Phi\in\mathcal A_{\mathcal F}$ satisfying $\Lip(\Phi)\le\Lambda(r)$.

\begin{enumerate}[label=\textup{(\roman*)},ref=\textup{(\roman*)},leftmargin=*]
\item Fix a nonnegative radial $\rho\in C_c^\infty(\R^d)$ with
$\int_{\R^d}\rho(x)\diff x=1$. The control family
\begin{equation}\label{eq:gm:bump_family}
\mathcal F_\rho\coloneqq
\left\{x\mapsto w\rho\left(\frac{x-a}{s}\right):
w,a\in\R^d,\ s>0\right\}
\end{equation}
admits the profile $\Lambda_\eta(r)=(1+\eta)\vee e^r$ for all $\eta>0$, but admits no profile with $\Lambda(0)=1$.

\item 
Appending the linear contraction removes this threshold: $\mathcal F_\rho\cup\{x\mapsto-x\}$ admits the optimal profile $\Lambda(r)=e^r$. 

\item
For $d\ge2$, the divergence-free family $\mathcal F_{\mathrm{div}}$ possesses the \UIP, but admits no profile $\Lambda$.

\item
For $d\ge2$ and $\upsigma(s)=s_+$, the anchored family $\mathcal F_{\upsigma,0}$ of Proposition~\ref{prop:gm:piUIP_not_WpUAP}\textup{(ii)} possesses the $W_p$-\UAP\ on $\Pac(\R^d)\cap\mathcal P_p(\R^d)$, yet fails the $\Lambda$-$W_p$-\UAP\ for every profile $\Lambda$, even when $\mu_{\mathrm B}$ is restricted to such base measures.
\end{enumerate}

The proofs of (i)--(iv) are provided in Appendix~\ref{app:wasserstein}.
\end{example}

}

Finally, for a family with $\Lip(f)\le 1$ (such as $\mathcal{F}_1$ in Remark~\ref{rem:gm:Lip_normalization}), restricting the attainable set to a fixed time horizon $T>0$ enforces the uniform bound $\Lip(\Phi)\le e^T$ for all $\Phi\in\mathcal A_{\mathcal{F},T}$. Within $\mathcal A_{\mathcal{F},T}$, the Lipschitz blow-up of Remark~\ref{rem:gm:divfree_lip_blowup} is ruled out a priori, so distributional reachability of $\mu_\star$ from $\mu_{\mathrm B}$ within horizon $T$ is equivalent to uniform reachability from empirical approximations of $\mu_{\mathrm B}$ within the same horizon:

\begin{corollary}\label{cor:gm:uniform_time_Wp_closure}
Fix $d\ge1$, $p\in[1,\infty)$, $T>0$, and a control family $\mathcal{F}$ on $\R^d$ with $\Lip(f)\le 1$ for all $f\in\mathcal{F}$. For $\mu_{\mathrm B},\mu_\star\in\mathcal P_p(\R^d)$, the following are equivalent:
\begin{enumerate}[label=\textup{(\roman*)}, ref=\textup{(\roman*)}]
\item For every $\varepsilon>0$, there exists $\eta>0$ such that any $\nu\in\mathcal P_{\mathrm{emp}}(\R^d)$ with $W_p(\nu,\mu_{\mathrm B})<\eta$ admits a flow $\Phi\in\mathcal A_{\mathcal{F},T}$ satisfying $W_p(\Phi_\#\nu,\mu_\star)<\varepsilon$.
\item $\mu_\star \in \overline{\mathcal A_{\mathcal{F},T}(\mu_{\mathrm B})}^{\,W_p}$.
\end{enumerate}
\end{corollary}


\subsection{Approximation of transport maps}\label{ss:transp.map}

We now describe a second route to the $W_p$-\UAP, using functional approximation as a bridge: convergence of maps in $L^p(\mu_{\mathrm B};\R^d)$ implies convergence of their pushforwards in $W_p$. Indeed, for any $p\in[1,\infty)$, $\mu_{\mathrm B}\in\mathcal P_p(\R^d)$, and $\mathsf R,\mathsf S\in L^p(\mu_{\mathrm B};\R^d)$, the coupling $(\mathsf R,\mathsf S)_\#\mu_{\mathrm B}$ gives
\begin{equation}\label{eq:gm:Wasserstein-Lp}
W_p(\mathsf R_\#\mu_{\mathrm B},\mathsf S_\#\mu_{\mathrm B})
\le \|\mathsf R-\mathsf S\|_{L^p(\mu_{\mathrm B})}.
\end{equation}

When $\mu_{\mathrm B}$ is atomless, every $\nu\in\mathcal P_p(\R^d)$ can be written as $\nu=\mathsf S_\#\mu_{\mathrm B}$ for some map $\mathsf S\in L^p(\mu_{\mathrm B};\R^d)$. By~\eqref{eq:gm:Wasserstein-Lp}, density of $\mathcal A_{\mathcal{F}}$ in $L^p(\mu_{\mathrm B};\R^d)$ implies density of $\mathcal A_{\mathcal{F}}(\mu_{\mathrm B})$ in $(\mathcal P_p(\R^d),W_p)$. Our strategy for establishing this $L^p$-density depends strongly on the dimension.
\begin{itemize}[leftmargin=*]
    \item If $d=1$, the map $\mathsf S$ can be chosen nondecreasing, and the \UIP\ alone guarantees density of $\mathcal A_{\mathcal F}$ in the space ${C}^0_{\mathrm{nd}}(I)$ of nondecreasing continuous functions on any compact interval $I\subset\R$~\cite[Sec.~3.3.1]{Cheng2025Interpolation}:
\begin{equation}\label{eq:gm:1D_UIP_UAP}
\overline{\mathcal A_{\mathcal{F}}}^{\,L^\infty(I)} = {C}^0_{\mathrm{nd}}(I).
\end{equation}
    \item If $d\ge2$, the \UIP\ alone is insufficient, since a multidimensional flow can behave unpredictably between interpolation points. To recover full $L^p$-density, a standard strategy is to first compress the domain into localized clusters and then apply the \UIP\ to steer them to their prescribed targets~\citep{Li2023Deep,RuizBalet2023Neural}.
\end{itemize}

\begin{definition}[Diagonal Shrinking Property]\label{def:gm:shrinking}
The system driven by a control family $\mathcal{F}$ possesses the \emph{diagonal shrinking property (\DSP)} if, for every compact set $K\subset\R^d$ and every nondecreasing continuous function $h:\R \to \R$, we have
\begin{equation*}
x\mapsto
\left(h(x^{(1)}),\dots,h(x^{(d)})\right)
\in
\overline{\mathcal A_{\mathcal F}}^{\,L^\infty(K;\R^d)}.
\end{equation*}
\end{definition}

By \cite[Prop.~3.9]{Cheng2025Interpolation}, the \UIP\ combined with the \DSP\ implies the  $L^p$-\UAP\ for $d\ge2$. When $\mu_{\mathrm B}\in\Pac(\R^d)$ is compactly supported, this approximation yields global density in $L^p(\mu_{\mathrm B};\R^d)$, which lifts directly to the measure level via~\eqref{eq:gm:Wasserstein-Lp}.

\begin{proposition}\label{prop:gm:Cheng_lifting}
Let $d\ge2$, $p\in[1,\infty)$, and $\mathcal{F}$ be a control family on $\R^d$ whose associated system possesses both the \UIP\ and the \DSP. If $\mu_{\mathrm B}\in\Pac(\R^d)$ has compact support, then:
\begin{enumerate}[label=\textup{(\roman*)}, ref=\textup{(\roman*)}, leftmargin=*]
    \item\label{item:lifting_Lp} $\mathcal A_{\mathcal{F}}$ is dense in $L^p(\mu_{\mathrm B};\R^d)$;
    \item\label{item:lifting_Wp} $\mathcal A_{\mathcal{F}}(\mu_{\mathrm B})$ is dense in $(\mathcal{P}_p(\R^d),W_p)$ (so the system possesses the $W_p$-\UAP\ on the class of compactly supported measures in $\Pac(\R^d)$).
\end{enumerate}
\end{proposition}

\begin{remark}[Control of the tails]\label{rem:gm:local_to_global}
Compact support of $\mu_{\mathrm B}$ bypasses the need to control the flows at infinity. For a general base $\mu_{\mathrm B}\in\Pac(\R^d)\cap\mathcal P_p(\R^d)$ with unbounded support, one must split the $L^p$ error between the approximants $\Phi_n$ and the target map $\mathsf S$ over a ball $B_R(0)$ and its complement:
\begin{equation*}
\|\Phi_n-\mathsf S\|_{L^p(\mu_{\mathrm B})}^p
=
\int_{B_R(0)}\|\Phi_n-\mathsf S\|^p\diff\mu_{\mathrm B}
+
\int_{\R^d\setminus B_R(0)}\|\Phi_n-\mathsf S\|^p\diff\mu_{\mathrm B}.
\end{equation*}
For fixed $R$, the first integral vanishes as $n\to\infty$ by local approximation. Controlling the tail requires uniform bounds at infinity, such as uniform $\mu_{\mathrm B}$-integrability, to restore density.
\end{remark}


For affine-invariant control families such as \eqref{eq:gm:node}, the presence of a non-affine vector field guarantees both the \UIP\ and the \DSP~\citep{Cheng2025Interpolation}. Proposition~\ref{prop:gm:Cheng_lifting} therefore yields the following corollary.

\begin{corollary}\label{cor:gm:WpUAP_affine}
Let $d \ge 2$, and let $\mathcal{F} \subset {C}^{0,1}(\R^d;\R^d)$ be an affine-invariant control family containing at least one non-affine vector field. Then, for every $p\in[1,\infty)$ and every compactly supported $\mu_{\mathrm B}\in\Pac(\R^d)$, $\mathcal A_{\mathcal{F}}(\mu_{\mathrm B})$ is dense in $(\mathcal{P}_p(\R^d),W_p)$.
\end{corollary}

In summary, Wasserstein universality can be obtained either through quantitatively controlled interpolation or through $L^p$-approximation of transport maps. We now turn to total variation, for which these mechanisms no longer apply directly.

\section{Total variation distance}\label{sec:gm:TV}

For any pair of finite measures $\mu$ and $\nu$ on $\R^d$, the \emph{total variation} ($\TV$) distance is 
\begin{equation}\label{def:gm:tv}
|\mu-\nu|_{\TV}\coloneqq\sup_{E\in\mathcal{B}(\R^d)}|\mu(E)-\nu(E)|.
\end{equation}
For $\mu,\nu\in\Pac(\R^d)$ with densities $\rho_\mu$ and $\rho_\nu$, this reduces to\footnote{For distinct total masses, \eqref{eq:tvl1} holds upon adding $|\mu(\R^d)-\nu(\R^d)|/2$ to the right-hand side.}
\begin{equation}\label{eq:tvl1}
|\mu-\nu|_{\TV}=\tfrac12\|\rho_\mu-\rho_\nu\|_{L^1(\R^d)}.
\end{equation}

{

In generative modeling, a small $\TV$ error guarantees that the generated and target distributions assign similar probabilities to every measurable event, including regions corresponding to individual modes, regardless of their spatial separation.

Beyond this event-based view, $\TV$ admits a coupling characterization measuring the minimal mass that fails to be matched exactly: 
for any flow map $\Phi$,
\begin{equation}\label{eq:gm:TV_flow_matching}
|\Phi_\#\mu_{\mathrm B}-\mu_\star|_{\TV}
=
\min_{\gamma\in\Gamma(\mu_{\mathrm B},\mu_\star)}
\gamma\left(\{(x,y)\in\R^d\times\R^d:\Phi(x)\neq y\}\right),
\end{equation}
which follows by pushing forward couplings under $(x,y)\mapsto(\Phi(x),y)$. Under this perspective, an optimal coupling pairs initial points $x\sim\mu_{\mathrm B}$ with targets $y\sim\mu_\star$, so $|\Phi_\#\mu_{\mathrm B}-\mu_\star|_{\TV}<\varepsilon$ means that $\Phi(x)=y$ holds with probability at least $1-\varepsilon$.

For uniform empirical measures $\mu_N=\sum_{i=1}^N\delta_{x_i}/N$ and $\nu_N=\sum_{i=1}^N\delta_{y_i}/N$ with distinct atoms, this pairing can be chosen as a permutation, giving
\begin{equation}\label{eq:gm:TV_empirical_matching}
|\Phi_\#\mu_N-\nu_N|_{\TV}
=
\frac1N\min_{\pi\in S_N}
\#\left\{i\in[N]:\Phi(x_i)\neq y_{\pi(i)}\right\}.
\end{equation}
Compared to \eqref{eq:gm:Wp_empirical_assign}, the spatial transport cost is replaced by a discrete mismatch indicator. Any error below $1/N$ therefore forces exact matching, making the $\TV$-\UAP\ on $\mathcal P_{\mathrm{emp}}(\R^d)$ equivalent to the exact \uUIP.\footnote{The $\TV$-compatible targets of a uniform $N$-point base are exactly the uniform $N$-point measures with distinct atoms. All such targets are reachable by unrestricted Lipschitz flows, and this class is $\TV$-closed.}

Unlike in Wasserstein space, however, one cannot pass from $\mathcal P_{\mathrm{emp}}(\R^d)$ to continuous laws via empirical approximation: empirical and absolutely continuous measures are mutually singular, so their $\TV$ distance is identically equal to $1$. Since generative models aim to learn densities, we focus on $\Pac(\R^d)$, where compatibility admits a clean characterization.}

\begin{proposition}\label{prop:gm:TV_compat_Pac}
Let $d\ge1$, $\mu_{\mathrm B}\in\Pac(\R^d)$, and $\mu_\star\in\mathcal P(\R^d)$. Then $(\mu_{\mathrm B},\mu_\star)$ is $\TV$-compatible if and only if $\mu_\star\in\Pac(\R^d)$.
\end{proposition}

{
Although \eqref{eq:gm:TV_flow_matching} links $\TV$-approximation to exact matching on a set of large coupling mass, the $\TV$-\UAP\ on $\Pac(\R^d)$ is logically independent of finite-point interpolation. Indeed, the divergence-free and anchored families from Proposition~\ref{prop:gm:piUIP_not_WpUAP} also separate these properties in total variation, using a smooth activation in the anchored case.

\begin{proposition}\label{prop:uip_tv_independent}
Let $d\ge2$ and consider the families $\mathcal F_{\mathrm{div}}$ and $\mathcal F_{\upsigma,0}$ from \eqref{eq:VrhoBdivfree}--\eqref{eq:Fsigma0}. Then:
\begin{enumerate}[
    label=\textup{(\roman*)},
    ref=\textup{(\roman*)},
    leftmargin=*
]
\item
The system associated with $\mathcal F_{\mathrm{div}}$ possesses the \UIP, but fails the $\TV$-\UAP\ on $\Pac(\R^d)$.

\item
If $\upsigma\in C^{0,1}(\R)\cap C^1(\R)$ is non-affine, the system associated with $\mathcal F_{\upsigma,0}$ possesses the $\TV$-\UAP\ on $\Pac(\R^d)$, but fails the approximate \uUIP.
\end{enumerate}
\end{proposition}}

A further obstruction concerns the approximation of transport maps. While convergence of maps in $L^p(\mu_{\mathrm B})$ yields $W_p$-convergence of their pushforwards by \eqref{eq:gm:Wasserstein-Lp}, even uniform convergence of flows need not yield $\TV$-convergence, as the next example shows.

\begin{remark}[Uniform map convergence does not yield $\TV$ convergence]\label{rem:TV_fails_flows}
There exists a control family $\mathcal F\subset C^\infty(\R)$ with uniformly bounded Lipschitz constants and $\Phi_n\in\mathcal A_{\mathcal F,1}$ such that $\Phi_n\to\Id$ uniformly on $[0,1]$, hence $(\Phi_n)_\#\mu\to\mu$ in $W_p$ for $\mu\coloneqq\mathcal U([0,1])$, yet, for some constant
$c_0>0$ independent of $n$,
\begin{equation*}
\left|(\Phi_n)_\#\mu-\mu\right|_{\TV}
=c_0>0
\qquad\text{for every }n\ge 1.
\end{equation*}
The construction, detailed in Appendix~\ref{app.multi.d.tv}, relies on autonomous $f_n$ with uniformly bounded first derivatives while $\Lip(Df_n)\to\infty$; see also \cite[Sec.~1.2.1]{Geshkovski2026Constructive}.
\end{remark}

{
In general, neither $\TV$- nor $W_p$-convergence implies the other on $\Pac(\R^d)$ (see Example~\ref{ex:TV_vs_Wp_incomparable}). Passing from $\TV$ to $W_p$ only requires tail control: since $\TV$-convergence implies weak convergence, it yields $W_p$-convergence whenever $p$-th moments are uniformly integrable. Conversely, under uniform regularity bounds and a fixed time horizon, \(W_p\)-approximation can be upgraded to \(\TV\)-approximation, as shown by the following result.

\begin{proposition}\label{prop:gm:Wp_to_TV}
Let $p\in[1,\infty)$, let $\mu_{\mathrm B}\in\Pac(\R^d)$ be compactly supported, and let $\mu_\star\in\Pac(\R^d)\cap\mathcal P_p(\R^d)$. Assume that, for some $T>0$,
\begin{equation}\label{eq:fix.time.assump}
\mu_\star\in
\overline{\mathcal A_{\mathcal F,T}(\mu_{\mathrm B})}^{\,W_p}.
\end{equation}
Suppose moreover that $\mathcal F\subset W^{2,\infty}(\R^d;\R^d)$ and
\begin{equation}\label{eq:gm:C11_bound}
\|Df\|_{L^\infty(\R^d)}+\Lip(Df)\le1
\qquad\text{for every }f\in\mathcal F.
\end{equation}
Then there exists a sequence $\{\Phi_n\}\subset\mathcal A_{\mathcal F,T}$ such that $|(\Phi_n)_\#\mu_{\mathrm B}-\mu_\star|_{\TV}\to0$.
\end{proposition} }

We next establish \(\TV\)-universality through a different approach, based on stronger approximation properties of transport maps. In one dimension, we use \(W^{1,1}\)-approximation of monotone rearrangements; in higher dimensions, we combine coordinatewise transport with shear transformations to generate dependence between coordinates.

\subsection{One-dimensional case}\label{ss:1d}

For any $\mu_{\mathrm B},\mu_\star\in\Pac(\R)$, there is a canonical transport map between them given by the monotone rearrangement:
\begin{equation}\label{eq:gm:mon_rearrange}
\mathsf{M}_{\mathrm{B} \to \star}
\coloneqq F_{\mu_\star}^{-1}\circ F_{\mu_{\mathrm B}},
\qquad
F_\mu(x)\coloneqq \mu((-\infty,x]).
\end{equation}
Because one-dimensional flows are increasing homeomorphisms, $\mathsf{M}_{\mathrm{B} \to \star}$ serves as a natural reference for flow-based transport. However, approximating this map uniformly is not enough to ensure convergence of the pushforward densities (recall Example~\ref{ex:TV_vs_Wp_incomparable}). We use the stronger topology of $W^{1,1}_{\mathrm{loc}}$ to obtain a sufficient condition for $\TV$-universality.


\begin{proposition}\label{prop:d1interp}
Let $\mathcal{F}$ be a Lipschitz control family on $\R$ for which $\mathcal A_{\mathcal{F}}$ is dense in the set $W^{1,1}_{\mathrm{nd},\mathrm{loc}}(\R)$ of nondecreasing functions in $W^{1,1}_{\mathrm{loc}}(\R)$, meaning that\footnote{By a standard diagonal argument, Condition~\eqref{eq:gm:d1approx} is equivalent to the existence of a single sequence in $\mathcal A_{\mathcal{F}}$ converging to $h$ in $W^{1,1}_{\mathrm{loc}}(\R)$ (i.e., on all compact intervals simultaneously).}
\begin{equation}\label{eq:gm:d1approx}
h\in \overline{\mathcal A_{\mathcal{F}}}^{\,W^{1,1}(I)} \qquad \text{for every } h\in W^{1,1}_{\mathrm{nd},\mathrm{loc}}(\R) \text{ and every compact interval } I\subset\R.
\end{equation}
Then the system driven by $\mathcal{F}$ possesses the $\TV$-\UAP\ on $\Pac(\R)$.
\end{proposition}



The proof of Proposition~\ref{prop:d1interp} proceeds by truncation and mollification: replacing $\mu_{\mathrm B}$ and $\mu_\star$ with smooth, strictly positive densities on a compact interval makes $\mathsf{M}_{\mathrm{B} \to \star}$  an increasing diffeomorphism, whose $W^{1,1}$-approximation yields $L^1$-convergence of the resulting densities.

\subsection{Multi-dimensional case}\label{ss:multid}


In one dimension, $W^{1,1}$ approximation of monotone rearrangements provides sufficient control over transported densities. In higher dimensions, however, coordinatewise transformations preserve independence and therefore cannot generate dependence from a product base measure. { This is relevant to generative modeling, where base measures are usually product distributions, such as the standard Gaussian or $\mathcal U([0,1]^d)$, as sampling reduces to drawing the coordinates independently from their one-dimensional marginals. Let
\begin{equation}\label{eq:set.prod.meas}
 \mathcal{P}_{\mathrm{prod}}(\R^d) \coloneqq \left\{ \otimes_{i=1}^d \mu_i \;:\; \mu_i \in \mathcal{P}(\R) \right\}.
\end{equation}
The following proposition shows that decoupled control families cannot approximate dependent target laws from these bases.
}

\begin{proposition}\label{prop:decoupled-obstruction}
Let $d\ge2$ and let $\mathcal F$ be a Lipschitz control family that is \emph{decoupled}, meaning that every $f\in\mathcal F$ has the form
\begin{equation}\label{eq:gm:decouple}
f(x) = \left(f_1(x^{(1)}), \dots, f_d(x^{(d)})\right).
\end{equation}
Then the system generated by $\mathcal F$ does not possess the $\TV$-\UAP\ on $\Pac(\R^d)\cap \mathcal P_{\mathrm{prod}}(\R^d)$.
\end{proposition}


The same obstruction holds for the $W_p$-\UAP\ on $\Pac(\R^d)\cap\mathcal P_{\mathrm{prod}}(\R^d) \cap\mathcal P_p(\R^d)$, since product measures form a weakly closed class. Universality from product bases therefore requires transformations that couple coordinates. For $W_p$, sufficient mechanisms were given in Sections~\ref{ss.emp.approx} and~\ref{ss:transp.map}. For $\TV$, we combine coordinatewise approximation with shear flows.

\begin{theorem}\label{thm:TV-d}
Let $d\ge2$ and let $\mathcal A_{\mathcal{F}}$ be the attainable set generated by a Lipschitz control family $\mathcal{F}$ on $\R^d$. Assume the following conditions:
\begin{enumerate}[label=\textup{\textbf{(A\arabic*)}},ref=(A\arabic*),start=0,leftmargin=*]
  \item\label{ass:A0} 
  $\mathcal A_{\mathcal{F}}$ is closed under inversion: $\Phi\in\mathcal{A}_{\mathcal{F}}$ if and only if $\Phi^{-1}\in\mathcal{A}_{\mathcal{F}}$.

  \item\label{ass:A1} 
  For every compact interval $I\subset\R$, $\alpha\in W^{1,1}_{\mathrm{nd,loc}}(\R)$, $j\in[d]$, and $\varepsilon>0$, there exist $\phi:\R\to\R$ and $\Phi\in\mathcal{A}_{\mathcal{F}}$ of the form
  \begin{equation*}\Phi(x)=\left(x^{(1)},\dots,x^{(j-1)},\phi(x^{(j)}),x^{(j+1)},\dots,x^{(d)}\right)\qquad \text{with }\Vert{}\phi-\alpha\Vert{}_{W^{1,1}(I)}<\varepsilon.\end{equation*}

  \item\label{ass:A2} 
  For every compact interval $J\subset\R$, piecewise affine $\beta\in{C}^0_\mathrm{nd}(\R)$, distinct $i, j \in [d]$, and $\varepsilon>0$, there exist $\psi:\R\to\R$ and $\Phi\in\mathcal{A}_{\mathcal{F}}$ of the form
  \begin{equation*}\Phi(x)=\left(x^{(1)},\dots,x^{(j-1)},x^{(j)}+\psi(x^{(i)}),x^{(j+1)},\dots,x^{(d)}\right)\quad \text{with }\Vert{}\psi-\beta\Vert{}_{L^1(J)}<\varepsilon.\end{equation*}
\end{enumerate}
Then the system possesses the $\TV$-\UAP\ on $\Pac(\R^d)$.
\end{theorem}

Related criteria for $\TV$-universality are established by \citet[Prop.~40]{Ishikawa2023Universal} under approximation of the full transport maps in $W^{0,\infty}_{\mathrm{loc}}\cap W^{1,1}_{\mathrm{loc}}$, which allows one to control both inverse maps and Jacobian determinants. Our conditions instead separate and exploit the structure of coordinatewise maps and shears: \ref{ass:A1} requires $W^{1,1}$ approximation of one-dimensional coordinatewise maps, whereas \ref{ass:A2} requires only $L^1$ approximation of shear profiles. Indeed, for $i\neq j$, the shear $x\mapsto x+\psi(x^{(i)})e_j$ is volume-preserving independently of $\psi'$, allowing us to control the effect of shear approximation on pushforward densities without approximating the derivatives of $\psi$.

\begin{figure}[t]
\centering
\begin{subfigure}{\textwidth}
\centering
\includegraphics[width=0.48\textwidth]{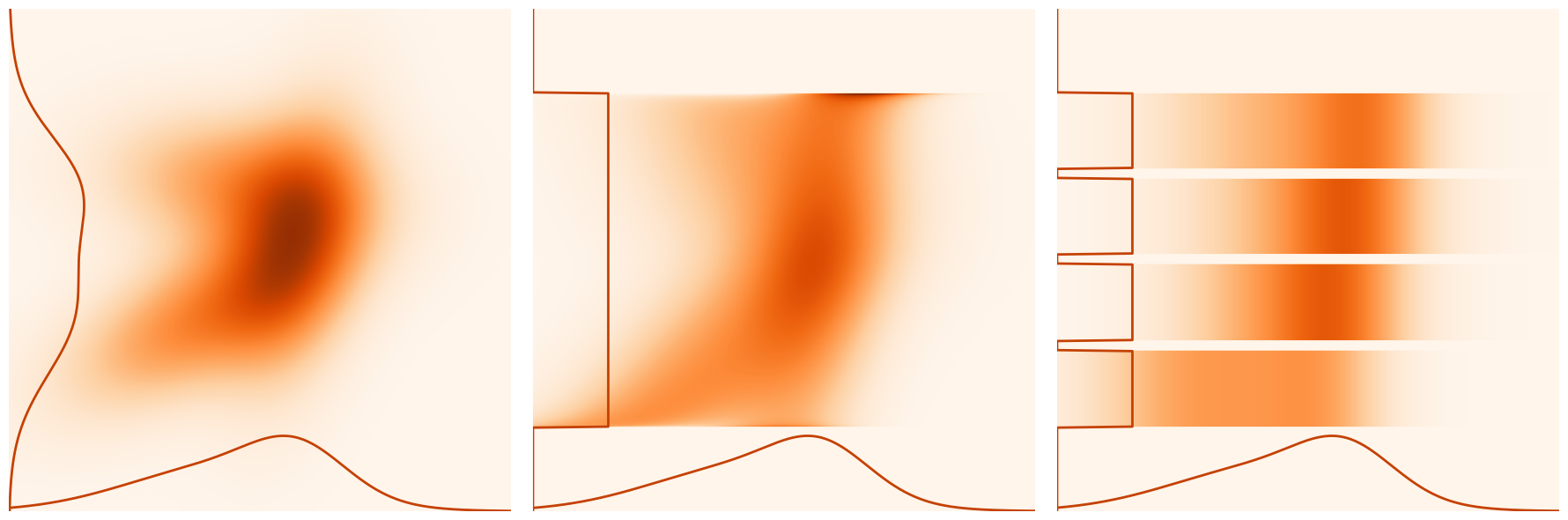}
\hfill
\includegraphics[width=0.48\textwidth]{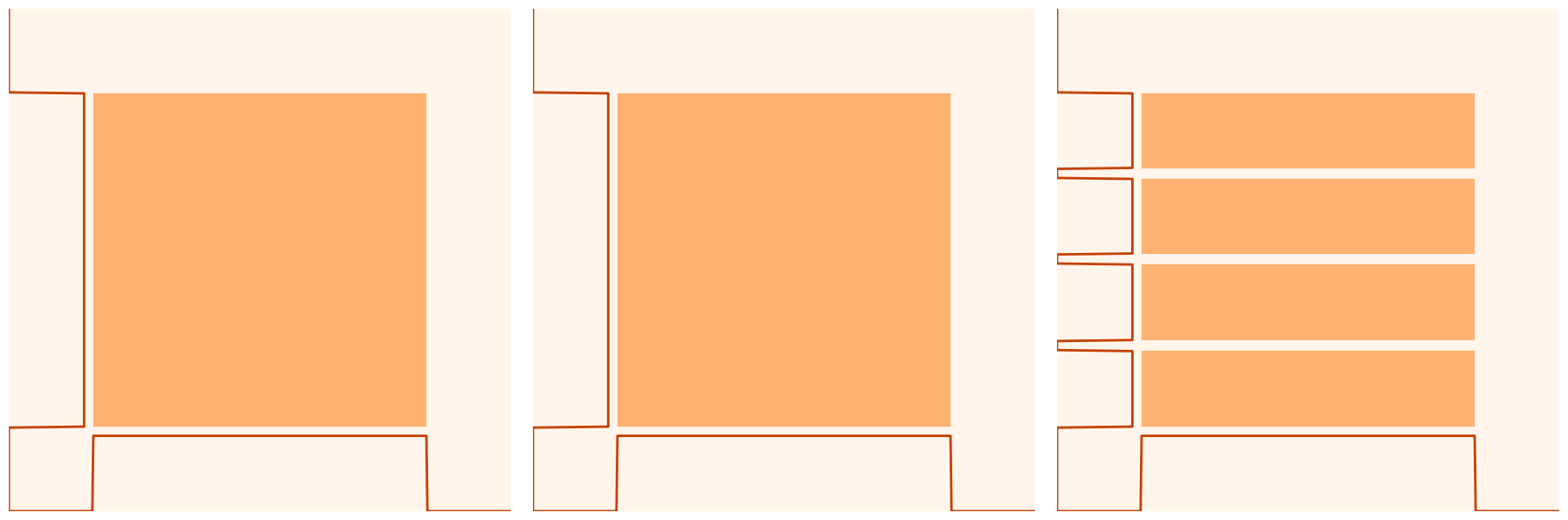}
\caption{Marginal matching and discretization \ref{step.S1}--\ref{step.S2}: after matching the first marginal, the initial measure (\textbf{left}) and the target measure (\textbf{right}, here $\mathcal{U}([0,1]^2)$) are partitioned into finitely many horizontal strips spanning the same vertical intervals and carrying equal mass. (For visual clarity, the vertical and horizontal axes represent $x^{(1)}$ and $x^{(2)}$, respectively.)}
\label{fig:TV-row1}
\end{subfigure}

\vspace{0.4cm}

\begin{subfigure}{\textwidth}
\centering
\includegraphics[width=0.48\textwidth]{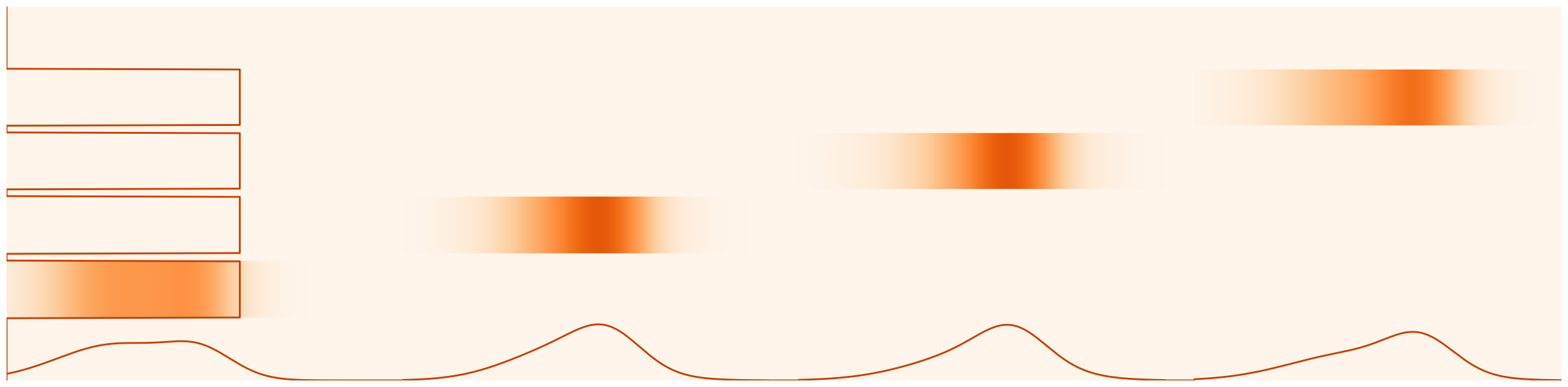}
\hfill
\includegraphics[width=0.48\textwidth]{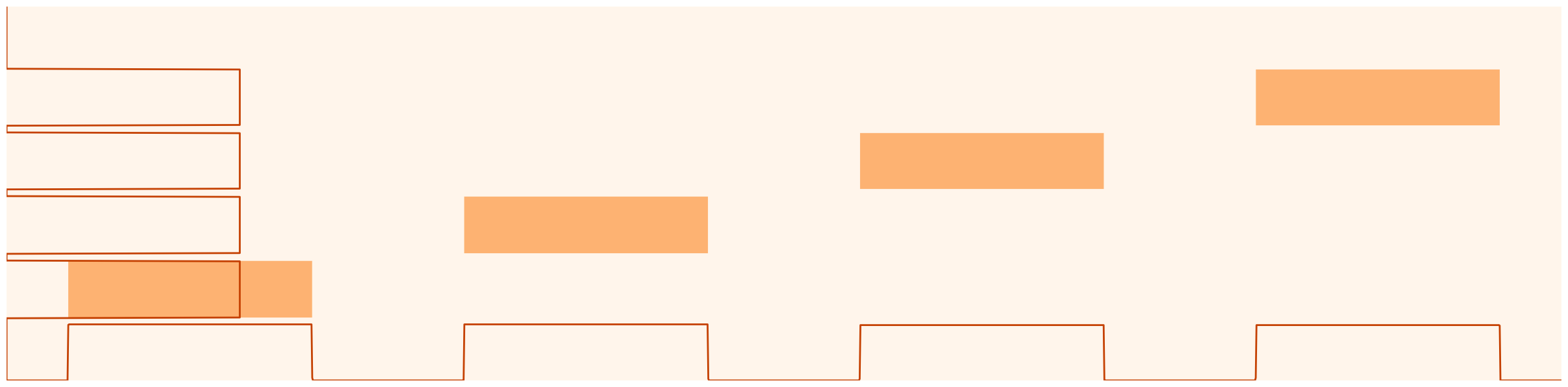}
\caption{Shearing and decoupled conditional transport \ref{step.S3}--\ref{step.S4}: horizontal shear separating the $x^{(2)}$-projections of the strips (\textbf{left}), followed by a single global 1D rearrangement along $x^{(2)}$ (\textbf{right}).}
\label{fig:TV-row2}
\end{subfigure}
\caption{Illustration of the four-step construction in the proof of Theorem~\ref{thm:TV-d} for $d=2$.}
\label{fig:TV-construction}
\end{figure}

Our proof is inspired by \citet[Thm.~1]{RuizBalet2024Control}. We illustrate the core mechanism for $d=2$ in four steps (see Figure~\ref{fig:TV-construction})\footnote{For $d \ge 3$, the proof proceeds inductively along the coordinates $(x^{(1)}, x^{(2)}, \dots,x^{(d)})$.}:

\begin{enumerate}[label=\textbf{(S\arabic*)}, ref=(S\arabic*),leftmargin=*]
    \item\label{step.S1} \textbf{Marginal matching:} Match the $x^{(1)}$-marginal via a coordinatewise map using~\textup{\ref{ass:A1}}.
    
    \item\label{step.S2} \textbf{Discretization:} To approximate a continuum of conditional maps, we partition the domain into finitely many horizontal strips and approximate the densities as constant in $x^{(1)}$ within each strip.
    
    \item\label{step.S3} \textbf{Shearing:} A shear map $(x^{(1)}, x^{(2)}) \mapsto (x^{(1)}, x^{(2)} + \psi(x^{(1)}))$ provided by~\textup{\ref{ass:A2}} translates the strips so that their projections onto the $x^{(2)}$-axis become mutually disjoint.
    
    \item\label{step.S4} \textbf{Decoupled conditional transport:} Since the strips are now spatially separated along $x^{(2)}$, a single global 1D map acting on $x^{(2)}$ via~\textup{\ref{ass:A1}} can rearrange the mass within all strips simultaneously, before undoing the shear via~\textup{\ref{ass:A0}}.
\end{enumerate}

\begin{remark}[Structure of Constructed Flows]
\label{rem.KR}
The construction underlying Theorem~\ref{thm:TV-d} is reminiscent of the recursive structure of the Knothe--Rosenblatt ($\KR$) rearrangement, although we do not establish convergence of the flow maps to $\KR$ in any functional topology.

For $\mu_{\mathrm B},\mu_\star\in\Pac(\R^d)$, the $\KR$ rearrangement from $\mu_{\mathrm B}$ to $\mu_\star$ is defined as the $\mu_{\mathrm B}$-a.e. unique triangular transport map satisfying $\KR_\#\mu_{\mathrm B}=\mu_\star$, of the form
\begin{equation*}
\KR(x)=
\left(
\KR_1(x^{(1)}),
\KR_2(x^{(1)},x^{(2)}),
\dots,
\KR_d(x^{(1)},\dots,x^{(d)})
\right)
\end{equation*}
where each component $\KR_k$ is nondecreasing
in its last argument $x^{(k)}$ for almost every
fixed choice of the preceding coordinates, see \citep{Knothe1957Contributions}.

The map is constructed recursively through one-dimensional monotone rearrangements: $\KR_1$ matches the first marginals, while each subsequent component $\KR_k$ transports the conditional distribution of the $k$-th source coordinate, given the preceding coordinates, onto the corresponding target conditional distribution, given their transported values.

\end{remark}

To conclude this section, we apply Theorem~\ref{thm:TV-d} to affine-invariant families, for which \textup{\ref{ass:A0}} holds automatically by symmetry ($f\in\mathcal{F} \Longleftrightarrow -f\in\mathcal{F}$).

\begin{corollary}\label{cor:gm:gen_TV}
Let $d\ge1$ and let $\mathcal{F}\subset{C}^{0,1}(\R^d;\R^d)$ be affine-invariant. Assume that $\mathcal{F}$ contains a vector field of the form $x\longmapsto \upsigma(x^{(1)})e_1$, where either \textup{(i)} $\upsigma\in{C}^1(\R)$ is non-affine, or \textup{(ii)} $\upsigma(s)=s_+$. Then the system driven by $\mathcal{F}$ possesses the $\TV$-\UAP\ on $\Pac(\R^d)$.
\end{corollary}

This result translates the abstract conditions of Theorem~\ref{thm:TV-d} into standard machine learning architectures: affine invariance reflects the action of fully connected layers, while the requirement on $\upsigma$ is met by most common activation functions. Consequently, the architecture in \eqref{eq:gm:node} suffices to achieve the $\TV$-\UAP.

\section{Discussion}\label{s.discussion}
We conclude with some consequences of our results and several extensions of the framework.

\subsection{Further results on point interpolation}\label{s.discussion.interpolation}

{
\noindent\textit{{Labeled vs.\ unlabeled interpolation.}}
By definition, the \UIP\ implies the \uUIP. At the approximate level, the converse also holds:

\begin{proposition}\label{prop:approx_labeled_unlabeled}
Let $d\ge1$ and let $\mathcal F$ be a Lipschitz control family. The associated system possesses the approximate \uUIP\ if and only if it possesses the approximate \UIP.
\end{proposition}

For $d=1$, this follows directly from order preservation. For $d\ge2$, the proof relies on closure under concatenation and connectedness of the configuration space of distinct points.

One might expect that interpolating generic point configurations would suffice by continuity to guarantee the approximate \uUIP\ globally. The following example shows that this fails: even exact unlabeled interpolation on an open, dense set of full measure does not imply approximate interpolation on all configurations of distinct points.

\begin{example}[Generic exact interpolation vs.\ global failure]\label{ex:generic-unlabeled}
Let $d\ge2$, and consider the control family $\mathcal F$ in \eqref{eq:gm:node} with $\upsigma(s)=s_+$ and $A_{ij}=0$ for all $i\in[d]$ and $j\in\{2,\dots,d\}$ (so the fields depend only on the first coordinate). For each $N\ge1$, consider the open, dense subset of full Lebesgue measure given by configurations with distinct first coordinates:
\begin{equation*}
\mathcal G_N
\coloneqq
\left\{
X=(x_1,\dots,x_N)\in\R^{dN}:
x_i^{(1)}\neq x_j^{(1)}
\text{ whenever }i\neq j
\right\}.
\end{equation*}
Then the system exhibits the following contrast:
\begin{enumerate}[label=\textup{(\roman*)}, leftmargin=*]
    \item For $X,Y\in\mathcal G_N$, there exist $\Phi\in\mathcal A_{\mathcal F}$ and $\pi\in S_N$ such that $\Phi(x_i)=y_{\pi(i)}$ for all $i\in[N]$.
    \item The system satisfies neither the approximate \uUIP\ nor the approximate \UIP\ on $\R^d$.
\end{enumerate}
\end{example}

The obstruction is that attainable maps preserve both the ordering and any ties in the first coordinate ($x_i^{(1)} = x_j^{(1)} \implies \Phi(x_i)^{(1)} = \Phi(x_j)^{(1)}$). While configurations in $\mathcal G_N$ can be matched exactly after sorting, points lying on the same hyperplane $\{x^{(1)} = c\}$ can never be separated along the $x^{(1)}$-axis, preventing the approximation of target configurations with distinct first coordinates.

\medskip

\noindent\textit{{Approximate vs.\ exact interpolation.}}
While Proposition~\ref{prop:approx_labeled_unlabeled} operates up to an approximation error, approximate interpolation can be upgraded to exact interpolation under reversibility and a local rank condition. Our proof adapts the exactification strategy of \citep{LiuLuZuazua2026Autonomous}: instead of tuning the internal parameters of an autonomous vector field, we eliminate the residual error by varying the switching times of concatenated flows.

\begin{proposition}\label{prop:approx_exact_UIP}
Let $d\ge1$ and let $\mathcal F$ be a Lipschitz control family such that $\mathcal A_{\mathcal F}$ is closed under inversion and the associated system possesses the approximate \UIP. Suppose moreover that, for every $N\ge1$, there exist distinct points $\bar x_1,\dots,\bar x_N\in\R^d$ such that
\begin{equation}\label{ass.maxrank}
\operatorname{span}\left\{ \left(f(\bar x_1),\dots,f(\bar x_N)\right): f\in\mathcal F \right\} =\R^{dN}.
\end{equation}
Then the system possesses the exact \UIP.
\end{proposition}

Combining Propositions~\ref{prop:approx_labeled_unlabeled} and~\ref{prop:approx_exact_UIP}, we conclude that the approximate \uUIP\ implies the exact \UIP\ whenever $\mathcal A_{\mathcal F}$ is closed under inversion and satisfies the rank condition~\eqref{ass.maxrank}.

}

\subsection{Beyond global Lipschitz regularity}

While global Lipschitz regularity guarantees bi-Lipschitz flows and stability bounds, relaxing it to local Lipschitz continuity is natural to encompass polynomial control systems, neural ODEs with gated or multiplicative interactions, and flows driven by superlinear confining potentials. When such fields are \emph{complete}, their flows remain well-defined for all $t\in\R$; however, completeness alone does not ensure that $\Phi_\#\mu_{\mathrm B}$ retains finite $p$-th moments when $\mu_{\mathrm B}\in\mathcal P_p(\R^d)$ has unbounded support (so $\Phi_\#\mu_{\mathrm B}$ may leave $\mathcal P_p(\R^d)$).

Several results nonetheless persist under this relaxation. The empirical equivalence in Theorem~\ref{thm:gm:Wp_permUIP} relies only on global existence and injectivity on finite configurations. The arguments in Propositions~\ref{prop:gm:TV_compat_Pac}, \ref{prop:d1interp} and Theorem~\ref{thm:TV-d} are essentially local, suggesting extensions by restriction to compact sets. In contrast, Theorem~\ref{thm:gm:Lip_WpUAP_equiv} and Corollary~\ref{cor:gm:uniform_time_Wp_closure} require global Lipschitz control or suitable substitutes at infinity, and extending Corollary~\ref{cor:gm:gen_TV} would additionally require verifying the activation's approximation properties on unbounded domains beyond localizing the Lie--Trotter argument.

\subsection{Other discrepancies}

{
The matching interpretation developed above extends beyond Wasserstein and total variation. Given a coupling cost  $c: \mathbb{R}^d \times \mathbb{R}^d \to [0, \infty]$, consider
$$
\mathsf D_c(\mu,\nu) \coloneqq \inf_{\gamma\in\Gamma(\mu,\nu)} \int c(x,y)\diff\gamma(x,y).
$$
The choice of $c$ determines how point interpolation is relaxed after source and target particles are optimally paired. The cost $c(x,y)=\|x-y\|^p$ yields $W_p^p$ and penalizes geometric mismatch, whereas $c(x,y)=\mathbf 1_{\{x\neq y\}}$ yields $\TV$ and penalizes inexact matching. On $\mathcal P_{\mathrm{emp}}(\R^d)$, these lead respectively to approximate and exact \uUIP. A related notion is provided by the Prokhorov metric, defined by
$$
d_{\mathrm P}(\mu,\nu) = \inf\left\{ \varepsilon>0: \exists\gamma\in\Gamma(\mu,\nu),\ \gamma\left(\{\|x-y\|>\varepsilon\}\right)\le\varepsilon \right\}.
$$
It thus allows both a geometric error and an exceptional mass of size at most $\varepsilon$.

Beyond this matching interpretation, the empirical strategy of Section~\ref{ss.emp.approx} suggests extensions to other metrics $\mathsf D$, provided uniform empirical measures are dense, perturbations of finite configurations are continuous, and Lipschitz pushforwards satisfy
\begin{equation}\label{eq:D_pushforward_stability}
\mathsf D(\Phi_\#\mu,\Phi_\#\nu) \le\omega_L\left(\mathsf D(\mu,\nu)\right) \qquad\text{if }\Lip(\Phi)\le L,
\end{equation}
where $\omega_L$ is nondecreasing and $\omega_L(r)\to0$ as $r\to0$. Prokhorov and bounded-Lipschitz distances fit this framework, while Maximum Mean Discrepancies require suitable kernel assumptions. Empirical density alone is insufficient for an analogue of Theorem~\ref{thm:gm:Lip_WpUAP_equiv}: uniform Lipschitz control of the interpolating maps and pushforward stability remain essential.
}

Furthermore, discrepancies controlled by $\TV$ inherit our results from Section~\ref{sec:gm:TV}. One example is the Jensen-Shannon divergence, which is of particular relevance in generative modeling (e.g., as the original GAN objective). It is defined for $\mu,\nu\in\Pac(\R^d)$ by
$$
\JS(\mu,\nu) \coloneqq \frac12 \KL(\mu \| \lambda) + \frac12 \KL(\nu \| \lambda), \qquad \text{where } \lambda\coloneqq\frac{\mu+\nu}{2}.
$$
Since $\JS(\mu,\nu) \le (\log 2) \, |\mu-\nu|_{\TV}$, the assumptions of Theorem~\ref{thm:TV-d} (and thus Corollary~\ref{cor:gm:gen_TV}) automatically guarantee the $\JS$-\UAP on $\Pac(\R^d)$.

    


\appendix

\section{Preliminary results}
\label{app:preliminaries}
This appendix collects general properties of flow maps and auxiliary lemmas.

\subsection{Flow maps and pushforward measures}
We begin by recalling fundamental flow estimates and their action on the Lebesgue decomposition of probability measures.
The flow map $\Phi_t^\theta\colon\R^d\to \R^d$ generated by $\dot x(t)=f_{\theta(t)}\left(x(t)\right)$ is a bi-Lipschitz homeomorphism satisfying
\begin{equation}\label{eq:bilip}
\Lip\left(\Phi_t^\theta\right)\vee\Lip\left((\Phi_t^\theta)^{-1}\right)\le e^{\int_0^t \Lip\left(f_{\theta(s)}\right)\diff s},
\end{equation}
and the growth estimate (via Gr\"onwall's inequality)
\begin{equation*}
\|\Phi_t^\theta(x)\| \le e^{\int_0^t \Lip\left(f_{\theta(s)}\right)\diff s}\,\|x\|
+\int_0^t e^{\int_s^t \Lip\left(f_{\theta(u)}\right)\diff u}\,\|f_{\theta(s)}(0)\|\diff s.
\end{equation*}

Any $\mu_{\mathrm{B}}\in\mathcal P(\R^d)$ admits a unique decomposition
\begin{equation*}
\mu_{\mathrm{B}}=\rho_{\mathrm{B}}\,\mathcal L^d+\mu_{\mathrm{B}}^{\mathrm{sc}}+\sum_{k}\alpha_k\,\delta_{x_k},
\end{equation*}
where $\rho_{\mathrm{B}}\in L^1(\R^d)$ is the density function, $\mu_{\mathrm{B}}^{\mathrm{sc}}$ is a singular and atomless measure, and $\alpha_k=\mu_{\mathrm{B}}(\{x_k\})>0$ for all $k$. The pushforward measure
\begin{equation*}
\mu_t^\theta\coloneqq(\Phi_t^\theta)_\#\mu_{\mathrm{B}}
= \rho_t^\theta\,\mathcal L^d + (\Phi_t^\theta)_\#\mu_{\mathrm{B}}^{\mathrm{sc}} + \sum_{k}\alpha_k\,\delta_{\Phi_t^\theta(x_k)},
\end{equation*}
preserves each component in the following sense:
\medskip

\noindent \textbf{(i)} $\rho_t^\theta\,\mathcal L^d=(\Phi_t^\theta)_\#(\rho_{\mathrm{B}}\,\mathcal L^d)$ and, by the method of characteristics, for a.e.\ $x$
\begin{equation}\label{eq:Liouville}
\rho_t^\theta(x)
=\rho_{\mathrm{B}}\left((\Phi_t^\theta)^{-1}(x)\right)\,e^{-\int_0^t \operatorname{div}_x f_{\theta(s)}\left(\Phi_s^\theta\circ(\Phi_t^\theta)^{-1}(x)\right)\diff s}.
\end{equation}

\noindent \textbf{(ii)} The singular--continuous component remains singular--continuous:
\begin{equation*}
(\Phi_t^\theta)_\#\mu_{\mathrm{B}}^{\mathrm{sc}} \perp \mathcal L^d,
\qquad
(\Phi_t^\theta)_\#\mu_{\mathrm{B}}^{\mathrm{sc}} \text{ has no atoms.}
\end{equation*}

\noindent \textbf{(iii)} No new atoms are created, and each existing atom is transported by the flow:
\begin{equation*}
(\Phi_t^\theta)_\#(\alpha_k\delta_{x_k})=\alpha_k\,\delta_{\Phi_t^\theta(x_k)}.
\end{equation*}






 Assuming that $\operatorname{div} f_\theta$ is uniformly bounded in space, \eqref{eq:Liouville} yields for $p\in [1,\infty]$:
\begin{equation}\label{eq:Lpbounds}
e^{-\frac{p-1}{p}\int_0^t \|\operatorname{div} f_{\theta(s)}\|_{L^\infty}\diff s}\, \|\rho_{\mathrm{B}}\|_{L^p}
\le \|\rho_t^\theta\|_{L^p} \le
e^{\frac{p-1}{p}\int_0^t \|\operatorname{div} f_{\theta(s)}\|_{L^\infty}\diff s}\, \|\rho_{\mathrm{B}}\|_{L^p}.
\end{equation}
Moreover, from \eqref{eq:Liouville} we also get 
\begin{equation*}
\rho_t^\theta(x) \ge \rho_{\mathrm{B}}\left((\Phi_t^\theta)^{-1}(x)\right)\,e^{-\int_0^t \|(\operatorname{div} f_{\theta(s)})_+\|_{L^\infty}\diff s}\qquad \text{for a.e. }x\in\R^d.
\end{equation*}
Thus, whenever $\rho_{\mathrm{B}}\ge m>0$ on a compact set $\Omega$, then
\begin{equation*}
\rho_t^\theta(x) \ge m\,e^{-\int_0^t \|(\operatorname{div} f_{\theta(s)})_+\|_{L^\infty}\diff s} > 0\qquad \text{for a.e. }x\in\Phi^\theta_t(\Omega).
\end{equation*}
In particular, it holds that  $\supp(\mu_t^\theta)=\Phi_t^\theta\left(\supp\mu_{\mathrm{B}}\right)$.

\subsection{Approximation and realization by flows}
We next prove some technical lemmas on the approximation and realization of target maps.

\begin{lemma}\label{lem:gm:LieTrotter}
Let $f_1,\dots,f_p\in{C}^{0,1}(\R^d;\R^d)$, and set $f=\sum_{i=1}^p f_i$.
For $t>0$ and $n\in\N$, define
\begin{equation*}
S_{n,t}
=
\left(
\Phi_{t/n}^{f_p}\circ\cdots\circ\Phi_{t/n}^{f_1}
\right)^n
\end{equation*}
where $\Phi^{f_i}_s$ denotes the time-$s$ flow of $f_i$. Then, for every compact $K\subset\R^d$,
\begin{equation*}
\|S_{n,t}-\Phi_t^f\|_{C^0(K)}\to0.
\end{equation*}
If, in addition, the $f_i$ are $C^1$, the convergence
holds in $C^1(K)$.
\end{lemma}

\begin{proof}[Proof of Lemma~\ref{lem:gm:LieTrotter}]
Fix $t>0$ and a compact set $K\subset\R^d$. Write $h=t/n$ and $Q_h=\Phi_h^{f_p}\circ\cdots\circ\Phi_h^{f_1}$, so that $S_{n,t}=Q_h^n$. Global Lipschitz bounds place all exact and splitting trajectories starting from $K$ in a common compact neighborhood $K'\subset\R^d$ independently of $n$, on which all local estimates below hold uniformly.

Since $\|\Phi_s^{f_i}(x)-x\|\le Cs$ and $f_i$ is Lipschitz, the integral equation gives
\begin{equation*}
\Phi_h^{f_i}(x)-x-hf_i(x)
=\int_0^h\left(f_i(\Phi_s^{f_i}(x))-f_i(x)\right)\diff s
=O(h^2).
\end{equation*}
The same argument on $f$ and composing the expansions yields $\Phi_h^f=\Id+hf+O(h^2)$ and $Q_h=\Id+hf+O(h^2)$ in $C^0(K')$, whence $\|Q_h-\Phi_h^f\|_{C^0(K')}\le Ch^2$. Using $\Lip(\Phi_h^f)\le e^{h\Lip(f)}$, 
\begin{equation*}
\|Q_h^{k+1}-\Phi_{(k+1)h}^f\|_{C^0(K)}
\le e^{h\Lip(f)}\|Q_h^k-\Phi_{kh}^f\|_{C^0(K)}+Ch^2.
\end{equation*}
Since the initial error is zero, iteration gives
\begin{equation*}
\max_{0\le k\le n}\|Q_h^k-\Phi_{kh}^f\|_{C^0(K)}
\le Ch^2\sum_{j=0}^{n-1}e^{jh\Lip(f)}
\le C_t h\to0.
\end{equation*}

If the $f_i$ are $C^1$, the variational equation gives
\begin{equation*}
D\Phi_h^f(x)-\Id-hDf(x)
=\!\!\int_0^h
\!\left[\left(Df(\Phi_s^f(x))-Df(x)\right)D\Phi_s^f(x)\!+\!Df(x)\left(D\Phi_s^f(x)\!-\!\Id\right)\right]\!\!\diff s.
\end{equation*}
Since $\|\Phi_s^f(x)-x\|\le Cs$, $\|D\Phi_s^f(x)-\Id\|\le Cs$, and $Df$ is uniformly continuous on $K'$, the integrand is $o(1)$ uniformly on $K'$. Applying the same argument to each $f_i$ and differentiating $Q_h$ via the chain rule, as intermediate points lie within $O(h)$ of $x$, we obtain
\[
D\Phi_h^f=\Id+hDf+o(h)
\qquad\text{and}\qquad
DQ_h=\Id+hDf+o(h)
\quad\text{in }C^0(K').
\]
The Lipschitz bounds give $\|DQ_h\|_{C^0(K')}\le1+Ch$ and $\max_{0\le k\le n}\|D\Phi_{kh}^f\|_{C^0(K)}\le C_t$. Combining the above expansions with the $C^0$ convergence and the uniform continuity of $Df$ on $K'$ yields
\begin{equation*}
\varepsilon_n\coloneqq\max_{0\le k<n}\frac1h
\|DQ_h\circ Q_h^k-D\Phi_h^f\circ\Phi_{kh}^f\|_{C^0(K)}
\le
\max_{0\le k<n}\|Df\circ Q_h^k-Df\circ\Phi_{kh}^f\|_{C^0(K)}+o(1)\to0.
\end{equation*}
Finally, the chain rule gives
\begin{equation*}
D(Q_h^{k+1})-D\Phi_{(k+1)h}^f
=(DQ_h\circ Q_h^k)
\left(D(Q_h^k)-D\Phi_{kh}^f\right)+\left(DQ_h\circ Q_h^k-D\Phi_h^f\circ\Phi_{kh}^f\right)
D\Phi_{kh}^f,
\end{equation*}
which implies $\|D(Q_h^{k+1})-D\Phi_{(k+1)h}^f\|_{C^0(K)}\le(1+Ch)\|D(Q_h^k)-D\Phi_{kh}^f\|_{C^0(K)}+C_t h\varepsilon_n$. 

Iterating from zero initial error yields
\begin{equation*}
\|D(Q_h^n)-D\Phi_t^f\|_{C^0(K)}
\le C_t h\varepsilon_n\sum_{j=0}^{n-1}(1+Ch)^j
\le C_t'\varepsilon_n\to0.
\end{equation*}
\let\BlackBox\relax
\end{proof}

\begin{lemma}\label{lem:gm:Lp_mu_flow_density}
For $d\ge 2$, $p\in[1,\infty)$ and $\mu\in\mathcal P_p(\R^d)$, $\mathcal A_{{C}^{0,1}}$ is dense in $L^p(\mu;\R^d)$.
\end{lemma}

\begin{proof}[Proof of Lemma~\ref{lem:gm:Lp_mu_flow_density}]
Fix $T\in L^p(\mu;\R^d)$ and $\varepsilon>0$. Because $\mu\in\mathcal P_p(\R^d)$ it follows that $\lim_{R\to\infty} R^p \mu(\{ \|x\| > R \}) = 0$. Defining the truncation
\begin{equation*}
T_R(x)\coloneqq
\begin{cases}
T(x), & \|x\|\le R \text{ and } \|T(x)\|\le R,\\
x, & \text{otherwise},
\end{cases}
\end{equation*}
we have $T_R\to T$ in $L^p(\mu)$ as $R\to\infty$ by dominated convergence. Thus, we can fix $R>0$ sufficiently large such that approximating $T_R$ is sufficient. We henceforth assume $T\equiv T_R$.

By Lusin's theorem, we can find a compact $K\subset \overline{B(0,R)}$ such that $T|_K$ is continuous and $\mu(\overline{B(0,R)}\setminus K)$ is arbitrarily small. By uniform continuity, for any $\delta \in (0, 1)$, there exists $\rho>0$ such that $\|x-z\|<2\rho$ implies $\|T(x)-T(z)\|<\delta$ for all $x,z\in K$.

By the Vitali covering theorem applied to \(\mu|_K\),
for every \(\eta>0\) we can choose finitely many pairwise disjoint balls
\(B_i=\overline{B(x_i,r_i)}\), with \(x_i\in K\) and
\(r_i<\rho\wedge 1\), such that $\mu(K\setminus\bigcup_{i=1}^N B_i)<\eta$. For each $i$, pick $y_i$ so that $\|y_i-T(x_i)\|<\delta$, ensuring by infinitesimal shifts that the points $y_i$ are pairwise distinct. Because the family is finite, we may choose pairwise disjoint open balls $D_i\subset B(y_i,\delta)$.

Since $d\ge 2$, there exists $\Psi\in\mathcal A_{{C}^{0,1}}$ mapping each $B_i$ strictly inside $D_i$. Because \(r_i,\delta<1\), the source balls and their target balls lie in
\(B(0,R+2)\). Moreover, $\Psi$ can be chosen to act only within $B(0,R+2)$ and be the identity outside.

We bound the pointwise error over four regions:
\begin{enumerate}[leftmargin=*]
    \item For $x\in K \cap B_i$: $\|\Psi(x)-T(x)\| \le \|\Psi(x)-y_i\| + \|y_i-T(x_i)\| + \|T(x_i)-T(x)\| < 3\delta$.
    \item For  $x\in E_{\mathrm{in}} \coloneqq \overline{B(0,R)}\setminus (K\cap \bigcup_{i=1}^N B_i)$ we have $\|\Psi(x)\|+\|T(x)\|<2R+4$.
    \item For $x\in E_{\mathrm{shell}} \coloneqq B(0,R+2)\setminus \overline{B(0,R)}$, again $\|\Psi(x)\|+\|T(x)\|<2R+4$.
    \item Outside $B(0,R+2)$, we have $\Psi(x)=x=T(x)$.
\end{enumerate}
Combining these, 
\begin{equation*}
\|\Psi - T\|^p_{L^p(\mu)}\le (3\delta)^p\mu(\R^d) + (2R+4)^p \left( \mu(E_{\mathrm{in}}) + \mu(\{ \|x\| > R \}) \right).
\end{equation*}
By choosing $\delta$ small, picking $K$ and Vitali cover to minimize $\mu(E_{\mathrm{in}})$, and having chosen $R$ large enough initially, the total error can be made strictly less than $\varepsilon$.
\end{proof}

\begin{lemma}\label{lem:gm:C1_to_W11}
Let $\mathcal{F}$ be a  Lipschitz control family dense in ${C}^1_{\mathrm{loc}}(\R)$. Then, for every increasing ${C}^1$-diffeomorphism $h:\R\to\R$, every compact interval $I$, and every $\varepsilon>0$, there exists $\Phi\in\mathcal A_{\mathcal{F}}$ such that $\|h-\Phi\|_{W^{1,1}(I)}<\varepsilon$.
\end{lemma}

\begin{proof}[Proof of Lemma~\ref{lem:gm:C1_to_W11}] Define $\Psi_t \coloneqq (1-t)\Id + th$ for $t \in [0,1]$.
Since $h$ is an increasing ${C}^1$-diffeomorphism, each $\Psi_t$ is again an increasing ${C}^1$-diffeomorphism. Let  $I\subset\R$ be a compact interval and define $K \coloneqq \mathrm{conv}( I \cup h(I) )$, so that the tube of trajectories satisfies $\{\Psi_t(x) : t\in[0,1],\, x\in I\} \subseteq K$. Fix an open neighborhood $U$ of $K$ and a compact interval $K_1$ such that $K \subset U \subset K_1$. Define
\[
u_t=\partial_t \Psi_t\circ\Psi_t^{-1}=(h-\Id)\circ\Psi_t^{-1}.
\]
For $K_1=[a,b]$, monotonicity implies
\[
\Psi_t^{-1}(K_1)\subset
J\coloneqq
[a\wedge h^{-1}(a),b\vee h^{-1}(b)]
\qquad \text{for }t\in[0,1].
\]
Since $\min_J h'>0$, the inverse maps depend continuously
on $(t,y)$, and
\[
u_t'(y)=
\frac{h'(\Psi_t^{-1}(y))-1}
     {1-t+t\,h'(\Psi_t^{-1}(y))}
\]
is continuous on $[0,1]\times K_1$.
Thus $u\in C([0,1];C^1(K_1))$ and
$L\coloneqq\sup_{t\in[0,1]}\Lip(u_t|_{K_1})<\infty$.

Since the map $t \mapsto u_t$ is uniformly continuous with values in ${C}^1(K_1)$, and by the density of $\mathcal{F}$ in ${C}^1_{\mathrm{loc}}(\R)$, for any $\delta>0$ we can choose a sufficiently fine time partition to construct a piecewise constant-in-time vector field $w_t$ with values in $\mathcal{F}$ such that
\begin{equation}\label{eq:gm:ut_wt_revised}
\sup_{t\in[0,1]} \|u_t - w_t\|_{{C}^1(K_1)} < \delta.
\end{equation}
Let $\Phi_t$ be the flow of $\dot x = w_t(x)$ on $[0,1]$. For $\delta$ sufficiently small, a standard bootstrap argument ensures that $\Phi_t(I)$ remains inside the open neighborhood $U \subset K_1$ for all $t \in [0,1]$. Indeed, as long as $\Phi_t(I) \subset K_1$, we have
\begin{equation*}
|\partial_t (\Psi_t - \Phi_t)| =| u_t(\Psi_t) - w_t(\Phi_t)| \le \left|u_t(\Psi_t) - u_t(\Phi_t)\right| + \left|u_t(\Phi_t) - w_t(\Phi_t)\right|\le L|\Psi_t - \Phi_t| + \delta.
\end{equation*}
Grönwall's lemma then implies, assuming $(e^L-1)/L=1$ when $L=0$,
\begin{equation}\label{eq:gm:Psit_Phit_revised}
 \sup_{t\in[0,1]}\|\Psi_t - \Phi_t\|_{L^\infty(I)} \le \delta \frac{e^L-1}{L} \eqqcolon C_0 \delta.
\end{equation}
Choosing $\delta$ such that $C_0\delta < \operatorname{dist}(K, \partial U)$ guarantees $\Phi_t(I)\subset U\subset K_1$ for all $t \in [0,1]$.

For the derivatives, set $Y_t(x)\coloneqq \partial_x\Psi_t(x)$ and $Z_t(x)\coloneqq \partial_x\Phi_t(x)$. They satisfy
\begin{equation*}
\partial_t Y_t(x) = u_t'(\Psi_t(x))\,Y_t(x),
\qquad
\partial_t Z_t(x) = w_t'(\Phi_t(x))\,Z_t(x).
\end{equation*}
Let $\Delta_t\coloneqq Y_t-Z_t$. Subtracting the equations gives
\begin{equation}\label{eq:Deltat}
\partial_t \Delta_t
=\left[u_t'(\Psi_t)-u_t'(\Phi_t)\right]Y_t
+\left[u_t'(\Phi_t)-w_t'(\Phi_t)\right]Y_t
+w_t'(\Phi_t)\Delta_t.
\end{equation}
We bound the right-hand side of \eqref{eq:Deltat} evaluated on $I$ term by term:

(i) Since $u'$ is uniformly continuous on $[0,1]\times K_1$, let $\omega_{u'}$ be its modulus of continuity. The first term is bounded by $C_Y\,\omega_{u'}(\|\Psi_t-\Phi_t\|_{L^\infty(I)}) \le C_Y\,\omega_{u'}(C_0\delta)$, where $C_Y\coloneqq\sup_{(t,x)\in[0,1]\times I}|Y_t(x)|$.

(ii) By \eqref{eq:gm:ut_wt_revised}, the second term is strictly bounded by $C_Y\delta$.

(iii) The term $|w_t'(\Phi_t)|$ is bounded by $\|u_t'\|_{{C}^0(K_1)} + \|w_t'-u_t'\|_{{C}^0(K_1)} < L + \delta$ on $I$.

Taking absolute values in \eqref{eq:Deltat} and integrating yields an inequality of the form
\begin{equation*}
|\Delta_t| \le \int_0^t \left[ C_Y\,\omega_{u'}(C_0\delta) + C_Y\delta \right] \diff s + \int_0^t (L+\delta)|\Delta_s| \diff s.
\end{equation*}
Applying Grönwall's lemma yields
\begin{equation*}
\|\Delta_1\|_{L^\infty(I)} \le \eta(\delta)\,e^{L+\delta},
\end{equation*}
where
\begin{equation*}
\eta(\delta)\coloneqq C_Y\delta + C_Y\omega_{u'}(C_0\delta)\to0\qquad \text{as }\delta\to0.
\end{equation*}
Therefore $\|\partial_x h - \partial_x \Phi_1\|_{L^\infty(I)}\to0$ as $\delta\to0$, so we also have
$\|\partial_x h - \partial_x \Phi_1\|_{L^1(I)}\to0$. Combining this with \eqref{eq:gm:Psit_Phit_revised} gives
$\|h-\Phi_1\|_{W^{1,1}(I)}<\varepsilon$ for $\delta$ sufficiently small.
\end{proof}

\begin{lemma}\label{lem:gm:W11_relu_d1}
Let $I\subset\R$ be a compact interval and $u\in W^{1,1}(I)$ a nondecreasing function. Then, for any $\varepsilon>0$ there exists a strictly  increasing, continuous piecewise affine function $\phi:I\to\R$ with finitely many pieces such that $\|u-\phi\|_{W^{1,1}(I)}<\varepsilon.$
\end{lemma}

\begin{proof}[Proof of Lemma~\ref{lem:gm:W11_relu_d1}]
Let $I=[a,b]$ with $a<b$ (the case $a=b$ is trivial). Since $u\in W^{1,1}(I)$ is nondecreasing, we have $u'\ge 0$ a.e.\ on $I$ and $u(x)=u(a)+\int_a^x u'(t)\diff t$ for all $x\in I$. By density of step functions in $L^1(I)$, there exists a nonnegative step function $g$ with finitely many pieces such that $\|u'-g\|_{L^1(I)}<\frac{\varepsilon}{2(1+|I|)}$. Fixing $\eta>0$ with $\eta|I|(1+|I|)<\varepsilon/2$, we define
\begin{equation*}
\phi(x)\coloneqq u(a)+\int_a^x \left(g(t)+\eta\right)\diff t, \qquad x\in I.
\end{equation*}
Then $\phi$ is continuous and piecewise affine with finitely many pieces, and $\phi'=g+\eta\ge\eta>0$ a.e.\ on $I$. Since $|u(x)-\phi(x)| = \left|\int_a^x \left(u'(t)-\phi'(t)\right)\diff t\right| \le \|u'-\phi'\|_{L^1(I)}$ for every $x\in I$, then
\begin{equation*}
\|u-\phi\|_{W^{1,1}(I)}
\le (1+|I|)\|u'-\phi'\|_{L^1(I)}
\le (1+|I|)\left(\|u'-g\|_{L^1(I)}+\eta|I|\right)
<\varepsilon.
\end{equation*}
\end{proof}

\begin{lemma}\label{lem:gm:PWA_exact_compact}
Consider the control family \eqref{eq:gm:node} on $\R$ with $\upsigma(s)=s_+$. Let $I$ be a compact interval and let $h\in{C}^0(I)$ be strictly increasing and piecewise affine. Then there exists an attainable flow map $\Phi$ such that $\Phi(x) = h(x)$ for all $x \in I$.
\end{lemma}

\begin{proof}[Proof of Lemma~\ref{lem:gm:PWA_exact_compact}]
A strictly increasing, piecewise affine function $h$ on $I=[a,b]$ is uniquely determined by its initial value $h(a)$ and its slopes $m_0,\dots,m_N>0$ on the subintervals defined by the knots 
\begin{equation*}
a=z_0 < z_1 < \dots < z_{N+1}=b.
\end{equation*}
We construct $\Phi$ by composing a sequence of flows, matching the target function $h$ from left to right. First, the constant field $f_0(x) = h(a) - a$ belongs to $\mathcal{F}$ (using $a=0$, $b=1$, $w = h(a)-a$). Its time-$1$ flow is the translation $\Phi_0(x) = x + (h(a)-a)$, which matches $h$ at $z_0$ and has a constant slope $S_0 = 1$.

For $k=0, \dots, N$, suppose $\Phi_k$ matches $h$ exactly on $[a, z_k]$ and has a constant slope $S_k$ on $(z_k, b]$. Let $y_k \coloneqq h(z_k)$. We generate a flow $\Psi_k$ using the vector field 
\begin{equation*}
f_k(x) = \alpha_k (x - y_k)_+ \in \mathcal{F}, \qquad \text{where } \alpha_k = \log(m_k / S_k).
\end{equation*}
At $t=1$, this flow fixes $x \le y_k$, and rescales distances by $e^{\alpha_k} = m_k/S_k$ for $x > y_k$.

We define the composition $\Phi_{k+1} \coloneqq \Psi_k \circ \Phi_k$. Because $\Phi_k$ is strictly increasing, we have $\Phi_k(x) \le y_k$ if and only if $x \le z_k$. Consequently, $\Phi_{k+1}$ acts identically to $\Phi_k$ on $[a, z_k]$, thus preserving the exact match up to $z_k$. On the remainder $(z_k, b]$, the previous slope $S_k$ is multiplied by $m_k/S_k$, setting the new slope to exactly $m_k$.

After $N+1$ steps, $\Phi \coloneqq \Phi_{N+1}$ matches the slope of $h$ on every subinterval of $I$ and agrees with $h(a)$, meaning $\Phi(x) = h(x)$ for all $x \in I$.
\end{proof}

\subsection{Total variation stability}
Finally, we record two useful stability properties of the total variation distance.

\begin{lemma}\label{lem:TV-contraction}
Let $\mathsf{T}:\R^d\to\R^d$ be measurable and let $\mu,\nu$ be finite measures. Then
\begin{equation}\label{eq:contraction.ineq.tv}
 \left|\mathsf{T}_\#\mu-\mathsf{T}_\#\nu\right|_{\TV}\ \le\ |\mu-\nu|_{\TV}.   
\end{equation}
If, in addition, $\mathsf{T}$ is bijective and $\mathsf{T}^{-1}$ is measurable, then equality holds:
\begin{equation}\label{eq:contraction.eq.tv}
\left|\mathsf{T}_\#\mu-\mathsf{T}_\#\nu\right|_{\TV}\ =\ |\mu-\nu|_{\TV}.    
\end{equation}
\end{lemma}

\begin{proof}[Proof of Lemma~\ref{lem:TV-contraction}]
For every Borel set $B\subset\R^d$,
$|(\mathsf T_\#\mu)(B)-(\mathsf T_\#\nu)(B)|
=|\mu(\mathsf T^{-1}(B))-\nu(\mathsf T^{-1}(B))|
\le|\mu-\nu|_{\TV}$.
Taking the supremum proves contraction, by \eqref{def:gm:tv}. If $\mathsf T$ is a
bimeasurable bijection, applying contraction to $\mathsf T^{-1}$
gives the reverse inequality.
\end{proof}

\begin{remark}
The $L^1$-error in an initial density is preserved by its evolution
under a common bi-Lipschitz flow. Indeed, if $\rho_t^\eta$ and
$\rho_t^0$ solve the continuity equation with the same velocity field
and initial data $\rho_{\mathrm B}+\eta$ and $\rho_{\mathrm B}$,
respectively, then
$(\rho_t^\eta-\rho_t^0)\mathcal L^d
=(\Phi_t)_\#(\eta\mathcal L^d)$.
Then \eqref{eq:contraction.eq.tv} with $\mathsf{T}=\Phi_t$ gives
$\|\rho_t^\eta-\rho_t^0\|_{L^1(\R^d)}=\|\eta\|_{L^1(\R^d)}$ for all $t\ge0$.
\end{remark}

\begin{lemma}\label{lem:gm:TV_C1_stability}
Let $\mu\in\Pac(\R^d)$ and let
$\psi_n,\psi:\R^d\to\R^d$ be $C^1$-diffeomorphisms such that
$\psi_n\to\psi$ in $C^1_{\mathrm{loc}}(\R^d;\R^d)$.
Then $|(\psi_n)_\#\mu-\psi_\#\mu|_{\TV}\to0$.
\end{lemma}

\begin{proof}[Proof of Lemma~\ref{lem:gm:TV_C1_stability}]
Set $T_n=\psi^{-1}\circ\psi_n$, so that $T_n\to\Id$ in
$C^1_{\mathrm{loc}}$. By Lemma~\ref{lem:TV-contraction},
$|(\psi_n)_\#\mu-\psi_\#\mu|_{\TV}=|(T_n)_\#\mu-\mu|_{\TV}$.
First suppose $\mu=\rho\mathcal L^d$ with $\rho\in C_c(\R^d)$,
and set $K=\supp\rho$ and $J_n=|\det DT_n|$.
Change of variables and equality of the total masses give
\begin{equation*}
|(T_n)_\#\mu-\mu|_{\TV}
=\frac12\int_{\R^d}|\rho-\rho(T_n)J_n|\diff x
=\int_K(\rho-\rho(T_n)J_n)_+\diff x\to0.
\end{equation*}
Indeed, $T_n\to\Id$ and $J_n\to1$ uniformly on $K$,
while $\rho$ is bounded and uniformly continuous.
Thus the integrand converges uniformly to zero on $K$,
which has finite Lebesgue measure.

For general $\mu$, choose probability measures $\mu^{(k)}$
with continuous compactly supported densities such that
$|\mu^{(k)}-\mu|_{\TV}\to0$.
Contraction and the triangle inequality yield
\begin{equation*}
\limsup_{n\to\infty}|(T_n)_\#\mu-\mu|_{\TV}
\le2|\mu-\mu^{(k)}|_{\TV}.
\end{equation*}
Letting $k\to\infty$ proves the claim.
\end{proof}

\section{Proofs of Section \ref{sec:gm:wasserstein} (Wasserstein)}
\label{app:wasserstein}

\begin{proof}[Proof of Proposition~\ref{prop:gm:Wp_compat}]
Suppose
$(\mu_{\mathrm B},\mu_\star)$ is $W_p$-compatible, and let $\Psi_k\in\mathcal A_{C^{0,1}}$ such that
\begin{equation*}
\nu_k\coloneqq(\Psi_k)_\#\mu_{\mathrm B}
\to\mu_\star
\qquad\text{in }W_p.
\end{equation*}

Let $\{x_i\}_{i \in I}$ be the atoms of $\mu_{\mathrm B}$ (where $I$ is countable), and set $m_i \coloneqq \mu_{\mathrm B}(\{x_i\}) > 0$. For every $i \in I$, the sequences $z_i^k \coloneqq \Psi_k(x_i)$ are uniformly bounded in $\R^d$ due to the uniformly bounded $p$-th moments of $\nu_k$. By a diagonal extraction, we may pass to a subsequence such that $z_i^k \to y_i \in \R^d$ for every $i \in I$.

Define $g(x_i) \coloneqq y_i$. To verify this assignment, fix $y \in \R^d$ and a finite subset $J \subset \{i \in I : y_i = y\}$. For every $r>0$, we have $z_i^k \in \overline B(y, r)$ for all $i \in J$ whenever $k$ is sufficiently large. Hence, $\nu_k(\overline B(y, r)) \ge \sum_{i \in J} m_i$. By the Portmanteau theorem,
$$
\mu_\star(\overline B(y, r)) \ge \limsup_{k \to \infty} \nu_k(\overline B(y, r)) \ge \sum_{i \in J} m_i.
$$
Letting $r \to 0$ and taking the supremum over all such finite $J$ yields $\mu_\star(\{y\}) \ge \sum_{i \in I : y_i = y} m_i$. Thus, each $y_i$ is an atom of $\mu_\star$, and $g$ satisfies the required constraint.

Conversely, assume that such a map $g$ exists. Decompose $\mu_{\mathrm B} = \sum_{i \in I} m_i \delta_{x_i} + \mu_{\mathrm B}^{\mathrm c}$, where $\mu_{\mathrm B}^{\mathrm c}$ is atomless, and set $y_i \coloneqq g(x_i)$. The measure
$$
\eta \coloneqq \mu_\star - \sum_{i \in I} m_i \delta_{y_i}
$$
is non-negative, has a finite $p$-th moment, and satisfies $\eta(\R^d) = \mu_{\mathrm B}^{\mathrm c}(\R^d)$.  If this common mass is positive, there exists a Borel map \(T_{\mathrm c}\) such that $(T_{\mathrm c})_\#\mu_{\mathrm B}^{\mathrm c}=\eta.$ If the common mass is zero, choose \(T_{\mathrm c}\) arbitrarily. We define $T: \R^d \to \R^d$ by $T(x_i) = y_i$ for $i \in I$, and \(T = T_{\mathrm c}\) on \(\R^d\setminus\{x_i:i\in I\}\). Since $\mu_{\mathrm B}^{\mathrm c}$ does not charge atoms, $T_\# \mu_{\mathrm B} = \sum_{i \in I} m_i \delta_{y_i} + \eta = \mu_\star$. 

Moreover, $T \in L^p(\mu_{\mathrm B}; \R^d)$ because
$$
\int_{\R^d} \|T(x)\|^p \diff\mu_{\mathrm B}(x) = \int_{\R^d} \|y\|^p \diff\mu_\star(y) < \infty.
$$
By Lemma~\ref{lem:gm:Lp_mu_flow_density}, there exist $\Psi_n \in \mathcal A_{{C}^{0,1}}$ such that $\|\Psi_n - T\|_{L^p(\mu_{\mathrm B})}\to 0$, so \eqref{eq:gm:Wasserstein-Lp} implies that $(\mu_{\mathrm B}, \mu_\star)$ is $W_p$-compatible.
\end{proof}

\subsection{Proofs of Subsection \ref{ss.emp.approx}}

\begin{proof}[Proof of Theorem~\ref{thm:gm:Wp_permUIP}]
We prove the two implications separately.

\noindent\textbf{($\Leftarrow$).} Assume the system possesses the $W_p$-\UAP\ on $\mathcal P_{\mathrm{emp}}(\R^d)$. Fix $\varepsilon>0$ and two finite sets of pairwise distinct points $\{x_i\}_{i=1}^N$ and $\{y_i\}_{i=1}^N$ in $\R^d$. The measures 
$$
\mu_{\mathrm B}^N = \frac{1}{N}\sum_{i=1}^N\delta_{x_i}\qquad\text{and}\qquad\mu_\star^N = \frac{1}{N}\sum_{i=1}^N\delta_{y_i}
$$ 
are $W_p$-compatible, as the \UIP\ for ${C}^{0,1}(\R^d;\R^d)$ guarantees a flow mapping $x_i \mapsto y_i$ (in $d=1$, after reordering the target points increasingly). Applying the $W_p$-\UAP\ with tolerance $\varepsilon/N^{1/p}$ yields $\Phi\in\mathcal A_{\mathcal{F}}$ such that $W_p(\Phi_\#\mu_{\mathrm B}^N,\mu_\star^N)<\varepsilon/N^{1/p}$.

Since $\Phi$ is injective, the points $\{\Phi(x_i)\}_{i=1}^N$ are pairwise distinct. By \eqref{eq:gm:Wp_empirical_assign}, there exists a permutation $\pi\in S_N$ such that:
$$\max_{1\le i\le N}\|\Phi(x_i)-y_{\pi(i)}\|  \le \left(\sum_{i=1}^N \|\Phi(x_i)-y_{\pi(i)}\|^p\right)^{1/p} = N^{1/p}W_p(\Phi_\#\mu_{\mathrm B}^N,\mu_\star^N) < \varepsilon.$$
This proves the approximate \uUIP.

\smallskip
\noindent\textbf{($\Rightarrow$)}
Conversely, assume the approximate \uUIP. Let $\mu_{\mathrm B}=\frac{1}{N}\sum_{i=1}^N\delta_{x_i}\in \mathcal P_{\mathrm{emp}}(\R^d),$ let $\mu_\star$ be a $W_p$-compatible target, and fix $\varepsilon>0$. By Definition~\ref{def:gm:D-compatible}, there exists a sequence of $\Psi_k\in \mathcal A_{{C}^{0,1}}$ such that $W_p((\Psi_k)_\#\mu_{\mathrm B},\mu_\star)\to 0.$ Choose $k$ large enough so
\begin{equation*}
W_p((\Psi_k)_\#\mu_{\mathrm B},\mu_\star)<\frac{\varepsilon}{2}.
\end{equation*}
Since $\Psi_k$ is injective, all $y_i \coloneqq \Psi_k(x_i)$ are pairwise distinct and $(\Psi_k)_\#\mu_{\mathrm B} = \frac{1}{N}\sum_{i=1}^N\delta_{y_i}.$ Applying the approximate \uUIP, we find $\Phi\in\mathcal A_{\mathcal{F}}$ and $\pi\in S_N$ such that
\begin{equation*}
\max_{1\le i\le N}\|\Phi(x_i)-y_{\pi(i)}\|<\frac{\varepsilon}{2}.
\end{equation*}
The permutation $\pi$ defines a transport plan between $\Phi_\#\mu_{\mathrm B}$ and $(\Psi_k)_\#\mu_{\mathrm B}$, hence
\begin{equation*}
W_p^p(\Phi_\#\mu_{\mathrm B},(\Psi_k)_\#\mu_{\mathrm B})
\le
\frac{1}{N}\sum_{i=1}^N \|\Phi(x_i)-y_{\pi(i)}\|^p
<\left(\frac{\varepsilon}{2}\right)^p.
\end{equation*}
Finally, the triangle inequality yields
\begin{equation*}
W_p(\Phi_\#\mu_{\mathrm B},\mu_\star)
\le
W_p(\Phi_\#\mu_{\mathrm B},(\Psi_k)_\#\mu_{\mathrm B})
+
W_p((\Psi_k)_\#\mu_{\mathrm B},\mu_\star)
<
\frac{\varepsilon}{2}+\frac{\varepsilon}{2}
=
\varepsilon.
\end{equation*}
This proves the $W_p$-\UAP\ on $\mathcal P_{\mathrm{emp}}(\R^d)$.
\end{proof}

\begin{proof}[Proof of Proposition~\ref{prop:gm:piUIP_not_WpUAP}]
\textbf{Case (i).}
The system possesses the \UIP\ by
\cite[Thm.~1]{AlvarezLopez2024Interplay}.
Let $\mu_{\mathrm B}$ and $\mu_\star$ be the uniform
probability measures on
\begin{equation*}
Q=(0,1)^d,
\qquad H=(0,1/2)\times(0,1)^{d-1},
\end{equation*}
respectively. This pair is $W_p$-compatible by
Proposition~\ref{prop:gm:Wp_compat}.

Every vector field in $\mathcal F_{\mathrm{div}}$ is globally
Lipschitz with zero divergence a.e., therefore $\Phi_\#\mu_{\mathrm B}$ has density bounded by $1$.
In particular, $\Phi_\#\mu_{\mathrm B}(H)\le\mathcal L^d(H)=1/2$. 
If $(\Phi_n)_\#\mu_{\mathrm B}\to\mu_\star$ in $W_p$,
weak convergence and Portmanteau applied to 
$H$ would give
\begin{equation*}
1=\mu_\star(H)
\le\liminf_{n\to\infty}(\Phi_n)_\#\mu_{\mathrm B}(H)
\le\frac12,
\end{equation*}
a contradiction.

\medskip
\noindent\textbf{Case (ii).}
Set $\upsigma(s)=s_+$ and fix
$\mu_{\mathrm B}\in\Pac(\R^d)\cap\mathcal P_p(\R^d)$.
Every vector field in $\mathcal F_{\upsigma,0}$ vanishes at the
origin, so every attainable map fixes this point.
Thus approximate \uUIP\ fails already for $N=1$,
$x_1=0$, and $y_1\ne0$.

We first prove weak density.
By Proposition~\ref{prop:uip_tv_independent}\textup{(ii)},
$\mathcal A_{\mathcal F_{\tanh,0}}(\mu_{\mathrm B})$
is TV-dense in $\Pac(\R^d)$.
We transfer this approximation to ReLU as follows. Choose  $q_n$ continuous and piecewise affine,
with finitely many pieces and constant on each tail,
such that $\|q_n-\tanh\|_{L^\infty(\R)}\to0$.
For each
$f(x)=w(\tanh(a\cdot x+b)-\tanh(b))
\in\mathcal F_{\tanh,0}$, set
\begin{equation*}
g_n(x)=w\left(q_n(a\cdot x+b)-q_n(b)\right).
\end{equation*}
Then $g_n\in\operatorname{span}(\mathcal F_{\upsigma,0})$, and
\begin{equation*}
\|g_n-f\|_{L^\infty(\R^d)}
\le2\|w\|\,\|q_n-\tanh\|_{L^\infty(\R)}
\to0.
\end{equation*}
Gronwall's inequality gives, for every fixed $t>0$,
\begin{equation*}
\sup_{x\in\R^d}\|\Phi_t^{g_n}(x)-\Phi_t^f(x)\|
\le t e^{t\Lip(f)}\|g_n-f\|_{L^\infty(\R^d)}
\to0.
\end{equation*}

For each $n$,  Lemma~\ref{lem:gm:LieTrotter} approximates
$\Phi_t^{g_n}$ locally uniformly by maps in
$\mathcal A_{\mathcal F_{\upsigma,0}}$.
A diagonal choice, followed by finite composition,
therefore approximates every finite composition of flows
from $\mathcal F_{\tanh,0}$ locally uniformly by maps $\Phi_n\in\mathcal A_{\mathcal F_{\upsigma,0}}$. If $\Phi_n\to\Psi$ locally uniformly, then, for every
bounded continuous $\varphi$, dominated convergence gives
\begin{equation*}
\int\varphi\diff((\Phi_n)_\#\mu_{\mathrm B})
=\int\varphi\circ\Phi_n\diff\mu_{\mathrm B}
\to
\int\varphi\circ\Psi\diff\mu_{\mathrm B}.
\end{equation*}
Thus $\Psi_\#\mu_{\mathrm B}$ belongs to the weak closure of
$\mathcal A_{\mathcal F_{\upsigma,0}}(\mu_{\mathrm B})$.
The proof of Proposition~\ref{prop:uip_tv_independent}\textup{(ii)}
approximates every absolutely continuous target in TV
using finite compositions from $\mathcal F_{\tanh,0}$.
Since TV convergence implies weak convergence, this weak
closure contains $\Pac(\R^d)$.
Finally, weak density of $\Pac(\R^d)$ in $\mathcal P(\R^d)$
proves the claimed weak density.

To obtain $W_p$-density, fix
$\mu_\star\in\mathcal P_p(\R^d)$ and $\varepsilon>0$.
Let $\Pi_M$ denote the coordinatewise projection onto $[-M,M]^d$.
Since $\mu_\star$ has finite $p$-th moment, choose $M>0$
such that
\begin{equation*}
W_p((\Pi_M)_\#\mu_\star,\mu_\star)<\varepsilon/3.
\end{equation*}
By weak density, continuity of $\Pi_M$, and equivalence
of weak and $W_p$-convergence on the compact cube,
there exists $\nu\in \mathcal A_{\mathcal F_{\upsigma,0}}(\mu_{\mathrm B})$ such that
\begin{equation*}
W_p((\Pi_M)_\#\nu,(\Pi_M)_\#\mu_\star)<\varepsilon/3.
\end{equation*}

For $0<\delta<1$, define
\begin{equation*}
H_{M,\delta}(y)\coloneqq\Pi_M(y)+\delta\left(y-\Pi_M(y)\right).
\end{equation*}
This map is attainable: in each coordinate $j$, compose
the time-one flows of
\begin{equation*}
x\mapsto(\log\delta)(x^{(j)}-M)_+e_j,
\qquad
x\mapsto-(\log\delta)(-x^{(j)}-M)_+e_j.
\end{equation*}
Both vector fields belong to $\mathcal F_{\upsigma,0}$ because
$M>0$. Then, by \eqref{eq:gm:Wasserstein-Lp},
\begin{equation*}
W_p((H_{M,\delta})_\#\nu,(\Pi_M)_\#\nu)
\le
\delta\left(\int\|y-\Pi_M(y)\|^p\diff\nu(y)\right)^{1/p}
<\varepsilon/3
\end{equation*}
for sufficiently small $\delta>0$.
The triangle inequality then yields
\begin{equation*}
W_p(H_{M,\delta\#}\nu,\mu_\star)
<\varepsilon,
\end{equation*}
which proves the claim.
\end{proof}

\begin{proof}[Proof of Theorem~\ref{thm:gm:Lip_WpUAP_equiv}]
\emph{($\Rightarrow$)}
Fix $r\ge0$, $\Psi\in\mathcal A_r$,
$\mu_{\mathrm B}\in\mathcal P_p(\R^d)$, and $\varepsilon>0$.
Choose 
$\mu_{\mathrm B}^N=\sum_{i=1}^N\delta_{x_i}/N\in \mathcal P_{\mathrm{emp}}(\R^d)$ such that $W_p(\mu_{\mathrm B}^N,\mu_{\mathrm B})\to0$.
For each $N$ and $\eta>0$, the $\Lambda$-approximate \uUIP\ gives
$\Phi\in\mathcal A_{\mathcal F}$ with $\Lip(\Phi)\le\Lambda(r)$
and $W_p(\Phi_\#\mu_{\mathrm B}^N,\Psi_\#\mu_{\mathrm B}^N)<\eta$.
By the triangle inequality,
\eqref{eq:gm:Wp_Lip}, and because
\begin{equation}\label{eq:gm:reference_Lip}
\Lip(\Psi)\vee\Lip(\Psi^{-1})\le e^r
\qquad\text{for every }\Psi\in\mathcal A_r,
\end{equation}
Gr\"onwall's inequality gives
\begin{equation*}
\begin{aligned}
W_p(\Phi_\#\mu_{\mathrm B},\Psi_\#\mu_{\mathrm B})
&\le
W_p(\Phi_\#\mu_{\mathrm B},\Phi_\#\mu_{\mathrm B}^N)
+W_p(\Phi_\#\mu_{\mathrm B}^N,\Psi_\#\mu_{\mathrm B}^N)
+W_p(\Psi_\#\mu_{\mathrm B}^N,\Psi_\#\mu_{\mathrm B})\\
&<\left(\Lambda(r)+e^r\right)
W_p(\mu_{\mathrm B},\mu_{\mathrm B}^N)+\eta.
\end{aligned}
\end{equation*}
Choosing $N$ sufficiently large and then $\eta<\varepsilon/2$
proves the $\Lambda$-$W_p$-\UAP.

\smallskip
\noindent\emph{($\Leftarrow$)}
Fix $r\ge0$, $\Psi\in\mathcal A_r$, distinct points
$x_1,\dots,x_N$, and $\varepsilon>0$.
Apply the $\Lambda$-$W_p$-\UAP to
$\mu_{\mathrm B}^N=\sum_{i=1}^N\delta_{x_i}/N$ with
accuracy $\varepsilon N^{-1/p}$.
The resulting map satisfies $\Lip(\Phi)\le\Lambda(r)$, and
\eqref{eq:gm:Wp_empirical_assign} gives
\begin{equation*}
\min_{\pi\in S_N}\max_{i\in[N]}
\|\Phi(x_i)-\Psi(x_{\pi(i)})\|
\le N^{1/p}W_p(\Phi_\#\mu_{\mathrm B}^N,
\Psi_\#\mu_{\mathrm B}^N)<\varepsilon.
\end{equation*}
This proves the $\Lambda$-approximate \uUIP.
\end{proof}

{

\begin{proof}[Proof of Example~\ref{ex:gm:basic_quant_profiles}]
For \textup{(i)} and \textup{(ii)}, it suffices to approximate each $\Psi\in\mathcal A_r$ locally uniformly by attainable maps with Lipschitz constant at most $\Lambda(r)$; this implies the $\Lambda$-approximate \uUIP\ with the identity permutation, so Theorem~\ref{thm:gm:Lip_WpUAP_equiv} applies.

We first show that every $f\in C^{0,1}(\R^d;\R^d)$ admits $g_n\in\operatorname{span}(\mathcal F_\rho)$ such that
\begin{equation}\label{eq:gm:bump_field_approx}
g_n\to f\quad\text{locally uniformly},
\qquad
\Lip(g_n)\le\Lip(f)+\frac1n.
\end{equation}
Mollifying $f$ and composing with the smooth $1$-Lipschitz map $S_R(x)=x/\sqrt{1+\|x\|^2/R^2}$ as the mollification scale tends to zero and $R\to\infty$ yields smooth bounded approximations $v$ with $\Lip(v)\le\Lip(f)$. Multiplying $v$ by $\chi(\cdot/M)$, where $0\le\chi\le1$ is a smooth compactly supported cutoff equal to $1$ near the origin, adds at most $\|v\|_\infty\|\nabla\chi\|_\infty/M$ to the Lipschitz constant. Thus $f$ can be approximated locally uniformly by fields $w\in C_c^\infty(\R^d;\R^d)$ with arbitrarily small Lipschitz excess. For each such $w$, convolution with $\rho_s(x)=s^{-d}\rho(x/s)$ converges in  $C^1$ as $s\to 0$, and the integral
\[
(w*\rho_s)(x)
=\int_{\R^d}w(a)s^{-d}\rho\left(\frac{x-a}{s}\right)\diff a
\]
can in turn be approximated in global $C^1$ by finite Riemann sums in $\operatorname{span}(\mathcal F_\rho)$, proving \eqref{eq:gm:bump_field_approx}.

Now let $\Psi=\Phi_{t_m}^{f_m}\circ\cdots\circ\Phi_{t_1}^{f_1}\in\mathcal A_r$ with $\sum_{j=1}^m t_j\Lip(f_j)\le r$, and approximate each $f_j$ by $g_{j,n}\in\operatorname{span}(\mathcal F_\rho)$ via \eqref{eq:gm:bump_field_approx}. By continuous dependence and Gr\"onwall's inequality, the corresponding flows converge locally uniformly to $\Psi$ with Lipschitz constant at most $\exp(\sum_{j=1}^m t_j\Lip(g_{j,n}))$. Next, apply Lemma~\ref{lem:gm:LieTrotter} to each finite sum $g_{j,n}$. For fixed $n$, all constituent bump fields share a common compact support, outside which both the exact flows and their Lie--Trotter splittings equal the identity; hence the $C^1$ convergence in Lemma~\ref{lem:gm:LieTrotter} is global and introduces arbitrarily small Lipschitz excess. Choosing the approximation and splitting tolerances sufficiently small yields $\Phi_n\in\mathcal A_{\mathcal F_\rho}$ satisfying
\begin{equation}\label{eq:gm:bump_flow_approx}
\Phi_n\to\Psi\quad\text{locally uniformly},
\qquad
\Lip(\Phi_n)\le e^r+\frac1n.
\end{equation}

\smallskip
\noindent\textup{(i)}
Fix $\eta>0$. For $r=0$, \eqref{eq:gm:bump_flow_approx} directly provides approximants with Lipschitz constant at most $1+\eta$. For $r>0$ and $n\ge 2$, set
\[
\Psi_n=\Phi_{(1-1/n)t_m}^{f_m}\circ\cdots\circ\Phi_{(1-1/n)t_1}^{f_1}\in\mathcal A_{(1-1/n)r},
\]
so that $\Psi_n\to\Psi$ locally uniformly. Since $e^{(1-1/n)r}<e^r$, applying \eqref{eq:gm:bump_flow_approx} to each $\Psi_n$ yields $\widehat\Phi_n\in\mathcal A_{\mathcal F_\rho}$ with $\|\widehat\Phi_n-\Psi_n\|_{C^0(\overline B_n(0))}<1/n$ and $\Lip(\widehat\Phi_n)\le e^r$, establishing the profile $\Lambda_\eta(r)=(1+\eta)\vee e^r$.

Conversely, every $\Phi\in\mathcal A_{\mathcal F_\rho}$ equals the identity outside a compact set. If $\Lip(\Phi)\le1$, then for any $x\in\R^d$, $u\in\mathbb S^{d-1}$, and sufficiently large $t>0$, we have $\|\Phi(x)-tu\|^2\le\|x-tu\|^2$. Expanding and letting $t\to\infty$ gives $u\cdot(\Phi(x)-x)\ge0$; replacing $u$ with $-u$ forces $\Phi(x)=x$. Thus $\Lambda(0)=1$ permits only the identity map, which cannot approximate the translation $x\mapsto x+e_1\in\mathcal A_0$ even on $\{0\}$.

\smallskip
\noindent\textup{(ii)}
For the approximants in \eqref{eq:gm:bump_flow_approx}, set $c_n=e^r/(e^r+1/n)$. Since the dilation $x\mapsto c_nx$ is the time-$(-\log c_n)$ flow of $x\mapsto-x$, the maps $c_n\Phi_n$ are attainable in the enlarged family, converge locally uniformly to $\Psi$, and satisfy $\Lip(c_n\Phi_n)\le e^r$. Optimality of $\Lambda(r)=e^r$ follows from \eqref{eq:lowbdlip}.

\smallskip
\noindent\textup{(iii)}
Any profile $\Lambda$ implies the qualitative $W_p$-\UAP: given a $W_p$-compatible pair $(\mu_{\mathrm B},\mu_\star)$ and $\varepsilon>0$, compatibility yields $\Psi\in\mathcal A_r$ (for some $r<\infty$) with $W_p(\Psi_\#\mu_{\mathrm B},\mu_\star)<\varepsilon/2$, and Theorem~\ref{thm:gm:Lip_WpUAP_equiv} then provides $\Phi\in\mathcal A_{\mathcal F}$ with $W_p(\Phi_\#\mu_{\mathrm B},\Psi_\#\mu_{\mathrm B})<\varepsilon/2$. For $\mathcal F_{\mathrm{div}}$, this contradicts Proposition~\ref{prop:gm:piUIP_not_WpUAP}\textup{(i)}, which also establishes the exact \UIP.

\smallskip
\noindent\textup{(iv)}
Qualitative universality follows from Proposition~\ref{prop:gm:piUIP_not_WpUAP}\textup{(ii)}. Suppose a profile $\Lambda$ exists and set $K=\Lambda(0)$. Since every attainable map fixes the origin, taking $\mu_a=\mathcal U(B_a(0))$ and $\tau(x)=x+e_1\in\mathcal A_0$, any attainable $\Phi$ with $\Lip(\Phi)\le K$ satisfies
\[
\operatorname{supp}(\Phi_\#\mu_a)\subseteq\overline B_{Ka}(0)
\quad\text{and}\quad
\operatorname{supp}(\tau_\#\mu_a)\subseteq\overline B_a(e_1),
\]
whence
$$W_p(\Phi_\#\mu_a,\tau_\#\mu_a)\ge (1-(K+1)a)_+.$$
Choosing $0<a<1/[2(K+1)]$ rules out arbitrarily accurate approximation, even for absolutely continuous bases.
\end{proof}
}

\begin{proof}[Proof of Corollary~\ref{cor:gm:uniform_time_Wp_closure}]
We rely on the stability estimate \eqref{eq:gm:Wp_Lip}.

\smallskip
\noindent (ii)$\Rightarrow$(i). Fix $\varepsilon>0$. Choose $\Phi\in\mathcal A_{\mathcal{F},T}$ such that $W_p(\Phi_\#\mu_{\mathrm B},\mu_\star)<\varepsilon/2$. Setting $\eta \coloneqq\varepsilon/(2e^T)$, any $\nu\in\mathcal P_{\mathrm{emp}}(\R^d)$ with $W_p(\nu,\mu_{\mathrm B})<\eta$ satisfies
\begin{equation*}
W_p(\Phi_\#\nu,\mu_\star) \le W_p(\Phi_\#\nu,\Phi_\#\mu_{\mathrm B}) + W_p(\Phi_\#\mu_{\mathrm B},\mu_\star) < e^T \eta + \frac{\varepsilon}{2} = \varepsilon.
\end{equation*}

\smallskip
\noindent (i)$\Rightarrow$(ii). Fix $\varepsilon>0$, and let $\eta>0$ be the radius given by (i) for tolerance $\varepsilon/2$. We can pick $\nu\in\mathcal P_{\mathrm{emp}}(\R^d)$ with $W_p(\nu,\mu_{\mathrm B}) < \eta\wedge \varepsilon/(2e^T)$. By (i), there exists $\Phi\in\mathcal A_{\mathcal{F},T}$ such that $W_p(\Phi_\#\nu,\mu_\star)<\varepsilon/2$. Then,
\begin{equation*}
W_p(\Phi_\#\mu_{\mathrm B},\mu_\star) \le e^T W_p(\mu_{\mathrm B},\nu) + W_p(\Phi_\#\nu,\mu_\star) < e^T \frac{\varepsilon}{2e^T} + \frac{\varepsilon}{2} = \varepsilon.
\end{equation*}
Thus, $\mu_\star \in \overline{\mathcal A_{\mathcal{F},T}(\mu_{\mathrm B})}^{\,W_p}$.
\end{proof}

\subsection{Proofs of Subsection~\ref{ss:transp.map}}

\begin{proof}[Proof of Corollary~\ref{cor:gm:WpUAP_affine}]
Fix $d\ge2$, $p\in[1,\infty)$, and let $\mu_{\mathrm B}\in\Pac(\R^d)\cap\mathcal{P}_p(\R^d)$ be compactly supported. By Proposition~\ref{prop:gm:Cheng_lifting}, it suffices to verify that the system possesses both the \UIP\ and the \DSP. The \UIP is a consequence of \cite[Thm.~2.6]{Cheng2025Interpolation}. 

For the second, fix a compact set $K \subset \R^d$ and $h \in {C}^0_{\mathrm{nd}}(\R)$. Let $f\in\mathcal{F}$ be non-affine. There exist $\xi, \eta, b \in \R^d$ such that the scalar function $\tilde{f}(s) \coloneqq \eta\cdot f(s\xi+b)$ is non-affine. For each $i\in[d]$, define the induced one-dimensional control family $\mathcal{F}^{(i)} \coloneqq \{ u\in{C}^{0,1}(\R) : (x \mapsto u(x^{(i)})\,e_i) \in \mathcal{F} \}$. 
This scalar family is affine-invariant: if $u \in \mathcal{F}^{(i)}$, the affine invariance of $\mathcal{F}$ ensures that $x \mapsto \lambda u(a x^{(i)} + c)e_i \in \mathcal{F}$, meaning $s \mapsto \lambda u(a s + c) \in \mathcal{F}^{(i)}$. Moreover, taking $A=\xi e_i^\top$ and $W=e_i\eta^\top$ in Definition~\ref{def:gm:scaling_affine} ensures $\tilde{f}\in\mathcal{F}^{(i)}$. Since $\mathcal{F}^{(i)}$ is an affine-invariant scalar family containing a non-affine element, its associated scalar system possesses the \UIP\ by virtue of \cite[Thm.~2.6]{Cheng2025Interpolation}.

Applying \eqref{eq:gm:1D_UIP_UAP} on the projection $I_i$ of $K$ onto the $i$-th axis, for every $\varepsilon>0$ there exists an attainable scalar flow $\phi^{(i)} \in \mathcal{A}_{\mathcal{F}^{(i)}}$ such that
\begin{equation*}
\sup_{s\in I_i}\left|\phi^{(i)}(s)-h(s)\right|<\varepsilon.
\end{equation*}
Therefore, the coordinatewise lift $\Phi^{(i)}$ (acting as $\phi^{(i)}$ on the $i$-th coordinate and as the identity elsewhere) belongs to $\mathcal{A}_{\mathcal{F}}$. Concatenating these lifted flows sequentially, we obtain a joint flow $\Phi \coloneqq \Phi^{(d)}\circ\dots\circ\Phi^{(1)} \in\mathcal A_{\mathcal{F}}$ satisfying
\begin{equation*}
\sup_{x\in K}\left\|\Phi(x)-\left(h(x^{(1)}),\dots,h(x^{(d)})\right)\right\| \le \sqrt d\,\varepsilon.
\end{equation*}
Thus, the diagonal shrinking property holds, and Proposition~\ref{prop:gm:Cheng_lifting} yields the result.
\end{proof}

\section{Proofs of Section \ref{sec:gm:TV} (total variation)} \label{ss:proofs.tv}

\begin{proof}[Proof of Proposition~\ref{prop:gm:TV_compat_Pac}]
Assume $(\mu_{\mathrm B},\mu_\star)$ is $\TV$-compatible. By definition, there exist maps $\Psi_n\in\mathcal A_{{C}^{0,1}}$ such that $|(\Psi_n)_\#\mu_{\mathrm B}-\mu_\star|_{\TV}\to 0$. Since $\mu_{\mathrm B}\ll\mathcal{L}^d$ and each $\Psi_n$ is a bi-Lipschitz homeomorphism, we have $(\Psi_n)_\#\mu_{\mathrm B}\ll\mathcal{L}^d$ for all $n$. Because $\Pac(\R^d)$ is closed in $\TV$, it follows that $\mu_\star\ll\mathcal{L}^d$, meaning $\mu_\star\in\Pac(\R^d)$.

Conversely, assume $\mu_{\mathrm B},\mu_\star\in\Pac(\R^d)$. Fix $\varepsilon>0$. We can choose a sufficiently large open ball $\Omega\subset\R^d$ and smooth densities $\rho_{\mathrm B}^\varepsilon,\rho_\star^\varepsilon \in {C}^\infty(\overline\Omega)$ that are strictly positive on $\overline\Omega$ and normalized to have mass one, such that
\begin{equation*}
    \left|\mu_{\mathrm B}-\rho_{\mathrm B}^\varepsilon\mathbf{1}_\Omega\mathcal{L}^d\right|_{\TV} < \frac{\varepsilon}{2},
    \qquad
    \left|\mu_\star-\rho_\star^\varepsilon\mathbf{1}_\Omega\mathcal{L}^d\right|_{\TV} < \frac{\varepsilon}{2}.
\end{equation*}

Let $T_\varepsilon:\Omega\to\Omega$ be the quadratic optimal transport map between $\rho_{\mathrm B}^\varepsilon\mathcal{L}^d$ and $\rho_\star^\varepsilon\mathcal{L}^d$. By \cite[Lem.~5]{CaponigroVicari2025Transport}, since $\Omega$ is a bounded ball (and thus uniformly convex) and the densities are smooth and strictly positive up to the boundary, $T_\varepsilon$ belongs to the identity component of the diffeomorphism group of $\overline\Omega$.

Applying \cite[Prop.~6 and Rem.~2]{CaponigroVicari2025Transport} to the constant coordinate fields $e_1,\dots,e_d$ on $\R^d$ yields a global flow composition
\begin{equation*}
    \Psi_\varepsilon \coloneqq \Phi_1^{f_N}\circ\cdots\circ\Phi_1^{f_1} \in \mathcal A_{{C}^{0,1}},
\end{equation*}
where each $f_j$ is a constant or smooth compactly supported scalar multiple of some $e_i$, such that $\Psi_\varepsilon|_\Omega = T_\varepsilon$ and thus $(\Psi_\varepsilon)_\#(\rho_{\mathrm B}^\varepsilon\mathbf{1}_\Omega\mathcal{L}^d) = \rho_\star^\varepsilon\mathbf{1}_\Omega\mathcal{L}^d$.

Finally, using Lemma~\ref{lem:TV-contraction} and the triangle inequality, we conclude
\begin{align*}
    |(\Psi_\varepsilon)_\#\mu_{\mathrm B}-\mu_\star|_{\TV}
    &\le
    |(\Psi_\varepsilon)_\#\mu_{\mathrm B} - (\Psi_\varepsilon)_\#(\rho_{\mathrm B}^\varepsilon\mathbf{1}_\Omega\mathcal{L}^d)|_{\TV}
    +
    |\rho_\star^\varepsilon\mathbf{1}_\Omega\mathcal{L}^d-\mu_\star|_{\TV} \\
    &\le
    |\mu_{\mathrm B}-\rho_{\mathrm B}^\varepsilon\mathbf{1}_\Omega\mathcal{L}^d|_{\TV} + |\rho_\star^\varepsilon\mathbf{1}_\Omega\mathcal{L}^d-\mu_\star|_{\TV} < \varepsilon.
\end{align*}
Thus, $(\mu_{\mathrm B},\mu_\star)$ is $\TV$-compatible.
\end{proof}

\begin{proof}[Proof of Proposition~\ref{prop:gm:Wp_to_TV}]
Choose $\Phi_n\in\mathcal A_{\mathcal F,T}$ such that $W_p((\Phi_n)_\#\mu_{\mathrm B},\mu_\star)\to0$. Then \eqref{eq:bilip} and \eqref{eq:gm:C11_bound} give
\begin{equation}\label{eq:gm:uniform_flow_bounds}
\Lip(\Phi_n)\vee\Lip(\Phi_n^{-1})\le e^T,
\end{equation}
together with uniform bounds on $\Lip(D\Phi_n)$ and $\Lip(D\Phi_n^{-1})$.

Set $K\coloneqq\supp\mu_{\mathrm B}$. Since $(\Phi_n)_\#\mu_{\mathrm B}\to\mu_\star$ in $W_p$, we have $\sup_n\int_K\|\Phi_n(x)\|^p\diff\mu_{\mathrm B}(x)\le M$ for some $M<\infty$, so for each $n$ there exists $x_n\in K$ with $\|\Phi_n(x_n)\|\le M^{1/p}$. Fixing $x_0\in K$, the Lipschitz bound \eqref{eq:gm:uniform_flow_bounds} and the identity $\Phi_n^{-1}(\Phi_n(x_0))=x_0$ yield
\begin{equation*}
\sup_n\|\Phi_n(x_0)\|\le M^{1/p}+e^T\diam K < \infty,
\qquad
\sup_n\|\Phi_n^{-1}(0)\|\le\|x_0\|+e^T\sup_n\|\Phi_n(x_0)\| < \infty.
\end{equation*}
Since $\{\Phi_n\}$ and $\{\Phi_n^{-1}\}$ are equi-Lipschitz by \eqref{eq:gm:uniform_flow_bounds}, boundedness at a single point implies uniform boundedness on every compact set. Applying the Arzel\`a--Ascoli theorem to the maps and their derivatives, followed by a diagonal extraction, we obtain (up to a subsequence)
\begin{equation}\label{eq:gm:C1_convergence}
\Phi_n\to\Phi,
\qquad\Phi_n^{-1}\to\Psi
\qquad\text{in }C^1_{\mathrm{loc}}(\R^d;\R^d)
\end{equation}
for some $\Phi,\Psi\in C^1(\R^d;\R^d)$. Passing to the limit in $\Phi_n^{-1}\circ\Phi_n=\Id$ and $\Phi_n\circ\Phi_n^{-1}=\Id$ gives $\Psi\circ\Phi=\Phi\circ\Psi=\Id$, so $\Phi$ is a global $C^1$-diffeomorphism with inverse $\Psi=\Phi^{-1}$. Lemma~\ref{lem:gm:TV_C1_stability} then yields $|(\Phi_n)_\#\mu_{\mathrm B}-\Phi_\#\mu_{\mathrm B}|_{\TV}\to0$. Since both $\TV$- and $W_p$-convergence imply weak convergence, uniqueness of weak limits gives $\Phi_\#\mu_{\mathrm B}=\mu_\star$, completing the proof.
\end{proof}

The next example illustrates the lack of mutual control between $\TV$ and $W_p$ on $\Pac$. Moreover, part~(ii) shows that uniform convergence of flow maps may fail to produce $\TV$-convergence of the pushforward laws.

\begin{example}[Incomparability of Wasserstein and Total Variation]\label{ex:TV_vs_Wp_incomparable}
Fix $p\in[1,\infty)$.

\noindent\emph{(i) $\TV\to0$ while $W_p$ stays bounded away from zero.}
Let $\mu=\mathcal{U}([0,1]^d)$ and set $\eta_R=(\tau_R)_\#\mu$, where $\tau_R(x)=x+Re_1$. For $\varepsilon_n=1/n$ and $R_n=n^{1/p}+2$, define
\begin{equation*}
    \nu_n\coloneqq(1-\varepsilon_n)\mu+\varepsilon_n\eta_{R_n}\ \in\ \Pac(\R^d)\cap\mathcal{P}_p(\R^d).
\end{equation*}
Since $R_n>1$, $\mu$ and $\eta_{R_n}$ have disjoint supports, so $|\mu-\nu_n|_{\TV}=\varepsilon_n\to0$. On the other hand, any coupling moves a mass of $\varepsilon_n$ across a distance of at least $R_n-1$, hence
\begin{equation*}
    W_p(\mu,\nu_n)^p\ge \varepsilon_n(R_n-1)^p\ge 1.
\end{equation*}

\noindent\emph{(ii) $W_p\to0$ while $\TV$ stays bounded away from zero.}
Let $\mu = \mathcal{U}([0,1])$ and consider the control family $\mathcal F\coloneqq\{f_n:n\in\N\}$ with $f_n(x)\coloneqq\frac1n\sin^2(\pi n x)$, which satisfies $\Lip(f_n) \le \pi$. Let $\Phi_n \coloneqq \Phi_1^{f_n} \in \mathcal A_{\mathcal F,1}$. Since $f_n(k/n)=0$ for all $k\in\Z$, each interval $[k/n, (k+1)/n]$ is $\Phi_n$-invariant. Thus $\|\Phi_n-\Id\|_{L^\infty([0,1])} \le 1/n$, which yields
\begin{equation*}
    W_p\left((\Phi_n)_\#\mu, \mu\right) \le \|\Phi_n-\Id\|_{L^p(\mu)} \le \frac{1}{n} \to 0.
\end{equation*}
However, the $\TV$ distance remains constant. Let $\phi$ be the time-$1$ flow of $\dot y=\sin^2(\pi y)$. By translation and scale invariance of the vector fields, for all $k\in\Z$ and $u\in[0,1]$,
\begin{equation*}
    \Phi_n\left(\frac{k+u}{n}\right) = \frac{k+\phi(u)}{n}
    \qquad\text{and}\qquad
    \Phi_n'\left(\frac{k+u}{n}\right)=\phi'(u).
\end{equation*}
Let $\rho_n = (\Phi_n')^{-1}\circ\Phi_n^{-1}$ be the density of $(\Phi_n)_\#\mu$. Changing variables $y = \Phi_n(x)$ and using the $1/n$-periodicity of $\Phi_n'$ across the $n$ subintervals gives
\begin{equation*}
    2\left|(\Phi_n)_\#\mu - \mu\right|_{\TV} 
    = \int_0^1 |\rho_n(y)-1| \diff y 
    = \int_0^1 |1-\Phi_n'(x)| \diff x
    = \int_0^1|1-\phi'(u)|\diff u \eqqcolon 2c_0.
\end{equation*}
Since $\phi \neq \Id$, its derivative is not identically $1$, so $|(\Phi_n)_\#\mu - \mu|_{\TV}=c_0>0$ for all $n$.

Finally, while $\|f_n'\|_{L^\infty} \le \pi$, the second derivatives diverge: $\Lip(f_n') = 2\pi^2 n \to \infty$. Thus, uniform bounds on first derivatives alone do not control the pushforward densities: without further regularity, the Jacobians may oscillate and $\TV$-convergence may fail.
\end{example}

\begin{proof}[Proof of Proposition~\ref{prop:uip_tv_independent}]
\textbf{Case (i).}
The \UIP\ follows from Proposition~\ref{prop:gm:piUIP_not_WpUAP}. Let $B\subset\R^d$
be a ball centered at the origin and fix $\lambda>1$. Set $\mu_{\mathrm B}\coloneqq\mathcal U(B)$ and $\mu_\star\coloneqq\mathcal U(\lambda B).$ By Proposition~\ref{prop:gm:TV_compat_Pac}, this pair is $\TV$-compatible. Since every $f \in \mathcal{F}_{\mathrm{div}}$ satisfies $\operatorname{div} f = 0$ almost everywhere, $\Phi_\#\mu_{\mathrm B}$ is exactly the uniform probability measure over $\Phi(B)$,  hence
\begin{equation*}
|\Phi_\#\mu_{\mathrm B}-\mu_\star|_{\TV}
\ge
1-\mu_\star(\Phi(B))
=
1-\frac{|\Phi(B)\cap\lambda B|}{|\lambda B|}
\ge
1-\frac{|\Phi(B)|}{|\lambda B|}
=
1-\lambda^{-d}>0.
\end{equation*}
Thus the $\TV$-\UAP fails.

\medskip
\noindent\textbf{Case (ii).}
Every $f\in\mathcal F_{\upsigma,0}$ satisfies $f(0)=0$. Hence every
$\Phi\in\mathcal A_{\mathcal F_{\upsigma,0}}$ fixes the origin. Taking
$N=1$, $x_1=0$ and $y_1\neq0$, we have $\|\Phi(x_1)-y_1\|=\|y_1\|$ for every  $\Phi\in\mathcal A_{\mathcal F_{\upsigma,0}}.$ Thus, choosing $\varepsilon<\|y_1\|$, the approximate \uUIP fails.

We prove the $\TV$-\UAP. Fix $\varepsilon>0$ and
$\mu_{\mathrm B},\mu_\star\in\Pac(\R^d)$. By truncation and mollification,
there exist $0<r<R$ and probability measures
$\widetilde\mu_{\mathrm B},\widetilde\mu_\star$ with smooth strictly
positive densities on $A=\{x\in\R^d:r\le\|x\|\le R\}$ such that
\begin{equation*}
|\mu_{\mathrm B}-\widetilde\mu_{\mathrm B}|_{\TV}
+
|\mu_\star-\widetilde\mu_\star|_{\TV}
<\frac{\varepsilon}{2}.
\end{equation*}
By \cite[Thm.~1]{DacorognaMoser1990Jacobian}, there is a smooth
diffeomorphism $\psi:A\to A$, isotopic to the identity,
such that $\psi_\#\widetilde\mu_{\mathrm B}=\widetilde\mu_\star.$ By \cite[Cor.~1.3]{GrongSchmeding2025Boundary}, 
\begin{equation*}
\psi
=
\Phi_1^{f_N}\circ\cdots\circ\Phi_1^{f_1}
\qquad\text{on } A,
\end{equation*}
where the $f_j$ are smooth vector fields on $ A$ tangent to
$\partial A$. Extend them smoothly to vector fields
$f_j\in C_c^\infty(\R^d\setminus\{0\};\R^d)$ and set
\begin{equation*}
\widehat\psi
\coloneqq
\Phi_1^{f_N}\circ\cdots\circ\Phi_1^{f_1}.
\end{equation*}
Then we have $\widehat\psi=\psi$ on $ A$, and therefore $\widehat\psi_\#\widetilde\mu_{\mathrm B}
=
\widetilde\mu_\star.$

Since $\upsigma$ is globally Lipschitz and non-affine, it is non-polynomial, so finite sums of the form $g(x)=\sum_{k=1}^m w_k\upsigma(a_k\cdot x+b_k)$ are dense in $C^1_{\mathrm{loc}}(\R^d;\R^d)$ by~\cite[Thm.~4.1]{Pinkus1999Approximation}. Approximating on compact sets containing $0$ and using $f_j(0)=0$, we may replace $g$ by
\begin{equation*}
g(x)-g(0) = \sum_{k=1}^m w_k\left(\upsigma(a_k\cdot x+b_k)-\upsigma(b_k)\right) \in \operatorname{span}(\mathcal F_{\upsigma,0}),
\end{equation*}
which preserves derivatives and vanishes at the origin.

Continuous dependence of flows in $C^1$, followed by
Lemma~\ref{lem:gm:LieTrotter}, shows that $\Phi_1^{f_j}$ can be
approximated in $C^1$ on compact sets by maps in
$\mathcal A_{\mathcal F_{\upsigma,0}}$. Composing these approximations and taking a
diagonal sequence over growing compact sets gives $\Phi_n\in\mathcal A_{\mathcal F_{\upsigma,0}}$ with
$\Phi_n\to\widehat\psi$ in $C^1_{\mathrm{loc}}(\R^d;\R^d)$.
Lemma~\ref{lem:gm:TV_C1_stability} therefore yields $|(\Phi_n)_\#\widetilde\mu_{\mathrm B}
-\widetilde\mu_\star|_{\TV}\to0.$ Choosing $n$ sufficiently large and using
Lemma~\ref{lem:TV-contraction},
\begin{equation*}
\begin{aligned}
|(\Phi_n)_\#\mu_{\mathrm B}-\mu_\star|_{\TV}
&\le
|\mu_{\mathrm B}-\widetilde\mu_{\mathrm B}|_{\TV}
+
|(\Phi_n)_\#\widetilde\mu_{\mathrm B}
-\widetilde\mu_\star|_{\TV}
+
|\widetilde\mu_\star-\mu_\star|_{\TV}
<\varepsilon.
\end{aligned}
\end{equation*}
Thus the system possesses the $\TV$-\UAP on $\Pac(\R^d)$.
\end{proof}

\subsection{Proofs of Subsection~\ref{ss:1d}}

\begin{lemma}\label{lem:1d-bdbyw11}
Let $\mu\in\Pac(\R)$ with density $\rho\in \BV_\mathrm{loc}(\R)$. Fix $R>1$ and let $I_R\coloneqq[-R,R]$.
Let $h:\R\to\R$ be an increasing diffeomorphism and define
\begin{equation*}
J_R\coloneqq h^{-1}(I_{R+1}),\qquad m_R\coloneqq \inf_{x\in J_R} h'(x).
\end{equation*}
Assume that $h|_{J_R} \in {C}^{1,1}(J_R)$ and $m_R>0$.
Let $g:\R\to\R$ be an increasing bi-Lipschitz homeomorphism satisfying
\begin{equation*}
\|g-h\|_{W^{1,1}(J_R)}\le 1\wedge C_{R}^{-1},
\end{equation*}
where $C_{R}$ is the embedding constant for $W^{1,1}(J_R)\hookrightarrow{C}^0(\overline{J_R})$.
Then there exists a constant $C=C(h|_{J_R},\rho,R,m_R)$ such that
\begin{equation*}
|g_{\#}\mu - h_{\#}\mu|_{\TV}
\;\le\; C\,\|g-h\|_{W^{1,1}(J_R)}
\;+\; (h_{\#}\mu)(\R\setminus I_R) \;+\; (h_{\#}\mu)(\R\setminus I_{R-1}).
\end{equation*}
\end{lemma}

\begin{proof}[Proof of Lemma~\ref{lem:1d-bdbyw11}]
Because $J_R$ is compact, $\rho|_{J_R} \in \BV(J_R) \subset L^\infty(J_R)$. Since by assumption $h|_{J_R}\in {C}^{1,1}$ and $g$ is an increasing bi-Lipschitz homeomorphism, 
\begin{equation*}
g'(x)\ge \operatorname{Lip}(g^{-1})^{-1}>0 \quad\text{for a.e. }x.
\end{equation*}
Thus, both $h^{-1}$ and $g^{-1}$ are Lipschitz (hence absolutely continuous) on the relevant domains, and the pushforward densities $q_h$ and $q_g$ exist a.e.\ and are given by \eqref{eq:Liouville}. Moreover, 
\begin{equation}\label{eq:gm:deriv_inv}
(h^{-1})'(y)=\frac{1}{h'(h^{-1}(y))},
\qquad
(g^{-1})'(y)=\frac{1}{g'(g^{-1}(y))},\qquad\text{for a.e. }y\in I_R.
\end{equation}
This justifies the denominators and the changes of variables in the subsequent steps.

We decompose the total variation distance as
\begin{align*}
| g_\# \mu - h_\# \mu |_{\TV}
&= \frac12\int_{\R} |q_g(y) - q_h(y)| \diff y \\
&\le \frac12\int_{I_R} |q_g(y) - q_h(y)| \diff y 
   + \frac12 (g_\# \mu)(\R\setminus I_R) 
   + \frac12 (h_\# \mu)(\R\setminus I_R).
\end{align*}
Using $|ab-cd|\le |a-c|\cdot|b| + |c|\cdot|b-d|$, we bound the integral term on $I_R$:
\begin{align}
\frac12\int_{I_R} |q_g(y) - q_h(y)| \diff y
=&\frac12\int_{I_R} \left| \rho\left(g^{-1}(y)\right)\,(g^{-1})'(y) - \rho\left(h^{-1}(y)\right)\,(h^{-1})'(y)\right| \diff y \nonumber \\
\le& \frac12\int_{I_R}\left| \rho\left(g^{-1}(y)\right) - \rho\left(h^{-1}(y)\right) \right|(g^{-1})'(y)\diff y \nonumber \\
&+\frac12\int_{I_R}\rho\left(h^{-1}(y)\right)\left|(g^{-1})'(y) - (h^{-1})'(y)\right|\diff y \eqqcolon\mathcal I_1+ \mathcal I_2. \label{eq:gm:split_I1_I2}
\end{align}

\noindent\textbf{Step 1.} By Sobolev embedding on $J_R$,
\begin{equation*}
\|g-h\|_{L^\infty(J_R)} \le C_R\|g-h\|_{W^{1,1}(J_R)}\le 1.
\end{equation*}
Write $J_R=[a,b]$. Since $h(J_R)=I_{R+1}=[-R-1,R+1]$, we have $h(a)=-R-1$ and $h(b)=R+1$.
Using $|g-h|\le \|g-h\|_{L^\infty(J_R)}\le 1$ at the endpoints,
\begin{equation*}
g(a) \le h(a)+1 = -R, \qquad g(b) \ge h(b)-1 = R.
\end{equation*}
Since $g$ is increasing, $g(J_R)=[g(a),g(b)] \supset [-R,R] = I_R$, and for $x\notin J_R$ monotonicity implies $g(x)\notin I_R$. Hence
\begin{equation}\label{eq:gm:inclusion}
g^{-1}(I_R) \subset J_R.
\end{equation}
Moreover, if $z\in h^{-1}(I_{R-1})$, then $h(z)\in I_{R-1}$ and $|g(z)-h(z)|\le 1$ implies $g(z)\in I_R$.
Thus $h^{-1}(I_{R-1})\subset g^{-1}(I_R)$, i.e.\ $g^{-1}(\R\setminus I_R)\subset h^{-1}(\R\setminus I_{R-1})$, and therefore
\begin{equation}\label{eq:gm:outer}
(g_\#\mu)(\R\setminus I_R) \le (h_\#\mu)(\R\setminus I_{R-1}).
\end{equation}

\noindent\textbf{Step 2.} Let $A \coloneqq g^{-1}(I_R)$ and $T \coloneqq h^{-1}\circ g$. The change of variables $y=g(x)$ yields
\begin{equation*}
\mathcal I_1 = \frac12\int_A \left|\rho(x)-\rho(T(x))\right|\diff x.
\end{equation*}
By \eqref{eq:gm:inclusion}, $A \subset J_R$. If $x \in A$, then $g(x) \in I_R$, so $T(x) \in h^{-1}(I_R) \subset J_R$. Since $h^{-1}$ is $(1/m_R)$-Lipschitz on $I_{R+1}$, we have
\begin{equation*}
\delta \coloneqq \|T-\Id\|_{L^\infty(A)} \le \frac{\|g-h\|_{L^\infty(J_R)}}{m_R} \le \frac{C_R}{m_R}\|g-h\|_{W^{1,1}(J_R)}.
\end{equation*}
(Notice also that $\|g^{-1}-h^{-1}\|_{L^\infty(I_R)} = \delta$ for later use in Step 3). 
Choosing a good representative $\widetilde\rho$ of $\rho$, we have for a.e.\ $x\in A$,
\begin{equation*}
|\widetilde\rho(x)-\widetilde\rho(T(x))| \le |D\rho|\left((x\wedge T(x),\, x\vee T(x)]\right).
\end{equation*}
Integrating over $A$ and applying Tonelli's theorem, we get
\begin{align*}
\int_A |\widetilde\rho(x)-\widetilde\rho(T(x))|\diff x 
&\le \int_{J_R} \left( \int_{A \cap [z-\delta, z+\delta]} \diff x \right) \diff|D\rho|(z) 
\le 2\delta |D\rho|(J_R),
\end{align*}
because if $z$ lies between $x$ and $T(x)$ then $|x-z| \le |x-T(x)| \le \delta$. Thus,
\begin{equation}\label{eq:gm:I1-final}
\mathcal I_1 \le \frac{C_R}{m_R} |D\rho|(J_R)\, \|g-h\|_{W^{1,1}(J_R)}.
\end{equation}

\noindent\textbf{Step 3.} Since $h^{-1}(y) \in J_R$ for $y\in I_R$, we can bound 
\begin{equation*}
\mathcal I_2 \le \frac12\|\rho\|_{L^\infty(J_R)} \int_{I_R} \left|(g^{-1})'(y)-(h^{-1})'(y)\right|\diff y.
\end{equation*}
Using \eqref{eq:gm:deriv_inv}, for a.e.\ $y\in I_R$,
\begin{align*}
|(g^{-1})'(y)-(h^{-1})'(y)|
&=\left|\frac{1}{g'(g^{-1}(y))}-\frac{1}{h'(h^{-1}(y))}\right|
\\
&\le \left|\frac{1}{g'(g^{-1}(y))}-\frac{1}{h'(g^{-1}(y))}\right|
   + \left|\frac{1}{h'(g^{-1}(y))}-\frac{1}{h'(h^{-1}(y))}\right|.
\end{align*}
For the first term, making the change of variables $y=g(x)$ and using $g^{-1}(I_R)\subset J_R$:
\begin{equation*}
\int_{I_R} \left|\frac{1}{g'(g^{-1}(y))}-\frac{1}{h'(g^{-1}(y))}\right| \diff y
= \int_{g^{-1}(I_R)} \frac{|h'(x)-g'(x)|}{h'(x)}\diff x
\le \frac{1}{m_R} \|g'-h'\|_{L^1(J_R)}.
\end{equation*}
For the second term, since $h\in {C}^{1,1}$ on the compact interval $J_R$, the map $x\mapsto 1/h'(x)$ is Lipschitz on $J_R$ with constant $\Lip_{J_R}(h')/m_R^2$. Therefore
\begin{equation*}
\int_{I_R}\left|\frac{1}{h'(g^{-1}(y))}-\frac{1}{h'(h^{-1}(y))}\right|\diff y
\le 2R\,\frac{\Lip_{J_R}(h')}{m_R^2}\,\|g^{-1}-h^{-1}\|_{L^\infty(I_R)}.
\end{equation*}
Recall from Step 2 that $\|g^{-1}-h^{-1}\|_{L^\infty(I_R)} \le \frac{C_R}{m_R}\|g-h\|_{W^{1,1}(J_R)}$. Substituting these bounds yields
\begin{equation}\label{eq:gm:I2-final}
\mathcal I_2 \le \frac12\|\rho\|_{L^\infty(J_R)} \left[ \frac{1}{m_R} + 2R\frac{\Lip_{J_R}(h') C_R}{m_R^3} \right] \|g-h\|_{W^{1,1}(J_R)}.
\end{equation}

Combining the bounds from Steps 1, 2, and 3, the result follows.
\end{proof}

\begin{proof}[Proof of Proposition~\ref{prop:d1interp}]
Let $\mu_{\mathrm B},\mu_\star\in \Pac(\R)$ and denote their densities by $\rho_{\mathrm{B}}$ and $\rho_\star$.
Fix a Gaussian mollifier $\eta_\delta$ and define smooth approximations $\mu_{{\mathrm{B}},\delta}$ and
$\mu_{\star,\delta}$ by convolving the densities. These approximations satisfy
\begin{equation}\label{eq:gm:mollifier}
  |\mu_{{\mathrm{B}},\delta}-\mu_{\mathrm B}|_{\TV} + |\mu_{\star,\delta}-\mu_\star|_{\TV} \to 0
  \qquad \text{as }\delta\to 0.
\end{equation}
Since the convolved densities are positive and smooth, their monotone rearrangement 
$\mathsf{M}_\delta$ (see \eqref{eq:gm:mon_rearrange}) is a smooth increasing diffeomorphism satisfying $(\mathsf{M}_\delta)_{\#}\mu_{{\mathrm{B}},\delta} = \mu_{\star,\delta}$.

Fix $\varepsilon>0$. First, choose $\delta>0$ such that
\begin{equation}\label{eq:gm:delta-choice}
|\mu_{{\mathrm{B}},\delta}-\mu_{\mathrm B}|_{\TV}
+|\mu_{\star,\delta}-\mu_\star|_{\TV}
<\frac{\varepsilon}{3}.
\end{equation}
Next, choose $R>1$ large enough so that $I_R\coloneqq[-R,R]$ satisfies:
\begin{equation}\label{eq:gm:R-choice}
\mu_{\star,\delta}(\R\setminus I_R) + \mu_{\star,\delta}(\R\setminus I_{R-1})
<\frac{\varepsilon}{3}.
\end{equation}

Define $J_R \coloneqq \mathsf{M}_\delta^{-1}(I_{R+1})$. Since $\mathsf{M}_\delta$ is continuous and proper,
$J_R$ is a compact interval. Furthermore, since $\mathsf{M}_\delta$ is smooth and strictly increasing,
\begin{equation*}
m_R \coloneqq \inf_{x\in J_R}\mathsf{M}_\delta'(x)>0.
\end{equation*}
By \eqref{eq:gm:d1approx} applied with $h=\mathsf{M}_\delta$ and $I=J_R$, for any $\eta>0$
there is $\Phi\in\mathcal A_{\mathcal{F}}$ such that
\begin{equation*}
\|\Phi-\mathsf{M}_\delta\|_{W^{1,1}(J_R)}\le \eta,
\qquad
\|\Phi-\mathsf{M}_\delta\|_{L^\infty(J_R)}\le C_R\,\eta,
\end{equation*}
where $C_R$ is the embedding constant $W^{1,1}(J_R)\hookrightarrow{C}^0(\overline{J_R})$.
We may assume $\eta\le 1\wedge C_R^{-1}$ (decreasing $\eta$ if necessary). Note that $\Phi$ is a
bi-Lipschitz increasing homeomorphism, so we can apply Lemma~\ref{lem:1d-bdbyw11} with
$\mu=\mu_{{\mathrm{B}},\delta}$, $h=\mathsf{M}_\delta$, and $g=\Phi$.
Since $(\mathsf{M}_\delta)_\#\mu_{{\mathrm{B}},\delta}=\mu_{\star,\delta}$, we get 
\begin{equation}\label{eq:gm:middle-estimate}
|\Phi_{\#}\mu_{{\mathrm{B}},\delta}-\mu_{\star,\delta}|_{\TV}
\le C\,\eta
+ \mu_{\star,\delta}(\R\setminus I_R) + \mu_{\star,\delta}(\R\setminus I_{R-1}),
\end{equation}
for some constant $C$ depending on $(\mathsf{M}_\delta|_{J_R},\rho_{\mathrm{B}}*\eta_\delta,R,m_R)$.

Finally, using the triangle inequality and Lemma~\ref{lem:TV-contraction},
\begin{align*}
|\Phi_{\#}\mu_{\mathrm B}-\mu_\star|_{\TV}
&\le |\Phi_{\#}\mu_{\mathrm B}-\Phi_{\#}\mu_{{\mathrm{B}},\delta}|_{\TV}
   +|\Phi_{\#}\mu_{{\mathrm{B}},\delta}-\mu_{\star,\delta}|_{\TV}
   +|\mu_{\star,\delta}-\mu_\star|_{\TV}\\
&\le |\mu_{\mathrm B}-\mu_{{\mathrm{B}},\delta}|_{\TV}
   +|\Phi_{\#}\mu_{{\mathrm{B}},\delta}-\mu_{\star,\delta}|_{\TV}
   +|\mu_{\star,\delta}-\mu_\star|_{\TV}.
\end{align*}
Substituting \eqref{eq:gm:delta-choice}, \eqref{eq:gm:R-choice}, and \eqref{eq:gm:middle-estimate},
and choosing $\eta<\varepsilon/(3C)$, the RHS is $<\varepsilon$.
\end{proof}

\subsection{Proofs of Subsection~\ref{ss:multid}}\label{app.multi.d.tv}

\begin{proof}[Proof of Proposition~\ref{prop:decoupled-obstruction}]
Since \eqref{eq:gm:decouple} splits the controlled ODE
into independent scalar equations, every attainable
map acts coordinatewise:
$\Phi(x)=(\phi_1(x^{(1)}),\dots,\phi_d(x^{(d)}))$.
Consequently,
\begin{equation*}
\Phi_\#\left(\bigotimes_{i=1}^d\mu_i\right)
=\bigotimes_{i=1}^d(\phi_i)_\#\mu_i,
\end{equation*}
as follows by evaluating both sides on Borel rectangles.
Thus attainable pushforwards preserve product measures.

This class is closed in $\TV$. Indeed, if
$\nu^n=\bigotimes_{i=1}^d\nu_i^n\to\nu$ in $\TV$, then
Lemma~\ref{lem:TV-contraction} gives
$\nu_i^n\to\nu_i\coloneqq(\pi^i)_\#\nu$ in $\TV$. For every
Borel rectangle $B=\prod_{i=1}^dB_i$,
\begin{equation*}
\nu(B)=\lim_n\nu^n(B)
=\lim_n\prod_{i=1}^d\nu_i^n(B_i)
=\prod_{i=1}^d\nu_i(B_i).
\end{equation*}
Uniqueness of probability measures on the generating class of
rectangles yields $\nu=\bigotimes_{i=1}^d\nu_i$
\cite[Thm.~2.1.6]{Durrett2019Probability}.
Hence the $\TV$-closure of the attainable set from a product base
contains only product measures.

Finally, $\mu_{\mathrm B}=\mathcal N(0,\Id)$ and
$\mu_\star=\mathcal N(0,\Sigma)$, with $\Sigma$ positive definite
and possessing a nonzero off-diagonal entry, form a $\TV$-compatible
pair by Proposition~\ref{prop:gm:TV_compat_Pac}. The base is a product measure, whereas the target is
not, since its covariance matrix is not diagonal. Hence the $\TV$-\UAP fails.
\end{proof}

\begin{proof}[Proof of Theorem~\ref{thm:TV-d}]

Fix $\mu_{\mathrm B},\mu_\star\in\Pac(\R^d)$ and $\varepsilon>0$. It suffices to construct two maps $\Phi_{\rightarrow},\Phi_{\leftarrow}\in\mathcal A_{\mathcal{F}}$ and an intermediate measure $\nu$ such that
$$
\left|\,(\Phi_{\rightarrow})_\#\mu_{\mathrm B}-\nu\,\right|_{\TV}<\frac{\varepsilon}{2}
\quad\text{and}\quad
\left|\,(\Phi_{\leftarrow})_\#\mu_\star-\nu\,\right|_{\TV}<\frac{\varepsilon}{2}.
$$
Indeed, by \ref{ass:A0} we have $\Phi_{\leftarrow}^{-1}\in\mathcal A_{\mathcal{F}}$, so that $\Phi\coloneqq \Phi_{\leftarrow}^{-1}\circ\Phi_{\rightarrow}\in\mathcal A_{\mathcal{F}}$ (by time concatenation of controls), and using Lemma~\ref{lem:TV-contraction} we conclude:
$$
\begin{aligned}
\left|\,\Phi_\#\mu_{\mathrm B}-\mu_\star\,\right|_{\TV}
&=\left|\,(\Phi_{\leftarrow})_\#(\Phi_\#\mu_{\mathrm B})-(\Phi_{\leftarrow})_\#\mu_\star\,\right|_{\TV}
=\left|\,(\Phi_{\rightarrow})_\#\mu_{\mathrm B}-(\Phi_{\leftarrow})_\#\mu_\star\,\right|_{\TV}\\
&\le \left|\,(\Phi_{\rightarrow})_\#\mu_{\mathrm B}-\nu\,\right|_{\TV}
     +\left|\,\nu-(\Phi_{\leftarrow})_\#\mu_\star\,\right|_{\TV}
<\varepsilon.
\end{aligned}
$$

We now build $\Phi_{\rightarrow}$ and $\Phi_{\leftarrow}$, first treating the case $d=2$ and then indicating the modifications required for $d>2$. 

\noindent\textbf{(S1) Matching the first marginal.} Let $\mu_{{\mathrm{B}},1}\coloneqq (\pi^1)_\#\mu_{\mathrm B}$ and $\mu_{\star,1}\coloneqq (\pi^1)_\#\mu_\star.$ By \textup{\ref{ass:A1}} with $j=1$, the family $\mathcal{F}$ induces a 1D control system with reachable set
$$
\mathcal A_{\mathcal{F}}^1\coloneqq
\left\{\phi:\R\to\R\ \mid\ \exists\,\Phi\in\mathcal A_{\mathcal{F}} \text{ such that }\ \Phi\left(x^{(1)},x^{(2)}\right)=\left(\phi(x^{(1)}),x^{(2)}\right)\right\},
$$
satisfying \eqref{eq:gm:d1approx}. By Proposition~\ref{prop:d1interp}, for any $\delta_1>0$ there is $\phi_1\in\mathcal A_{\mathcal{F}}^1$ such that
$$
\left|\,(\phi_1)_\#\mu_{{\mathrm{B}},1}-\mu_{\star,1}\,\right|_{\TV}<\delta_1.
$$
Let $\Phi_1\in\mathcal A_{\mathcal{F}}$ be defined by $\Phi_1(x^{(1)},x^{(2)})=(\phi_1(x^{(1)}),x^{(2)})$, and set $\mu_{\mathrm B}^{\mathrm{S1}}\coloneqq(\Phi_1)_\#\mu_{\mathrm B}$. Then
$$
\left|\,(\pi^1)_\#\mu_{\mathrm B}^{\mathrm{S1}}-\mu_{\star,1}\,\right|_{\TV}
=\left|\,(\phi_1)_\#\mu_{{\mathrm{B}},1}-\mu_{\star,1}\,\right|_{\TV}<\delta_1.
$$

\smallskip
\noindent\textbf{(S2) Layerwise discretization.} We approximate $\mu_{\mathrm B}^{\mathrm{S1}}$ and $\mu_\star$ on a large square by finite measures supported on vertical layers. On each layer, the two approximating densities are independent of $x^{(1)}$ and have the same mass.

Let $K_R\coloneqq[-R,R]^2$. For each integer $m\ge2$, set $\gamma_m\coloneqq R/m^2$ and $h_m\coloneqq 2R/m-(m-1)\gamma_m/m$. Partition $[-R,R]$ into $m$ intervals $(a_k,b_k)$ of length $h_m$, separated by $m-1$ gaps of length $\gamma_m$. Define the layers $L_k\coloneqq(a_k,b_k)\times[-R,R]$, their union $L\coloneqq\bigcup_{k=1}^m L_k$, and the gap region $G\coloneqq K_R\setminus L$. Note that $h_m\to0$ and $\mathcal L^2(G)=2R(m-1)\gamma_m\to0$ as $m\to\infty$.

Write $\mu_{\mathrm B}^{\mathrm{S1}}=\rho_{\mathrm{B}}^{\mathrm{S1}}\,\mathcal{L}^2$ and $\mu_\star=\rho_\star\,\mathcal{L}^2$. We define the $x^{(1)}$-averaged densities
\[
\hat\rho_{\mathrm{B}}(x)\coloneqq \frac{1}{h_m}\int_{a_k}^{b_k}\rho_{\mathrm{B}}^{\mathrm{S1}}\bigl(s,x^{(2)}\bigr)\diff s \quad \text{for } x\in L_k,\qquad\text{and}\qquad \hat\rho_{\mathrm{B}}(x)\coloneqq 0\quad \text{for } x\notin L
\]
(and analogously for $\hat\rho_\star$). To match the layer masses, set $m_k \coloneqq \mu_{\mathrm B}^{\mathrm{S1}}(L_k)\wedge\mu_\star(L_k)$ and define the discretized measures $\mu_{\mathrm B}^{\mathrm{S2}}$ and $\mu^{\mathrm{S2}}_{\star}$ via the rescaled densities
\begin{equation}\label{eq:RhoS2}
 \rho^{\mathrm{S2}}_{\mathrm{B}}\coloneqq \sum_{k=1}^m \frac{m_k}{\mu_{\mathrm B}^{\mathrm{S1}}(L_k)}\,\hat\rho_{\mathrm{B}}\,\mathbf{1}_{L_k}
\qquad\text{and}\qquad
\rho^{\mathrm{S2}}_{\star}\coloneqq \sum_{k=1}^m \frac{m_k}{\mu_\star(L_k)}\,\hat\rho_\star\,\mathbf{1}_{L_k},   
\end{equation}
with the convention $0/0=0$.

\noindent\emph{Claim.} Provided $\delta_1>0$ (from \textbf{(S1)}) is small enough, for each $\delta_2>0$ there exist $R>0$ and an integer $m\ge2$ such that
\[
\left|\mu_{\mathrm B}^{\mathrm{S2}}-\mu_{\mathrm B}^{\mathrm{S1}}\right|_{\TV}<\delta_2
\qquad\text{and}\qquad
\left|\mu_\star^{\mathrm{S2}}-\mu_\star\right|_{\TV}<\delta_2.
\]

\smallskip
\noindent\emph{Proof of Claim.} First, choose $R>0$ large enough that $\mu(\R^2\setminus K_R)<\delta_2/32$ for both $\mu\in\{\mu_{\mathrm B}^{\mathrm{S1}},\mu_\star\}$. Since $\mathcal L^2(G)\to0$ as $m\to\infty$, for every sufficiently large $m$ we also have $\mu(G)<\delta_2/32$, and hence $\mu(\R^2\setminus K_R)+\mu(G)<\delta_2/16$, for both $\mu\in\{\mu_{\mathrm B}^{\mathrm{S1}},\mu_\star\}$.

Because $\rho_{\mathrm{B}}^{\mathrm{S1}}\in L^1(K_R)$, its strip average $\hat\rho_{\mathrm{B}}$ satisfies $\|\rho_{\mathrm{B}}^{\mathrm{S1}}-\hat\rho_{\mathrm{B}}\|_{L^1(L)}\to0$ as $h_m\to0$ ($m\to\infty$). Thus, for $m$ large enough, the measure $\hat\mu_{\mathrm B}\coloneqq\hat\rho_{\mathrm B}\,\mathcal L^2$ satisfies
\begin{equation}\label{eq:gm:S2_1}
\left|\mu_{\mathrm B}^{\mathrm{S1}}-\hat\mu_{\mathrm B}\right|_{\TV}
\le \frac12\|\rho_{\mathrm{B}}^{\mathrm{S1}}-\hat\rho_{\mathrm{B}}\|_{L^1(L)} + \mu_{\mathrm B}^{\mathrm{S1}}(G) + \mu_{\mathrm B}^{\mathrm{S1}}(\R^2\setminus K_R)
<\frac{7\delta_2}{16}.
\end{equation}
On each $L_k$, we have $|\rho_{\mathrm{B}}^{\mathrm{S2}}-\hat\rho_{\mathrm{B}}| = |1-m_k/\mu_{\mathrm B}^{\mathrm{S1}}(L_k)|\hat\rho_{\mathrm{B}}$. Integrating over $L$ gives
\[
\|\rho_{\mathrm{B}}^{\mathrm{S2}}-\hat\rho_{\mathrm{B}}\|_{L^1(L)}
= \sum_{k=1}^m \left|\mu_{\mathrm B}^{\mathrm{S1}}(L_k)-m_k\right|
\le \sum_{k=1}^m \left|\mu_{\mathrm B}^{\mathrm{S1}}(L_k)-\mu_\star(L_k)\right|.
\]
Since the intervals $(a_k,b_k)$ are disjoint,
\begin{align*}
\sum_{k=1}^m\left|\mu_{\mathrm B}^{\mathrm{S1}}(L_k)-\mu_\star(L_k)\right|
&\le \sum_{k=1}^m\left|(\pi^1)_\#\mu_{\mathrm B}^{\mathrm{S1}}\bigl((a_k,b_k)\bigr)-(\pi^1)_\#\mu_\star\bigl((a_k,b_k)\bigr)\right| \\
&\quad +\mu_{\mathrm B}^{\mathrm{S1}}(\R^2\setminus K_R)+\mu_\star(\R^2\setminus K_R) \\
&\le 2\left|(\pi^1)_\#\mu_{\mathrm B}^{\mathrm{S1}}-(\pi^1)_\#\mu_\star\right|_{\TV}
+\frac{\delta_2}{16}
<2\delta_1+\frac{\delta_2}{16},
\end{align*}
where the last inequality uses \textbf{(S1)} and the choice of $R$. Consequently, $|\mu_{\mathrm B}^{\mathrm{S2}}-\hat\mu_{\mathrm B}|_{\TV} < 2\delta_1 + \delta_2/16$. Combining this with \eqref{eq:gm:S2_1} via the triangle inequality yields $|\mu_{\mathrm B}^{\mathrm{S2}}-\mu_{\mathrm B}^{\mathrm{S1}}|_{\TV} < 2\delta_1 + \delta_2/2$, and the same holds for $|\mu_\star^{\mathrm{S2}}-\mu_\star|_{\TV}$ by symmetry. Choosing $\delta_1<\delta_2/4$ completes the proof.

\smallskip

\noindent\textbf{(S3) Shearing.} We now separate the layers vertically to ensure that their projections onto the second coordinate are pairwise disjoint.

Fix a separation $S>0$ and define $S_k \coloneqq (k-1)(2R+S)$ for $k\in[m]$. Let $\beta:\R\to\R$ be the unique nondecreasing, continuous, piecewise affine profile that equals $S_k$ on each layer $[a_k, b_k]$, is affine on the gaps, and remains constant outside $[-R,R]$. Define the ideal shear map $T_\beta(x)\coloneqq (x^{(1)},\,x^{(2)}+\beta(x^{(1)}))$ and
$$
\mu_{\mathrm B}^{\mathrm{S3}} \coloneqq (T_\beta)_\# \mu_{\mathrm B}^{\mathrm{S2}}
\qquad\text{and}\qquad
\mu_\star^{\mathrm{S3}} \coloneqq (T_\beta)_\# \mu_\star^{\mathrm{S2}}.
$$
Since $\beta\equiv S_k$ on each layer $L_k$, the map $T_\beta$ acts on $L_k$ as a rigid translation by $S_k$ in the $x^{(2)}$-direction. Consequently, the sheared layers $\tilde L_k\coloneqq (a_k,b_k)\times([-R,R]+S_k)$ have pairwise disjoint projections in the second coordinate.

\smallskip
\noindent\emph{Claim.} For each $\delta_3>0$ there is $\Phi_{\rm sh}(x)=(x^{(1)},\,x^{(2)}+\psi(x^{(1)}))\in\mathcal A_{\mathcal{F}}$ such that
$$
\left|\,(\Phi_{\rm sh})_\#\mu_{\mathrm B}^{\mathrm{S2}}-\mu_{\mathrm B}^{\mathrm{S3}}\,\right|_{\TV}<\delta_3
\qquad\text{and}\qquad
\left|\,(\Phi_{\rm sh})_\#\mu_\star^{\mathrm{S2}}-\mu_\star^{\mathrm{S3}}\,\right|_{\TV}<\delta_3.
$$

\smallskip
\noindent\emph{Proof of Claim.} We apply \textup{\ref{ass:A2}} with $J=[-R,R]$, $i=1$, and $j=2$. There exists a sequence of $\Phi_n(x)=(x^{(1)},\,x^{(2)}+\psi_n(x^{(1)}))\in\mathcal A_{\mathcal{F}}$ such that $\|\psi_n-\beta\|_{L^1([-R,R])}\to 0$.

We verify the convergence for $\mu_{\mathrm B}^{\mathrm{S2}}$ (the argument for $\mu_\star^{\mathrm{S2}}$ is identical). Recall from \textbf{(S2)} that its density is supported on $L$ and depends only on $x^{(2)}$ within each layer. Let $\bar\rho_k(x^{(2)})\coloneqq\rho_{\mathrm{B}}^{\mathrm{S2}}(x^{(1)},x^{(2)})$ denote this profile for $x\in L_k$. Because the density vanishes on the gaps $G$, the $\TV$ distance simplifies to:
$$
\left|\,(\Phi_n)_\#\mu_{\mathrm B}^{\mathrm{S2}}-\mu_{\mathrm B}^{\mathrm{S3}}\right|_{\TV}
= \frac12\sum_{k=1}^m\int_{a_k}^{b_k}\int_\R
\left|\bar\rho_k\left(x^{(2)}-\psi_n(x^{(1)})\right)-\bar\rho_k\left(x^{(2)}-S_k\right)\right|
\diff x^{(2)}\diff x^{(1)}.
$$
Since $\|\psi_n-\beta\|_{L^1}\to 0$ and $\beta\equiv S_k$ on $(a_k,b_k)$, we can extract a subsequence (not relabeled) such that $\psi_n(x^{(1)})\to S_k$ for almost every $x^{(1)}\in(a_k,b_k)$. Because translation is continuous in $L^1(\R)$, the inner integral converges to zero pointwise a.e. in $x^{(1)}$, and it is uniformly bounded by $2\|\bar\rho_k\|_{L^1}$. By the dominated convergence theorem, the double integral vanishes as $n\to\infty$. Thus, choosing $n$ large enough satisfies the claim.

\smallskip
\noindent\textbf{(S4) Matching the second marginal.}
Recall from \textbf{(S3)} that $\mu_{\mathrm B}^{\mathrm{S3}}$ and $\mu_\star^{\mathrm{S3}}$ are supported on the strips $\tilde L_k \coloneqq (a_k,b_k)\times I_k$, where $I_k\coloneqq[-R+S_k,R+S_k]$ are pairwise disjoint. On each strip, the densities are independent of $x^{(1)}$ and share the mass $m_k$.

Fix $\delta_4>0$. Let $\mu_{{\mathrm{B}},2,k}$ and $\mu_{\star,2,k}$ be the $x^{(2)}$-marginals of the measures restricted to $\tilde L_k$. For any $\delta>0$ and each $k$ with $m_k>0$, we can pick densities $\rho_{{\mathrm{B}},k}^\delta,\rho_{\star,k}^\delta\in {C}^{1,1}(I_k)$ with mass $m_k$, bounded away from zero and infinity, such that their associated measures satisfy
\begin{equation*}
\left|\mu_{{\mathrm{B}},k}^\delta-\mu_{{\mathrm{B}},2,k}\right|_{\TV}+\left|\mu_{\star,k}^\delta-\mu_{\star,2,k}\right|_{\TV}<\frac{\delta}{2m}.
\end{equation*}
For $m_k=0$, we set
$\mu_{{\mathrm{B}},k}^\delta=\mu_{\star,k}^\delta=0$.

We now define a global transport map $\alpha:\R\to\R$. For $m_k>0$, let $\alpha|_{I_k}$ be the monotone rearrangement from $\mu_{{\mathrm{B}},k}^\delta$ to $\mu_{\star,k}^\delta$ (setting $\alpha|_{I_k}=\Id$ if $m_k=0$). By the density bounds, this is an increasing bi-Lipschitz ${C}^{1,1}$-diffeomorphism from $I_k$ onto itself. Because each local map has a strictly positive derivative, we can interpolate these maps across the gaps while maintaining a uniformly positive derivative. This yields a global increasing bi-Lipschitz ${C}^{1,1}$-diffeomorphism $\alpha$ on $\R$. 

We now approximate $\alpha$ using \textup{\ref{ass:A1}}. Choose $R_0$ large enough so that all supports of $\mu_{{\mathrm{B}},k}^\delta$ and their pushforwards lie in $I_{R_0-1}\coloneqq[-R_0+1,R_0-1]$, and let $J_{R_0}\coloneqq \alpha^{-1}(I_{R_0+1})$. Since $\alpha$ is a bi-Lipschitz ${C}^1$-diffeomorphism, $m_{R_0}\coloneqq\min_{J_{R_0}}\alpha'>0$. Let $C_k$ be the constants provided by Lemma~\ref{lem:1d-bdbyw11} for $h=\alpha$, $R=R_0$, and the measure $m_k^{-1}\mu_{{\mathrm B},k}^\delta$. We then define
\begin{equation*}
C_{\max}\coloneqq \max\{C_k\mid m_k>0\}.
\end{equation*}
Letting $C_{R_0}$ denote the Sobolev embedding constant of $W^{1,1}(J_{R_0})\hookrightarrow {C}^0(\overline{J_{R_0}})$, we choose $\eta>0$ satisfying
\begin{equation*}
\eta\le 1\wedge C_{R_0}^{-1},\qquad C_{\max}\eta<\frac{\delta_4}{2}.
\end{equation*}
By \textup{\ref{ass:A1}} with $j=2$, for this $\eta$ there exists $\Phi_2(x)=(x^{(1)},\phi(x^{(2)}))\in\mathcal A_{\mathcal{F}}$ such that
\begin{equation*}
\|\phi-\alpha\|_{W^{1,1}(J_{R_0})}<\eta,
\qquad
\|\phi-\alpha\|_{L^\infty(J_{R_0})}\le C_{R_0}\eta.
\end{equation*}
For each $k$ with $m_k>0$ we apply Lemma~\ref{lem:1d-bdbyw11} with $(h,g)=(\alpha,\phi)$, $R=R_0$ and $J_{R_0}=\alpha^{-1}(I_{R_0+1})$ to the measure $m_k^{-1}\mu_{{\mathrm{B}},k}^\delta$, whose density is extended by zero outside $I_k$ (and thus belongs to $\mathrm{BV}_{\mathrm{loc}}(\mathbb R)$) with tail terms vanishing since $\alpha_\#\mu_{{\mathrm B},k}^\delta=\mu_{\star,k}^\delta$ is supported in $I_k\subset I_{R_0-1}$. The constants $C_k$ then satisfy
\begin{equation*}
\left|\,\phi_\#\mu_{{\mathrm{B}},k}^\delta-\alpha_\#\mu_{{\mathrm{B}},k}^\delta\,\right|_{\TV}
\le m_k\,C_k\,\|\phi-\alpha\|_{W^{1,1}(J_{R_0})}.
\end{equation*}

Let
\begin{equation*}
\mu_{\mathrm B}^\delta \coloneqq \sum_{k=1}^m \mathcal U((a_k,b_k))\otimes\mu_{{\mathrm{B}},k}^\delta,
\qquad
\mu_\star^\delta \coloneqq \sum_{k=1}^m \mathcal U((a_k,b_k))\otimes\mu_{\star,k}^\delta.
\end{equation*}
By the definition of the global transport $\alpha$, we have $(\mathsf T_\alpha)_\#\mu_{\mathrm B}^\delta=\mu_\star^\delta$. Furthermore, summing the marginal bounds over all strips yields
\begin{equation*}
\left|\mu_{\mathrm B}^\delta-\mu_{\mathrm B}^{\mathrm{S3}}\right|_{\TV} + \left|\mu_\star^\delta-\mu_\star^{\mathrm{S3}}\right|_{\TV} < \frac{\delta}{2}.
\end{equation*}
Since $\sum_{k=1}^m m_k \le 1$, the triangle inequality gives
\begin{equation*}
\begin{aligned}
\left|\,(\Phi_2)_\#\mu_{\mathrm B}^\delta-(\mathsf T_\alpha)_\#\mu_{\mathrm B}^\delta\,\right|_{\TV}
\le \sum_{k=1}^m \left|\,\phi_\#\mu_{{\mathrm{B}},k}^\delta-\alpha_\#\mu_{{\mathrm{B}},k}^\delta\,\right|_{\TV} \le \left(\sum_{k=1}^m m_k\right) C_{\max}\eta < \frac{\delta_4}{2}.
\end{aligned}
\end{equation*}
By Lemma~\ref{lem:TV-contraction}, we have $|\,(\Phi_2)_\#\mu_{\mathrm B}^{\mathrm{S3}}-(\Phi_2)_\#\mu_{\mathrm B}^\delta\,|_{\TV} \le |\mu_{\mathrm B}^{\mathrm{S3}}-\mu_{\mathrm B}^\delta|_{\TV}$. Therefore, another application of the triangle inequality yields
\begin{equation*}
\left|\,(\Phi_2)_\#\mu_{\mathrm B}^{\mathrm{S3}}-\mu_\star^{\mathrm{S3}}\,\right|_{\TV}
\le \left|\mu_{\mathrm B}^{\mathrm{S3}}-\mu_{\mathrm B}^\delta\right|_{\TV}
+ \left|\,(\Phi_2)_\#\mu_{\mathrm B}^\delta-\mu_\star^\delta\,\right|_{\TV}
+ \left|\mu_\star^\delta-\mu_\star^{\mathrm{S3}}\right|_{\TV} < \frac{\delta}{2} + \frac{\delta_4}{2}.
\end{equation*}
Choosing $\delta \le \delta_4$ from the beginning guarantees that $|\,(\Phi_2)_\#\mu_{\mathrm B}^{\mathrm{S3}}-\mu_\star^{\mathrm{S3}}\,|_{\TV} < \delta_4$.

\smallskip
\noindent\textbf{Conclusion of the 2D case.} Based on the construction in \textbf{(S1)}--\textbf{(S4)}, we define
\begin{equation*}
\Phi_{\rightarrow}\coloneqq \Phi_2\circ\Phi_{\rm sh}\circ\Phi_1\in\mathcal A_{\mathcal{F}},
\qquad
\Phi_{\leftarrow}\coloneqq \Phi_{\rm sh}\in\mathcal A_{\mathcal{F}}.
\end{equation*}
Setting $\nu\coloneqq\mu_\star^{\mathrm{S3}}$, the triangle inequality over the successive approximations yields
\begin{equation*}
\left|\,(\Phi_{\leftarrow})_\#\mu_\star-\nu\,\right|_{\TV} < \delta_2+\delta_3,
\qquad
\left|\,(\Phi_{\rightarrow})_\#\mu_{\mathrm B}-\nu\,\right|_{\TV} < \delta_2+\delta_3+\delta_4.
\end{equation*}
Choosing $\delta_2,\delta_3,\delta_4>0$ sufficiently small, and then choosing
$\delta_1\le\delta_2/4$ as required in \textbf{(S2)}, ensures that both total
errors are strictly less than $\varepsilon/2$. The reduction at the beginning of the proof then yields $\Phi=\Phi_{\leftarrow}^{-1}\circ\Phi_{\rightarrow}\in\mathcal A_{\mathcal{F}}$ satisfying
\begin{equation*}
\left|\,\Phi_\#\mu_{\mathrm B}-\mu_\star\,\right|_{\TV}<\varepsilon.
\end{equation*}
This completes the proof for $d=2$.

\smallskip

\noindent\textbf{Extension to arbitrary dimension.}

The case $d>2$ follows by iterating the preceding construction to simultaneously eliminate the dependence of both densities on the coordinates $x^{(1)}, x^{(2)}, \dots, x^{(d-1)}$. As in the 2D case, it suffices to build $\Phi_{\rightarrow},\Phi_{\leftarrow}\in\mathcal A_{\mathcal{F}}$ pushing $\mu_{\mathrm B}$ and $\mu_\star$, respectively, to a common measure, each within a $\TV$ error of less than $\varepsilon/2$.

For the base case $j=1$, we follow exactly the 2D steps \textbf{(S1)} to \textbf{(S3)}: we globally match the $x^{(1)}$-marginals, truncate to a compact cube, partition into slices, average in $x^{(1)}$, and separate them via a shear in $x^{(2)}$.
Assume that at the beginning of step $j \in \{2,\dots,d-1\}$, both intermediate ideal measures are supported on the same finite family of disjoint product sets (blocks) whose projections onto the $x^{(j)}$-axis are pairwise disjoint. Furthermore, assume that the two measures assign the same mass to each block, and, within each block, their densities depend only on the remaining coordinates $(x^{(j)},\dots,x^{(d)})$. We advance to step $j+1$ through the following operations:

\noindent \textbf{(S1') Marginal matching.} On each block, we approximate both local $x^{(j)}$-marginals by ${C}^{1,1}$-densities of the same mass, bounded away from zero and infinity on the corresponding projection interval. We consider the local monotone rearrangement between these approximations. Since the blocks project disjointly onto the $x^{(j)}$-axis, we can glue these local transport maps into a single global increasing bi-Lipschitz ${C}^{1,1}$-diffeomorphism $\alpha_j:\R\to\R$. By \textup{\ref{ass:A1}} for the $j$-th coordinate and Lemma~\ref{lem:1d-bdbyw11}, there exists a flow modifying only the $x^{(j)}$-coordinate that matches these source marginals to the target marginals up to a prescribed $\TV$ error $\delta_{j,1}>0$.

\noindent \textbf{(S2') Layering.} We partition each common block into finitely many thin slices along the $x^{(j)}$-direction. On each block, we replace the densities of both measures with their $x^{(j)}$-averages scaled as in \eqref{eq:RhoS2} so that both carry exactly the same mass. This yields two measures whose densities are independent of $x^{(j)}$ on each slice. By choosing $\delta_{j,1}$ small enough and the slice partition sufficiently thin, the $\TV$ error introduced by this piecewise-constant approximation is bounded by a prescribed tolerance $\delta_{j,2}>0$.

\noindent\textbf{(S3') Shearing.} We apply the same ideal shear transformation $T_{\beta_j}$ to both measures to separate the slices along the $x^{(j+1)}$-axis. We choose increasing, sufficiently spaced translations $S_{j,k}$ and define a nondecreasing, piecewise affine $\beta_j$ such that $\beta_j\equiv S_{j,k}$ on the $k$-th slice. By \textup{\ref{ass:A2}}, there exists an attainable shear flow $\Phi_{\psi_j}$ given by
\begin{equation*}
\left(x^{(1)},\dots,x^{(j)},\,x^{(j+1)}+\psi_j(x^{(j)}),x^{(j+2)},\dots,x^{(d)}\right)\in\mathcal A_{\mathcal{F}},
\end{equation*}
where $\psi_j$ approximates $\beta_j$ in $L^1$. Because the densities within each slice are exactly independent of $x^{(j)}$ prior to shearing, the $\TV$ error generated by replacing $T_{\beta_j}$ with $\Phi_{\psi_j}$ is bounded by $\delta_{j,3}>0$. 
After this ideal shear, the densities depend only on $(x^{(j+1)},\dots,x^{(d)})$, and the new slices form the family of blocks for the next level.

\noindent\textbf{Conclusion.} At  $j=d-1$, both measures are supported on the same finite collection of tubes parallel to the $x^{(d)}$-axis with mutually disjoint $x^{(d)}$-projections. On each tube, the densities depend exclusively on $x^{(d)}$ and carry identical mass. This reduces the problem exactly to the 1D setting of \textbf{(S4)}. We construct a final decoupled flow acting on $x^{(d)}$ via \textup{\ref{ass:A1}} to match these remaining densities up to an error $\delta_{d}>0$.

Let $\Phi_{\rightarrow}$ and $\Phi_{\leftarrow}$ be the compositions of all attainable flows applied to the source and target sides, respectively. By choosing the individual tolerances at each inductive stage such that the sum of all accumulated errors is strictly bounded by $\varepsilon/2$, we guarantee $|(\Phi_{\rightarrow})_\#\mu_{\mathrm B} - (\Phi_{\leftarrow})_\#\mu_\star|_{\TV} < \varepsilon/2$. The proof is complete.
\end{proof}

\begin{proof}[Proof of Corollary~\ref{cor:gm:gen_TV}]
By affine invariance, the assumption $x\mapsto \upsigma(x^{(1)})e_1 \in \mathcal{F}$ implies that $ x \mapsto c\,\upsigma(a x^{(i)}+b)e_j \in \mathcal{F}$ for all $a,b,c\in\R$ and $i,j\in[d]$. This induces the scalar control family $ \mathcal G_{\upsigma} \coloneqq \{ s \mapsto c\,\upsigma(as+b) : a,b,c\in\R \}.$

We first establish the scalar approximation property: for every compact interval $I\subset\R$ and every $\alpha\in W^{1,1}_{\mathrm{nd,loc}}(\R)$, we claim that
\begin{equation}\label{eq:gm:scalar-W11-TV}
    \alpha\in \overline{\mathcal A_{\mathcal G_{\upsigma}}}^{\,W^{1,1}(I)}.
\end{equation}
We distinguish two cases for $\upsigma$:

\smallskip\noindent\textit{Case (i): $\upsigma\in{C}^1(\R)$ is non-affine.} Since $x\mapsto\upsigma(x^{(1)})e_1$ is globally Lipschitz, $\upsigma$ itself is globally Lipschitz and therefore non-polynomial. By \cite[Thm.~4.1]{Pinkus1999Approximation}, $\operatorname{span}(\mathcal G_{\upsigma})$ is dense in ${C}^1_{\mathrm{loc}}(\R)$. Because increasing ${C}^1$-diffeomorphisms are dense in $W^{1,1}_{\mathrm{nd,loc}}(\R)$, Lemma~\ref{lem:gm:C1_to_W11} ensures that $\alpha$ can be approximated on $I$ by some flow $\Phi\in\mathcal A_{\operatorname{span}(\mathcal G_{\upsigma})}$. Finally, by Lemma~\ref{lem:gm:LieTrotter}, each autonomous segment of this composition can be approximated in $W^{1,1}$ using flows generated strictly by elements of $\mathcal G_{\upsigma}$. Concatenating these segments yields \eqref{eq:gm:scalar-W11-TV}.

\smallskip\noindent\textit{Case (ii): $\upsigma(s)=s_+$.} Here, $\mathcal G_{\upsigma}$ is precisely the one-dimensional instance of the family in \eqref{eq:gm:node}. By Lemma~\ref{lem:gm:W11_relu_d1}, $\alpha$ can be approximated in $W^{1,1}(I)$ by strictly increasing continuous piecewise affine functions, each of which exactly coincides on $I$ with the restriction of a flow in $\mathcal A_{\mathcal G_{\upsigma}}$ (Lemma~\ref{lem:gm:PWA_exact_compact}). This proves \eqref{eq:gm:scalar-W11-TV}.

\smallskip
When $d=1$, our initial observation gives $\mathcal G_{\upsigma}\subset\mathcal F$, meaning $ \mathcal A_{\mathcal G_{\upsigma}} \subset \mathcal A_{\mathcal F}.$ The scalar approximation property thus implies \eqref{eq:gm:d1approx} for $\mathcal F$, and the conclusion follows immediately from Proposition~\ref{prop:d1interp}. Assume now that $d\ge2$. We verify the three assumptions of Theorem~\ref{thm:TV-d}:

\smallskip\noindent
\textbf{\textup{\ref{ass:A0}}.}
Affine invariance guarantees that $f\in\mathcal F$ implies $-f\in\mathcal F$. For any attainable map $\Phi_{t_m}^{f_m}\circ\cdots\circ\Phi_{t_1}^{f_1} \in \mathcal A_{\mathcal F}$, its inverse is explicitly given by
\begin{equation*}
    \left(\Phi_{t_m}^{f_m}\circ\cdots\circ\Phi_{t_1}^{f_1}\right)^{-1}
    =
    \Phi_{t_1}^{-f_1}\circ\cdots\circ\Phi_{t_m}^{-f_m}.
\end{equation*}
Since $-f_k\in\mathcal F$ for each $k$, the attainable set $\mathcal A_{\mathcal F}$ is closed under inversion.

\smallskip\noindent
\textbf{\textup{\ref{ass:A1}}.}
Fix $j\in[d]$, a compact interval $I\subset\R$, $\alpha\in W^{1,1}_{\mathrm{nd,loc}}(\R)$, and $\varepsilon>0$. By \eqref{eq:gm:scalar-W11-TV}, there exists $\phi\in\mathcal A_{\mathcal G_{\upsigma}}$ such that $\|\phi-\alpha\|_{W^{1,1}(I)}<\varepsilon$. Lifting this scalar flow to the $j$-th coordinate using vector fields of the form $x\mapsto c\,\upsigma(ax^{(j)}+b)e_j$ yields the map $x \mapsto (x^{(1)}, \dots, \phi(x^{(j)}), \dots, x^{(d)}) \in \mathcal A_{\mathcal F}$, achieving the required marginal transport.

\smallskip\noindent
\textbf{\textup{\ref{ass:A2}}.}
Fix distinct $i,j\in[d]$, a compact interval $J\subset\R$, a continuous, nondecreasing, piecewise affine $\beta$, and $\varepsilon>0$. Since $\upsigma$ is non-polynomial in both cases, the universal approximation theorem provides $m\in\N$ and  $(a_k,b_k,c_k)_{k=1}^m$ such that the function $\psi(s)\coloneqq \sum_{k=1}^m c_k\,\upsigma(a_k s+b_k)$ satisfies $\|\psi-\beta\|_{L^1(J)}<\varepsilon$. For each $k\in[m]$, define
\begin{equation*}
    f_k(x)\coloneqq c_k\,\upsigma(a_kx^{(i)}+b_k)e_j \ \in \ \mathcal F.
\end{equation*}
Because $i\neq j$, the $i$-th coordinate is preserved by the flow of $f_k$, meaning $\Phi^{f_k}_1(x) = x+c_k\,\upsigma(a_kx^{(i)}+b_k)e_j$. Composing these flows yields
\begin{equation*}
    \Phi^{f_m}_1\circ\cdots\circ\Phi^{f_1}_1(x) = x+\psi(x^{(i)})e_j.
\end{equation*}
Thus \textup{\ref{ass:A2}} holds, and the theorem is proved via Theorem~\ref{thm:TV-d}.
\end{proof}

\section{Proofs of Subsection~\ref{s.discussion.interpolation}}\label{app:interpolation}

\begin{proof}[Proof of Proposition~\ref{prop:approx_labeled_unlabeled}]
The approximate \UIP clearly implies the approximate \uUIP. We prove
the converse.

Fix $N\ge1$ and, for a map $G:\R^d\to\R^d$, write
\begin{equation}\label{eq:diag.map}
G^{[N]}(x_1,\dots,x_N)
\coloneqq \left(G(x_1),\dots,G(x_N)\right).
\end{equation}
Suppose first that $d=1$, and let
$X=(x_1,\dots,x_N)$ and $Y=(y_1,\dots,y_N)$ be increasingly
ordered configurations. Every attainable map is increasing. For $N=1$, there is
nothing to prove. For $N\ge2$,  if the interpolation error is smaller than half the minimum distance between two target points, any permutation realizing the approximate
\uUIP\ must preserve the order and therefore be the identity.
Thus the approximate \UIP\ follows.

Assume now that $d\ge2$ and set
\begin{equation*}
\mathcal C_N
\coloneqq
\left\{
(x_1,\dots,x_N)\in\R^{dN}:
x_i\neq x_j \text{ whenever } i\neq j
\right\},
\end{equation*}
which is connected. The diagonal action $G^{[N]}$ commutes with the
action of the group of permutations $S_N$ by relabeling coordinates. For $X\in\mathcal C_N$, define
\begin{equation*}
C_X
\coloneqq
\overline{
\left\{
\Phi^{[N]}(X):\Phi\in\mathcal A_{\mathcal F}
\right\}
}^{\,\mathcal C_N}
\end{equation*}
where the closure is taken relative to $\mathcal C_N$.
The set $C_X$ is forward invariant. Indeed, if $Y\in C_X$,
choose $\Phi_k\in\mathcal A_{\mathcal F}$ such that
$\Phi_k^{[N]}(X)\to Y$. Then, for every
$\Psi\in\mathcal A_{\mathcal F}$,
\begin{equation*}
(\Psi\circ\Phi_k)^{[N]}(X)
\to \Psi^{[N]}(Y),
\end{equation*}
and $\Psi\circ\Phi_k\in\mathcal A_{\mathcal F}$. Hence
$\Psi^{[N]}(Y)\in C_X$, and therefore
\begin{equation}\label{eq:CX_inclusion}
Y\in C_X
\quad\Longrightarrow\quad
C_Y\subseteq C_X.
\end{equation}
Moreover, commutation with permutations gives $C_{\sigma X}=\sigma C_X $ for every $\sigma\in S_N.$

We claim that equality holds in \eqref{eq:CX_inclusion}. Let
$Y\in C_X$. Applying the approximate \uUIP\ from $Y$ to $X$
with errors tending to zero, and using the finiteness of $S_N$,
we obtain $\sigma\in S_N$ such that $\sigma X\in C_Y$. Thus
\begin{equation*}
\sigma C_X
=
C_{\sigma X}
\subseteq C_Y
\subseteq C_X.
\end{equation*}
Since $\sigma$ has finite order, $\sigma C_X\subseteq C_X$ implies
$\sigma C_X=C_X$. Thus $C_Y=C_X$.

It follows that any two sets of the form $C_X$ are either equal or
disjoint. Indeed, if $Z\in C_X\cap C_Y$, then $C_X=C_Z=C_Y.$ Fix $X\in\mathcal C_N$. For every $Y\in\mathcal C_N$, applying again the
approximate \uUIP\ 
and the finiteness of $S_N$, we obtain 
$\sigma\in S_N$ such that $\sigma Y\in C_X$. Hence
\begin{equation*}
\mathcal C_N
=
\bigcup_{\sigma\in S_N}\sigma C_X.
\end{equation*}
Each $\sigma C_X=C_{\sigma X}$ is closed in $\mathcal C_N$, and
distinct sets in this finite cover are disjoint. Hence each is also
open. Since $\mathcal C_N$ is connected, the cover consists of a
single set. In particular,
\begin{equation*}
C_X=\mathcal C_N.
\end{equation*}
Thus every target configuration can be approximated with its
prescribed labeling. Since $N$ was arbitrary, the system possesses the
approximate \UIP.
\end{proof}

\begin{proof}[Proof of Example~\ref{ex:generic-unlabeled}]
Write $x=(s,z)\in\R\times\R^{d-1}$. Since every vector field in $\mathcal F$ depends only on $s$, every attainable map $\Phi\in\mathcal A_{\mathcal F}$ has the triangular form
\begin{equation}\label{eq:att.map.ex}
\Phi(s,z)=\left(\phi(s),\,z+H(s)\right),
\end{equation}
where $\phi:\R\to\R$ is strictly increasing and $H:\R\to\R^{d-1}$. Moreover, the complement of $\mathcal G_N$ in $\R^{dN}$ is a finite union of hyperplanes, so $\mathcal G_N$ is open, dense, and of full Lebesgue measure.

\smallskip
\noindent\emph{Proof of \textup{(i)}.}
Let $X,Y\in\mathcal G_N$. Without loss of generality (up to relabeling indices), assume $x_1^{(1)}<\cdots<x_N^{(1)}$, and choose $\pi\in S_N$ such that
\begin{equation}\label{eq:order.preserve}
y_{\pi(1)}^{(1)}<\cdots<y_{\pi(N)}^{(1)}.
\end{equation}
Pick a strictly increasing, continuous piecewise affine function $h:\R\to\R$ with finitely many pieces such that $h(x_i^{(1)})=y_{\pi(i)}^{(1)}$ for all $i\in[N]$. Since $\mathcal F$ contains all scalar ReLU fields acting on the first coordinate, applying Lemma~\ref{lem:gm:PWA_exact_compact} on a compact interval containing $\{x_1^{(1)},\dots,x_N^{(1)}\}$ yields an attainable map $\Phi_1(s,z)=(\phi_1(s),z)\in\mathcal A_{\mathcal F}$ with $\phi_1(x_i^{(1)})=y_{\pi(i)}^{(1)}$ for all $i\in[N]$.

Next, write $x_i=(x_i^{(1)},z_i)$ and $y_j=(y_j^{(1)},\zeta_j)$ in $\R\times\R^{d-1}$. Since the coordinates $y_{\pi(i)}^{(1)}$ are distinct, there exists a piecewise affine map $q:\R\to\R^{d-1}$ of the form
\begin{equation*}
q(s)=v_0+\sum_{k=1}^m v_k(s-a_k)_+
\qquad\text{such that}\qquad
q\left(y_{\pi(i)}^{(1)}\right)=\zeta_{\pi(i)}-z_i
\quad\text{for all }i\in[N],
\end{equation*}
where $v_0,\dots,v_m\in\R^{d-1}$ and $a_1,\dots,a_m\in\R$. The vector fields $(s,z)\mapsto(0,v_0)$ and $(s,z)\mapsto(0,v_k(s-a_k)_+)$ belong to $\mathcal F$ and leave $s$ fixed, so composing their time-$1$ flows yields $\Phi_2(s,z)=(s,z+q(s))\in\mathcal A_{\mathcal F}$. Thus, $\Phi\coloneqq\Phi_2\circ\Phi_1\in\mathcal A_{\mathcal F}$ maps each $x_i$ to $y_{\pi(i)}$.

\smallskip
\noindent\emph{Proof of \textup{(ii)}.}
Since $d\ge2$, we may take the configurations of distinct points $X=(x_1,x_2)=(0,e_2)$ and $Y=(y_1,y_2)=(0,e_1)$. By \eqref{eq:att.map.ex}, every $\Phi\in\mathcal A_{\mathcal F}$ maps both $x_1$ and $x_2$ to points with first coordinate $\phi(0)$, whereas $y_1^{(1)}=0$ and $y_2^{(1)}=1$. Hence, for every $\pi\in S_2$,
\begin{equation*}
\max_{i=1,2}\|\Phi(x_i)-y_{\pi(i)}\|
\ge |\phi(0)|\,\vee\,|\phi(0)-1|\ge\frac12.
\end{equation*}
Thus the system fails the approximate \uUIP\ (and consequently also the approximate \UIP). In fact, the approximate \UIP\ fails even when restricted to $\mathcal G_2$: for $X=(0,e_1)\in\mathcal G_2$ and $Y=(e_1,0)\in\mathcal G_2$, the monotonicity $\phi(0)<\phi(1)$  forces $\max_{i=1,2}\|\Phi(x_i)-y_i\|\ge 1/2$.
\end{proof}

\begin{proof}[Proof of Proposition~\ref{prop:approx_exact_UIP}]
Fix $N\ge1$ and $\bar X=(\bar x_1,\dots,\bar x_N)$
satisfying \eqref{ass.maxrank}, increasingly ordered if $d=1$. Recall from \eqref{eq:diag.map} the notation for a diagonal map. By \eqref{ass.maxrank}, there exist
$f_1,\dots,f_{dN}\in\mathcal F$ such that
\begin{equation*}
A=
\begin{bmatrix}
f_1^{[N]}(\bar X) & \cdots & f_{dN}^{[N]}(\bar X)
\end{bmatrix}
\in\R^{dN\times dN}
\end{equation*}
is invertible. Define $\Gamma:\R^{dN}\to\R^{dN}$ by
\begin{equation*}
\Gamma(\alpha)
=
\left(
\Phi_{\alpha_{dN}}^{f_{dN}}
\circ\cdots\circ
\Phi_{\alpha_1}^{f_1}
\right)^{[N]}(\bar X).
\end{equation*}
Negative-time flows are attainable because
$\mathcal A_{\mathcal F}$ is closed under inversion.
The local flow expansion established in the proof of
Lemma~\ref{lem:gm:LieTrotter} gives
$\Gamma(\alpha)=\bar X+A\alpha+o(\|\alpha\|)$
as $\alpha\to0$.

Because $A$ is invertible, we have $\sigma\coloneqq\min_{\|v\|=1}\|Av\|>0$. Taking $r>0$ sufficiently small,
$\|\Gamma(\alpha)-\bar X-A\alpha\|<\sigma r/2$ on $\|\alpha\|=r$.
Hence, for every $Z\in B_{\sigma r/2}(\bar X)$,
\begin{equation*}
\|\Gamma(\alpha)-Z-A\alpha\|
<\sigma r
\le \|A\alpha\|,
\qquad \|\alpha\|=r.
\end{equation*}
Shrinking $r$, we may also assume that
$B_{\sigma r/2}(\bar X)$ consists of configurations of distinct
points, increasingly ordered when $d=1$. By homotopy invariance of the Brouwer degree,
\begin{equation}\label{eq:brouwer}
\deg(\Gamma-Z,B_r(0),0)
=
\deg(A,B_r(0),0)
\neq0.    
\end{equation}
Thus, for every $Z\in B_{\sigma r/2}(\bar X)$, there exists
$\alpha\in B_r(0)$ such that $\Gamma(\alpha)=Z$. 

Now, for any configuration of distinct points $X$, approximate \UIP yields
$\Psi\in\mathcal A_{\mathcal F}$ with $\Psi^{[N]}(X)\in B_{\sigma r/2}(\bar X)$.
By \eqref{eq:brouwer}, we can choose $\Phi\in\mathcal A_{\mathcal F}$ satisfying
$\Phi^{[N]}(\bar X)=\Psi^{[N]}(X)$; then
$(\Psi^{-1}\circ\Phi)^{[N]}(\bar X)=X$.
Every configuration of distinct points (of
increasing order if $d=1$) is therefore exactly reachable from $\bar X$.
Reversibility and concatenation then give the \UIP.

\end{proof}

\bibliographystyle{plainnat}
\bibliography{references}

\end{document}